\documentclass[final,5p,times,twocolumn]{elsarticle}

\usepackage{lineno}
\usepackage{comment}
\usepackage{amsmath}
\usepackage{float}
\usepackage{graphicx}
\usepackage{pdfpages} 
\modulolinenumbers[5]
\usepackage{pgfplots}
\usepackage{hyperref}
\usepackage{amssymb}
\usepackage{multirow}
\usepackage{gensymb}
\usepackage{multicol}
\usepackage{makecell}
\usepackage{subfigure}

\journal{arXiv}

\begin{document}
\begin{frontmatter}

\title{ \huge Deep Neural-Kernel Machines}

\author{Siamak Mehrkanoon\corref{cor1}}

\cortext[cor1]{Corresponding author}

\address{Department of Data Science and Knowledge Engineering, Maastricht University, The Netherlands}

\begin{abstract}\label{intro}
In this chapter we review the main literature related to the recent advancement of deep neural-kernel architecture, an approach that seek the synergy between two powerful class of models, i.e. kernel-based models and artificial neural networks. The introduced deep neural-kernel framework is composed of a hybridization of the neural networks architecture and a kernel machine. More precisely, for the kernel counterpart the model is based on Least Squares Support Vector Machines with explicit feature mapping. Here we discuss the use of one form of an explicit feature map obtained by random Fourier features. Thanks to this explicit feature map, in one hand bridging the two architectures has become more straightforward and on the other hand one can find the solution of the associated optimization problem in the primal, therefore making the model scalable to large scale datasets. We begin by introducing a neural-kernel architecture that serves as the core module for deeper models equipped with different pooling layers. In particular, we review three neural-kernel machines with average, maxout and convolutional pooling layers. In average pooling layer the outputs of the previous representation layers are averaged. The maxout layer triggers competition among different input representations and allows the formation of multiple sub-networks within the same model. The convolutional pooling layer reduces the dimensionality of the multi-scale output representations. Comparison with neural-kernel model, kernel based models and the classical neural networks architecture have been made and the numerical experiments illustrate the effectiveness of the introduced models on several benchmark datasets.
\end{abstract}

\begin{keyword}
Neural Networks \sep Kernel machines \sep convolutional neural network \sep explicit feature mapping
\end{keyword}
\end{frontmatter}

\section{Introduction}
Deep Learning techniques have recently achieved the state-of-the-art performances in different application domains. 
Thanks to their representation learning power, they can analyse complex tasks by learning from subtasks \cite{bengio2013representation,bengio2009learning} . In particular, it learns multiple levels of hierarchical representation from the given raw input data by means of successive nonlinear modules that are stacked in a hierarchical architecture. Thanks to the the staked nonlinear layers, the learnt representation (features) at one level is transformed into a 
slightly more abstract representation at a higher level \cite{mehrkanoon2018deep}. Recent years have witnessed the significant impact of various deep learning architectures including for instance Restricted Boltzmann Machines  \cite{salakhutdinov2009deep,hinton2010practical,nair2010rectified}, Stacked Denoising Autoencoders \cite{vincent2010stacked,hinton2006reducing}, , Convolutional Neural Networks \cite{lawrence1997face,krizhevsky2012imagenet}, Long Short Term Memories \cite{hochreiter1997long} among others. Deep architectures have shown their prominent superiority over other machine learning in several application domains \cite{bengio2013representation,lecun2015deep}. However, it should be noted that training deep neural networks is a labeled data demanding process and involves optimizing an expensive non-convex optimization problem. A deep artificial neural networks model often posses thousands to millions of parameters to be learned, which heavily influence its generalization performance. In addition, searching the proper architecture such as the type of activation functions, the number of hidden units/layers, the networks hyper-parameters, the learning rates among others, become a laborious task when the complexity of deep model increases. Moreover training deep networks demand lots of computing and memory resources. On the other kernel based models such as support vector machines (SVM) have also made a large impact in a wide range of application domains \cite{Vapnik98,shawe2004kernel,Bishop}. kernel based models are well established with strong foundations in learning theory and optimization. They are well suited for problems with limited training instances and are able to extend linear methods to nonlinear ones with theoretical guarantees \cite{mehrkanoon2018deep}. However, in their classical formulation, they cannot learn features from raw data and do not scale well to the size of the training datasets. One often requires to choose a relevant kernel amongst the well known ones a priori, or defines/learns an appropriate kernel for the data under study.
Deep learning models and Kernel machines have mostly been investigated separately. Furthermore, most of the developed successful deep models are based on artificial neural networks (ANN) architecture. On the other hand, deep kernel machines have not yet been explored in great detail. Both deep learning models and kernel machines have their strengths and weaknesses and can potentially be considered as complementary family of methods with respect to the settings where they are most relevant. Therefore, exploring hybridization between ANN and kernel machines appears worthwhile to pursue and can lead to the development of models that benefit from the advantages of the two frameworks. 

The literature has already witnessed such research direction for instance one can refer to kernel models for deep learning \cite{cho2009kernel}, deep Gaussian processes \cite{damianou2013deep,cutajar2017random}, convolutional kernel networks \cite{mairal2014convolutional} and a convex deep learning model using normalized kernels \cite{aslan2014convex}. The iterated compositions of Gaussian kernels are investigated in \cite{steinwart2016learning}. The authors in \cite{mairal2014convolutional} proposed a kernel based convolutional neural network.
Other deep kernel models that allow working with a hierarchy of representations are discussed in \cite{cho2009kernel,heinemann2016improper,jose2013local,montavon2011kernel}. A brief overview of existing research works on hybridization techniques for bridging artificial neural networks and kernel machines are discussed in \cite{Lluis}.

Mehrkanoon et~al. in \cite{siamakESANN17,mehrkanoon2018deep,mehrkanoon2019deep,mehrkanoon2019cross} introduced a new line of research that aims at combining the kernel machines and artificial neural networks architectures in order to achieve the best of two class of models. To this end, they introduced hybrid layers constructed by means of kernel functions that are used in a neural networks architecture. In particular, the authors in \cite{mehrkanoon2018deep} introduced neural-kernel model by means of deploying hybrid layers constructed by explicit feature map using random Fourier features in the neural network architectures. In related study, the author in \cite{mehrkanoon2019deep} explored the use of different choices of merging layers and introduced three types of kernel blocks. i.e. average, maxout and convolutional kernel blocks.

It is the purpose of this chapter to give a detailed overview of the recent advancements on the neural-kernel studies. 
Precisely, we start by presenting the basics of kernel as well as artificial neural networks based modelling and their differences.
Next the use of an explicit and implicit feature mapping in kernel models are discussed and in particular an overview of two popular approximation techniques for constructing the explicit feature mapping is given. Afterwards, the neural-kernel model formulation and its various choices of architectures are explained. Lastly, the performance of the described models on several real-life benchmark datasets are analysed.


\section{Kernel Machines vs Neural Networks }
\begin{figure}[ht]
	\begin{center}
		\includegraphics[width=3.5in, height=2in]{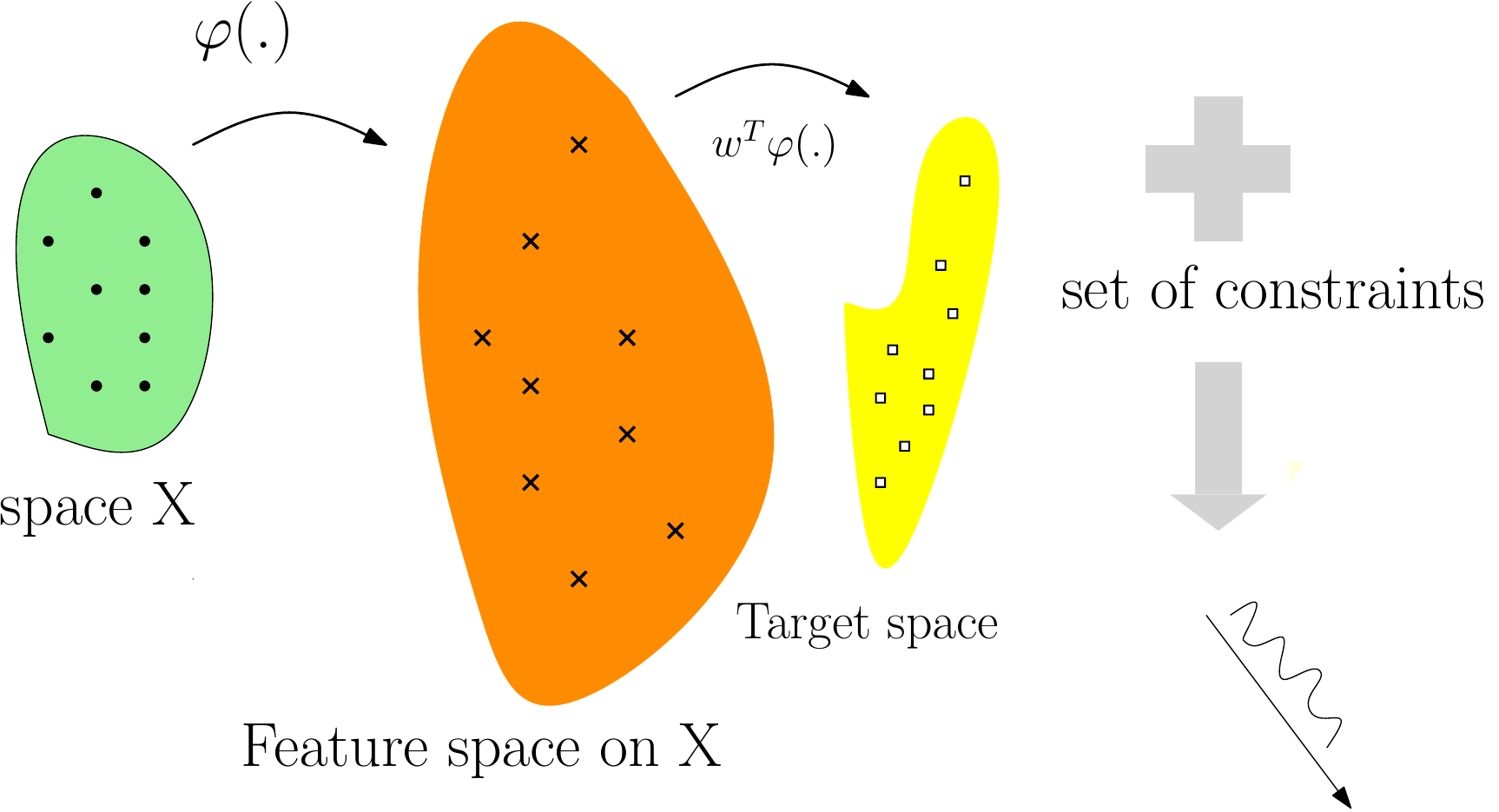}
	\end{center}
	\caption{ \label{fig_kenel_model} Schematic illustration of Kernel based modelling}
\end{figure}

Kernel machines framework and in particular Support Vector Machine (SVM) is a powerful methodology for solving complex pattern recognition and function estimation tasks. In this method, one often works with a primal-dual setting by expressing the solution in the primal in terms of feature maps and seeking the optimal representation of the model in the dual. In the primal, the input data are projected into a high dimensional feature space by means of an implicit feature mapping. The projected data points are then mapped to a target space using an inner product with a weight matrix \cite{mehrkanoon2018deep}, see Fig \ref{fig_kenel_model}. The feature mapping $\varphi(\cdot)$ is not in general explicitly known and can be infinite dimensional. Therefore, the kernel trick is applied and the solution is sought in the dual by solving a quadratic programming problem
\cite{Johan_book}. A variant of SVM is the Least Squares SVM (LS-SVM) where the problem formulation involves equality instead of inequality constraints. This leads to a system of linear system of equations at the dual level, in the context of regression, classification, semi-supervised learning and domain adaptation \cite{Johan_book,mehrkanoon_ssl2,mehrkanoon2017regularized}. Thanks to the primal-dual setting, the prior knowledge and side information can be incorporated in the primal formulation using proper regularization terms. For instance the authors in \cite{mehrkanoon2012approximate,mehrkanoon2012ls,mehrkanoon2015learning} proposed a systematic approach based on LS-SVM for learning the solutions of different types of differential equations. Furthermore, in \cite{mehrkanoon2014parameter,mehrkanoon2012parameter} an extension of LS-SVM formulation have been proposed for parameter estimation of dynamical systems governed by ordinary as well as delay differential equations. It should be noted that in the kernel based model the size of dual problem depends on the number of training samples. In case of large scale problems solving the problem becomes memory demanding. 
Therefore, when the number of training instances is much larger than that of feature dimensions, we can avoid solving the problem in the dual by constructing an explicit feature map and seeking the solution in the primal. In contrary, if the number of variables (feature dimensions) is much larger than the that of training instances then thanks the kernel trick one can seek the solution by solving the dual optimization problem.
The neural networks architecture differs from kernel based models. In neural networks modelling, the input data is projected into a new space via linear transformation (multiplication with a weight matrix) followed by applying a nonlinear activation function. One can consider the described operations in a module and form a deep architecture by stacking several of these modules. Thus, depending on the configurations of stacking these modules as well as the used connection scheme, one can generate several choices of architectures suitable for different modeling tasks. It should be noted that the projection space of an implicit feature mapping in kernel machines corresponds to a hidden layer in a neural network with infinite number of neurons \cite{Johan_book}. The bridge between neural network architectures and kernel models becomes even closer when an explicit feature map is used. More precisely, the number of hidden units in the hidden layer of a neural network architecture corresponds to the dimension of the explicit feature map used in kernel machines.

\subsection{Implicit and Explicit Feature Mapping}
\begin{figure}[htbp]  
	\begin{center}
		
		\includegraphics[width=3.5in, height=1.7in]{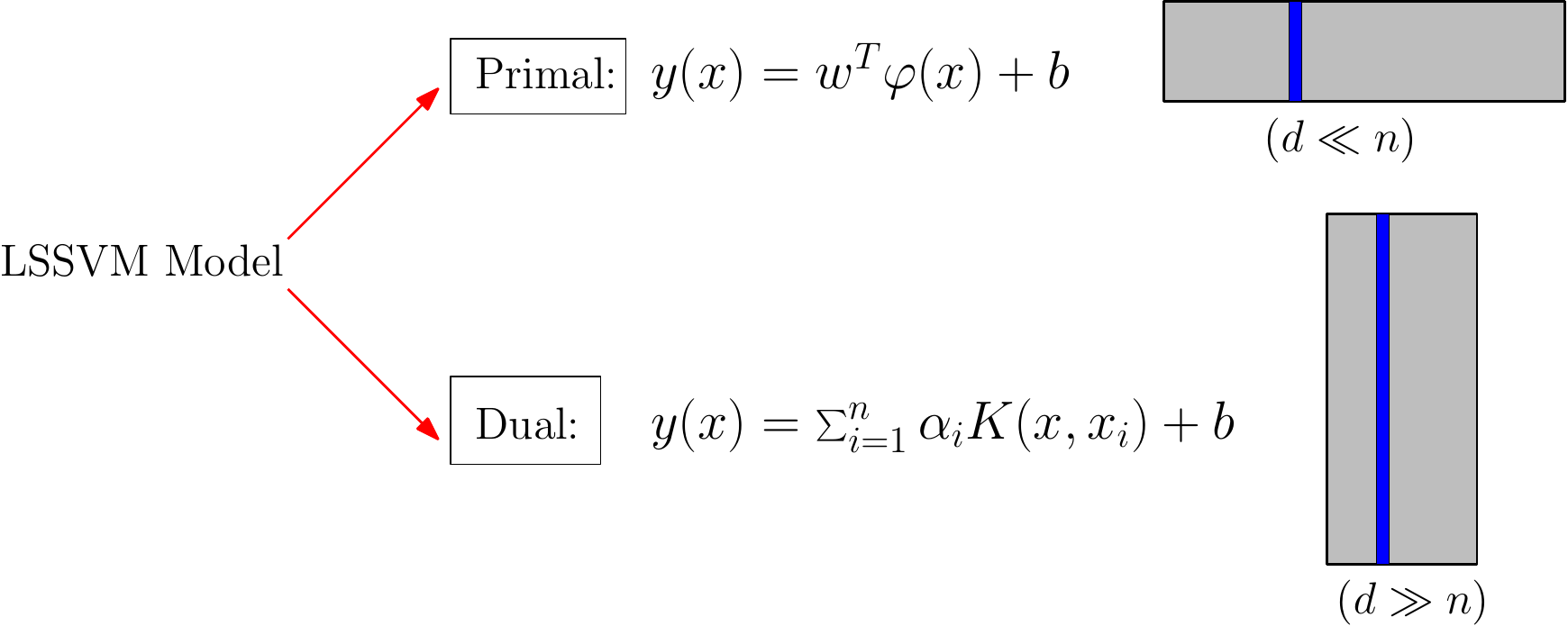} 
		
	\end{center}
	\caption{ \label{lssvm_fig} LS-SVM model formulation is the primal and dual. (Image adapted from \cite{mehrkanoon2018deep}).}
\end{figure}

\begin{figure}[ht]
	\begin{center}
		\begin{tabular}{c}
			

			\subfigure[]
			{
				\includegraphics[width=2.5in, height=0.8in]{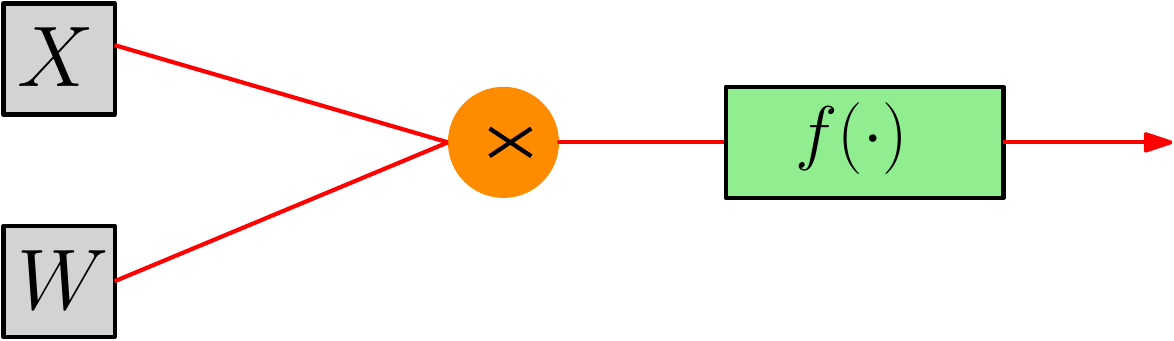}
			}
			
			\\\\
			
			\subfigure[]
			{
				\includegraphics[width=2.5in, height=0.8in]{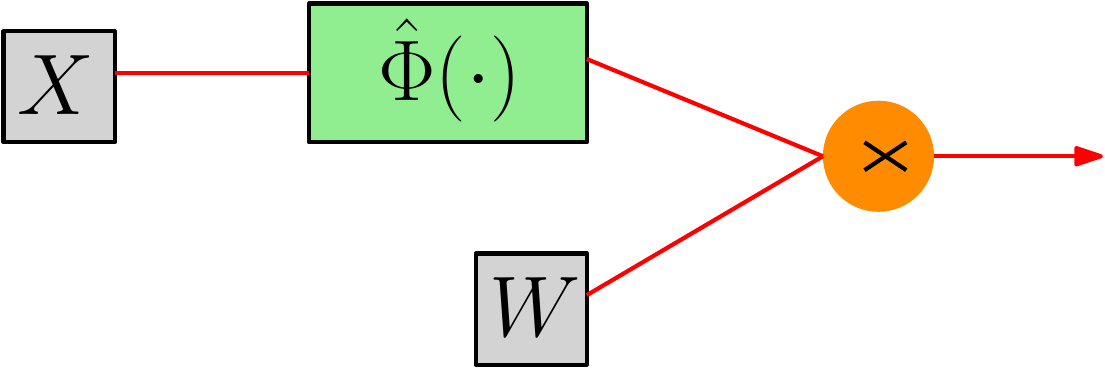}
			}

		\end{tabular}
	\end{center}
	\caption{ \label{fig_new} (a) A single module of a neural network model. (b) A single module of kernel based model with explicit feature mapping. (Images adapted from \cite{siamakESANN17}).}
\end{figure}

There are various kernel based models which follow different approaches. For instance in Reproducing Kernel Hilbert Spaces (RKHS) \cite{andrew2000introduction} the problem of function
estimation is treated as a variational problem. The Gaussian Processes (GP) \cite{rasmussen2003gaussian} follow a probabilistic Bayesian setting. Here we focus on the Least Squares Support vector machines (LS-SVM) which follows a primal-dual formulation and any additional constraints can be added to the formulation and therefore makes it straightforward to integrate prior knowledge into the model \cite{Johan_book}. Given training data points $\mathcal{D}=\{x_1,...,x_n\}$, where $\{x_i\}_{i=1}^n \in \mathbb{R}^{d}$ with targets $\{y_i\}_{i=1}^n$, LS-SVM framework for regression problems assumes that the underlying function describing the complex input-output relation of the system can be written in the primal as follows \cite{Johan_book}: 
\begin{equation}
y(x)=w^ T\varphi(x)+b,
\end{equation}
where $\varphi(\cdot):\mathbb{R}^d \rightarrow \mathbb{R}^{h}$ is the feature map and $h$ is the dimension of the feature space which can be infinite. Thanks to the nonlinear feature map, the input data are embedded into a feature space where the optimal solution is sought by minimizing the residual between the real measurements and the model outputs via solving the following optimization problem \cite{Johan_book}:
\begin{equation} \label{lssvmreg}
\begin{aligned}
& \underset{w,b,e}{\text{minimize}}
& & \frac{1}{2}w^Tw+\frac{\gamma}{2}e^Te \\
& \text{subject to} & & y_i=w^T \varphi(x_i)+b+e_i, \ \
i=1,...,n,
\end{aligned}
\end{equation}
where $\gamma \in \mathbb{R}^+, b \in \mathbb{R}$, $w \in
\mathbb{R}^{h}$. Here, one may choose to work with implicit or explicit feature maps. Implicit feature maps are usually unknown and infinite dimensional. Therefore, after obtaining the Karush-Kuhn-Tucker (KKT) optimality conditions, applying the kernel trick as well as eliminating the primal variables $e_i$ and $w$, the solution is obtained in the dual as follows \cite{Johan_book}:
\begin{equation} \label{dual_lssvm}
\left[%
\renewcommand*{\arraystretch}{1.4}
\begin{array}{c|c}
\Omega +  \gamma^{-1}\mathbf{\mathit{I}}_{n}& 1_n  \\ \hline
1_n^T & 0 \\
\end{array}
\right] \left[
\begin {array}{c}
\alpha \\  \hline
b
\end{array}
\right] = \left[
\begin {array}{c}
y \\ \hline
0
\end{array}
\right]
\end{equation}
where $\Omega_{ij}=K(x_i,x_j)=\varphi(x_i)^T
\varphi(x_j)$ is the $ij$-th entry of the positive definite
kernel matrix. $1_n=[1,\ldots,1]^T\in \mathbb{R}^n$,
$\alpha=[\alpha_1,\ldots,\alpha_n]^T$,
$y=[y_1,\ldots,y_n]^T$ and $I_n$ is the identity matrix. The LSSVM 
model for regression problem in the dual form becomes \cite{Johan_book}:
\begin{equation}
y(x)=w^T \varphi(x) +b = \sum_{i=1}^n \alpha_i K(x,x_i) +b.
\end{equation}
For large-scale problems, the cost of storing as well as computing the solution vectors of (\ref{dual_lssvm}) can be prohibitive due
to the size of the kernel matrix. An alternative approach would be for instance to use a low rank approximation of the kernel matrix. Among kernel approximation techniques are for instance Incomplete Cholesky Factorization \cite{golub2012matrix}, Nystr{\"o}m method \cite{williams2001}, randomized low-dimensional feature space \cite{rahimi2007random} or reduced kernel technique \cite{lee2001rsvm,mehrkanoon2014large}. In this way one can construct an explicit feature map, and rewrite the optimization problem (\ref{lssvmreg}) in the primal as follows: 
\begin{equation} \label{explicitsvm}
\small
\begin{aligned}
& \underset{\hat{w},b}{\min} J(\hat{w},b)= \hat{w}^{T} \hat{w}  + \frac{\gamma}{2} \sum_{i=1}^{n} (y_i- \hat{w}^T \hat{\varphi}(x_i) - b)^2, 
\end{aligned}
\end{equation}
where $\hat{\varphi}(\cdot)$ is an explicit finite dimensional feature map. The primal and dual LSSVM formulations corresponding to explicit and implicit feature maps are depicted in Fig. \ref{lssvm_fig}.
Next, let us analyse the difference between a single layer neural network model and a shallow kernel based model shown in Fig. \ref{fig_new}. 
In artificial neural network architectures the non-linearity is explicitly imposed by means of non-linear activation functions which operate on the weighted sum of the input layer. Among activation functions are hyperbolic tangent, sigmoid or Rectified Linear Units (ReLU).  
Considering a single-layer with activation function $f$, one can formulate the input-output relation as $y(x)=f(W x + b)$.
Here $x \in x \in \mathbb{R}^{d}$, $W \in x \in \mathbb{R}^{d_h \times d}$ and $b$ is the bias vector.
On the contrary, in the kernel machines the nonlinear feature maps are implicit, usually unknown and infinite dimensional. They directly operate on the input instances and project them to a feature space.  The model output then is formulated as a weighted sum over the dimension of the projected sample. In the case of explicit feature maps $\hat{\varphi}(\cdot)$, its dimension can be larger or smaller than the dimension of the training instances. 
Moreover, a kernel based model is linear in the weight matrix $W$, and therefore convex optimization techniques can be applied to obtain an optimal values of $W$. In what follows we discuss some of the approximation techniques for constructing the explicit feature maps. 


In the context of kernel machines, several methodologies are introduced in the literature to scale up the Kernel models for analysing large scale problems. Among them are kernel approximation techniques such greedy basis selection approach \cite{smola2000sparse}, incomplete Cholesky decomposition \cite{bach2005predictive}, and Nystr{\"o}m method \cite{williams2001using}, reduced kernel technique \cite{lee2001rsvm}, random features approximation \cite{yang2012nystrom}. Among the exiting methods, Nystr{\"o}m approximation \cite{williams2001using} and random features approximation \cite{rahimi2007random} are the two popular ones. 
The former aims at approximating the kernel matrix using a low-rank decomposition. The latter is based on mapping the input features into a feature space where dot products between elements of that space can well approximate the kernel function. In what follows these two successful techniques will be discussed.

\subsection{Nystr{\"o}m method}

Consider the previously defined training data points $\mathcal{D}=\{x_1,...,x_n\}$. One can use an eigenvalue decomposition of the kernel matrix $\Omega$ to obtain an explicit expression for feature map $\varphi$.
In particular, in Nystr{\"o}m approximation method, for the given instance $x \in \mathbb{R}^d$, the $i$-th component of the $n$-dimensional feature map $\hat{\varphi}: \mathbb{R}^d \rightarrow \mathbb{R}^n$ can be obtained as follows \cite{Johan_book}:
\begin{equation}  
\hat{\varphi}_i(x) = \frac{1}{ \sqrt{\lambda_i}}  \sum_{k=1}^n  u_{ki}\, K(x_k,x),
\end{equation}
where $K(\cdot,\cdot)$ is the kernel function. The eigenvalues and eigenvectors of the kernel matrix $\Omega_{n \times n}$ whose $(i,j)$-th element is defined as $K(x_i,x_j)$ are denoted by $\lambda_i$ and $u_i$.  The $k$-th element of the $i$-th eigenvector is also denoted by $u_{ki}$ denotes. It should be noted that when the sample size $n$ is large, one can alternatively work with a subsample of size $m \ll n$. In this case, the $m$-dimensional feature map $\hat{\varphi}: \mathbb{R}^d \rightarrow \mathbb{R}^m$ is approximated as follows \cite{Johan_book,mehrkanoon2018deep}:
\begin{equation} \label{omgea_eigen_decom}
\hat{\varphi}(x)=[\hat{\varphi}_1(x), \ldots, \hat{\varphi}_m(x)]^T,
\end{equation}
where
\begin{equation} \label{Nystrom}
\hat{\varphi}_i(x) = \frac{1}{\sqrt{\lambda_i}}  \sum_{k=1}^m  u_{ki}\, K(x_k,x), i=1,\ldots,m,
\end{equation}
where $\lambda_i$ and $u_i$ are eigenvalues and eigenvectors of the kernel matrix $\Omega_{m \times m}$ which is constructed using the selected subsamples (prototype vectors). The subsamples can be selected by several approaches including incomplete Cholesky factorization \cite{bach2005predictive}, clustering, entropy based methods and random selection. In the context of semi-supervised learning, the authors in \cite{mehrkanoon2016scalable} empirically compared the performance of three kernel approximation techniques obtained based on Nystr{\"o}m approximation, the reduced kernel techniques and random Fourier features. In particular, it was shown in \cite{mehrkanoon2016scalable} that the model with random Fourier features demands less training computation times while the keeping the test accuracy comparable to that of other two techniques. 

\subsection{Random Fourier features}
One of the most popular approaches for scaling up kernel based models is random Fourier features (RFF) proposed in \cite{rahimi2007random}.
In this method one approximates the original kernel by mapping the input features into a new space spanned by a small number of Fourier basis. The basic Principe of Random Fourier features boils down to computing feature map $\hat{\varphi}(\cdot)$, such that for the given two data points $x_i$ and $x_j$ the inner product $<\hat{\varphi}(x_i),\hat{\varphi}(x_j)>$ can well approximate the kernel function $K(x_i,x_j)$. The authors in \cite{rahimi2007random} exploited the classical Bochner's theorem in harmonic analysis and introduced Random Fourier features in the field of kernel methods.
A continuous kernel $K(x,y)$ defined on $\mathbb{R}^d$ is positive definite if and only if $K$ is the Fourier transform of a non-negative measure \cite{rahimi2007random}. If a shift-invariant kernel $k$ is properly scaled, its Fourier transform $p(\zeta)$ is a proper probability distribution \cite{mehrkanoon2018deep}. Thanks to this property, kernel functions can be approximated using linear projections on $D$ random features as follows \cite{rahimi2007random}:
\begin{equation}
K(x-y)=\int_{\mathbb{R}^d} p(\zeta) e^{j \zeta^T(x-y)}d\zeta = \mathbb{E}_\zeta[z_{\zeta}(x)z_{\zeta}(y)^*],
\end{equation}
where $z_{\zeta}(x)=e^{j\zeta^Tx}$. Here $z_{\zeta}(x) z_{\zeta}(y)^*$ is an unbiased estimate of $K(x,y)$ when $\zeta$ is drawn from $p(\zeta)$ (see \cite{rahimi2007random}).
To obtain a real-valued random feature for $K$, one replaces the $z_{\zeta}(x)$ by the mapping $z_{\zeta}(x)=[\cos(\zeta^Tx), \sin(\zeta^Tx)]$. The random Fourier features $\hat{\varphi}(x)$, for the sample $x$, are then defined as 
\begin{equation}
\hat{\varphi}(x)=\frac{1}{\sqrt{D}}[z_{\zeta_1}(x), \ldots, z_{\zeta_D}(x)]^T \in \mathbb{R}^{2D},
\end{equation}
where $\frac{1}{\sqrt{D}}$ is used as a normalization factor to reduce the variance of the estimate and $\zeta_1,\ldots,\zeta_D \in \mathbb{R}^d$ are sampled from $p(\zeta)$. If the variables $\zeta$ are drawn according to a Gaussian distribution then the method is shown to approximate the Gaussian kernel. In this case, they are drawn from a Normal distribution $\mathcal{N}(0,\sigma^2 I_d)$ where $\sigma$ is the parameter of the Gaussian distribution that generates the random features which at the same time is related to the kernel bandwidth parameter. It should be mentioned in practice there exists two approaches, i.e. \textit{data-independent} and \textit{data-dependent}, to select the variable $\zeta$. In standard RFF, they are selected independently from training data, so they are randomly sampled from $p(\zeta)$. In the second approach one uses the training data to guide the generation of random Fourier features. (see \cite{sinha2016learning,agrawal2018data}). The authors in \cite{cho2009kernel,pandey2014learning} showed that by replacing the $\cos(\cdot)$ and $\sin(\cdot)$ activation functions by a Rectified Linear Unit (ReLU), one can approximate the arc-cosine kernel instead of the radian basis function kernel.
The quality of the random Fourier features for shift-invariant kernels are analysed in \cite{sriperumbudur2015optimal}.

\section{Neural-Kernel Machines}\label{deep_nk}
In this section, we will give a detailed overview of deep neural-kernel machines. We start by introducing the deep neural-kernel networks which will further be used as first building blocks for developing neural-kernel models equipped with multi-scale representations and different choices of pooling Layers.

\subsection{Deep neural-kernel networks} 
\begin{figure}[h!]
	\centering
	\includegraphics[scale=0.58]{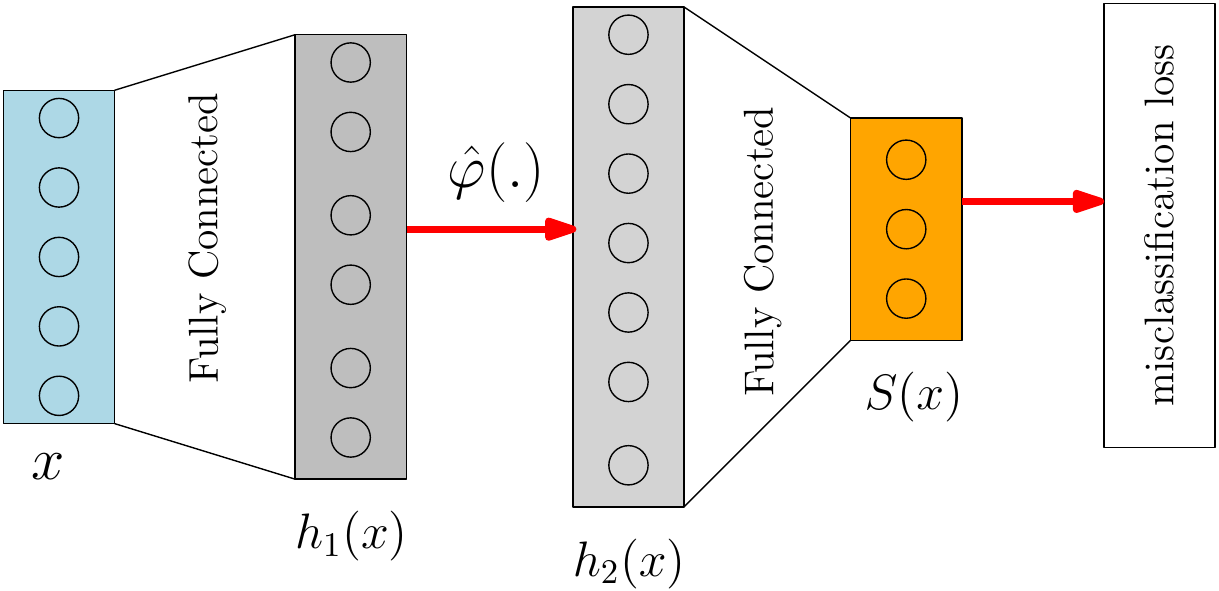}
	\caption{\label{Fig:deep} Two-layer hybrid neural kernel network architecture. (Image adapted from \cite{siamakESANN17}).}
\end{figure}

Bridging the neural networks and kernel machines frameworks can potentially lead to new model design that have the best of the two worlds. Authors in \cite{siamakESANN17,mehrkanoon2018deep} proposed a deep hybrid architecture that aims at bridging the artificial neural networks and kernel models. To this end, an explicit feature mapping based on random Fourier features is used for the kernel model. It should however be noted that other possible methods such as greedy based selection techniques \cite{smola2000sparse} or Nystr{\"o}m methods \cite{williams2001using} can alternatively be used for obtaining explicit feature maps. Assuming that there are $Q$ class in the given datasets, a two layer hybrid architecture using explicit feature map shown in Fig. \ref{Fig:deep} is formulated as follows \cite{siamakESANN17}:
\begin{equation} \label{hybrid_equation}
h_1 = W_1 x  + b_1, \;\;\; h_2=\hat{\varphi}(h_1), \;\;\; s = W_2 h_2  + b_2.
\end{equation}    
where $W_1 \in \mathbb{R}^{d_1 \times d}$ and $W_2 \in \mathbb{R}^{Q \times d_2}$ are weight matrices and the bias vectors are denoted by $b_1$ and $b_2$. 

\begin{figure*}[htbp]
	\centering
	\includegraphics[width=4in, height=1.5in]{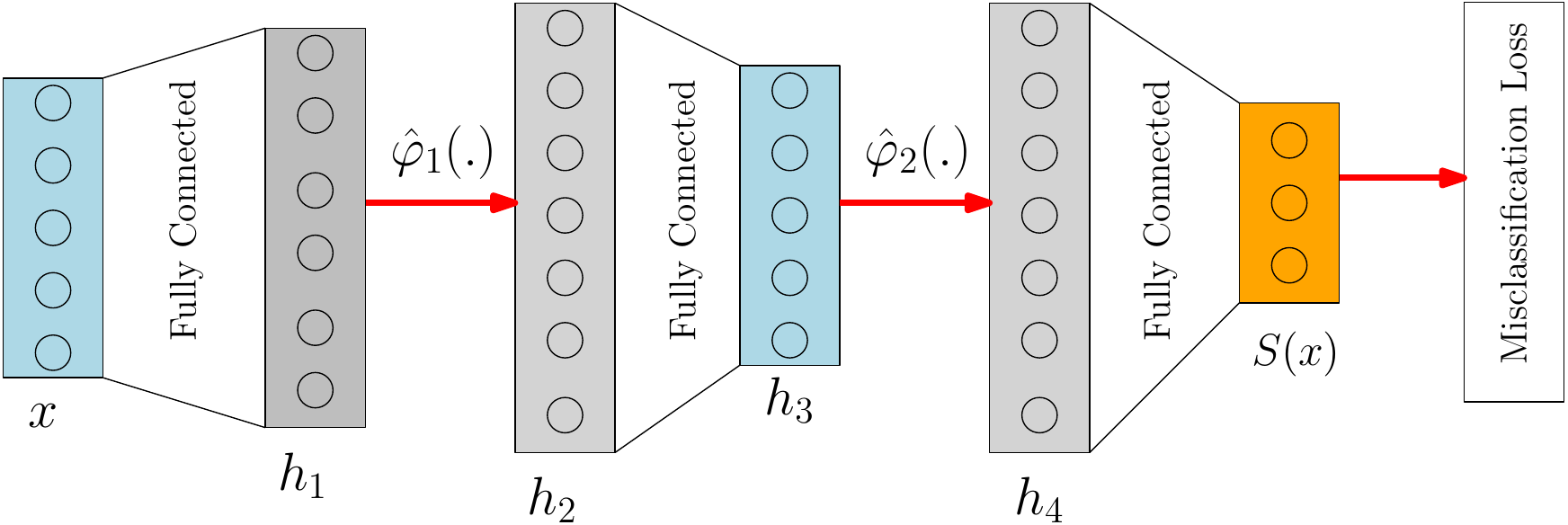}
	\caption{\label{Fig:stacked_deep} Deep hybrid neural kernel network architecture. (Image adapted from \cite{mehrkanoon2018deep}).}
\end{figure*}

The exiting connections with the standard neural networks architecture becomes more clear, if one re-writes (\ref{hybrid_equation}) as $s = W_2 \hat{\varphi}(W_1 x  + b_1) + b_2$. The dimensions of the hidden variables $h_1$ and $h_2$ are user defined parameters which here are set to $d_1$ and $d_2$. In order to learn the model parameters, one can formulate the following objective optimization problem:
\begin{equation} \label{PMSSKSC}
\underset{W_1,W_2,b_1,b_2}{\min} \;\; J(W_1,W_2,b_1,b_2)=\frac{\gamma}{2} \sum_{j=1}^{2} Tr(W_j W_j^T) + \frac{1}{n} \sum_{i=1}^{n} L(x_i,y_i).
\end{equation}
\normalsize
It consists of a misclassification loss function $L(\cdot)$ and the regularization terms. The added regularization terms aim at keeping the weights of the model small to avoid overfitting. The emphasis given to the regularization terms is controlled by $\gamma$ where needs to be properly selected as too low values can result in neglecting the regularization term and too high values encourage zero weights. Here we use the cross-entropy loss function (softmax function) which provides probabilistic membership assignments to each instance by minimizing the negative log likelihood of the correct class. The cross-entropy loss for the instance $x_i$ whose class scores are denoted by $s^{\ell}_i$ for $\ell=1,\ldots,Q$ are calculated as follows: 
\begin{equation} \label{loss}
L(x_i,y_i)=-\log\bigg(\frac{\exp(s_i^{y_i})}{\sum_{j=1}^{Q}\exp(s_i^{j})}\bigg),
\end{equation}
where $s_i^{j}$ denotes the $j$-th class score for the $x_i$ instance with ground truth label $y_i$. Given the instance $x_i$, the softmax classifier outputs a normalized probability assigned to the correct label $y_i$. 
The model can be trained using (stochastic) gradient descent algorithm. 
The class label for the test instance $x_{\text{test}}$ is computed as follows \cite{mehrkanoon2018deep}: 
\begin{equation}  \label{test_label}
\hat{y}_{\text{test}}= \underset{\ell=1,\ldots,Q}{\arg\!\max} (s_{\text{test}}),
\end{equation}
where $s_{\text{test}} = W_2 \; \tilde{\varphi}(W_1 x_{\text{test}} + b_1) + b_2$. It is shown in \cite{mehrkanoon2018deep} that one can develop deeper models by staking the hybrid model (\ref{hybrid_equation}), where in this case the input data can pass through several hybrid layers before reaching the output layer. (See Fig. \ref{Fig:stacked_deep}).
One can formulate the stacked hybrid deep neural kernel model as follows:
\begin{equation} \label{stacked_hybrid_equation}
\begin{aligned}
h_1 = W_1 x  + b_1, \\
h_2 = \hat{\varphi_1}(h_1),  \\
h_3= W_2 h_2 + b_2,  \\
h_4 = \hat{\varphi_2}(h_3), \\ 
s = W_3 h_4  + b_3. \\
\end{aligned}
\end{equation}    
Here, the input $x \in \mathbb{R}^{d}$, $\{W_i\}_{i=1}^{3}$ are weight matrices and the bias vectors are denoted by $b_1, b_2$ and $b_3$. The dimension of the hidden layers $h_1, h_2, h_3$ and $h_4$ are denoted by $d_1,d_2,d_3$ and $d_4$ respectively. 
Similar type of objective functions can be defined to learn the parameters of the model (\ref{stacked_hybrid_equation}). Several strategies can potentially be applied for training the stacked model. Similar to the idea related to transfer learning, one can use the previously learned weights and transfer them from model (\ref{PMSSKSC}) to the new model (\ref{stacked_hybrid_equation}) and keep it unchanged. This helps in reducing the training computation time as there are less number of parameters to be learned. 
Alternatively, the learned weights of the model (\ref{PMSSKSC}) can be used for the initialization of the first two layers in (\ref{stacked_hybrid_equation}). Here, we start by training the two-layer hybrid neural kernel networks (\ref{PMSSKSC}). Then we drop the last fully connected and the Softmax layer from model (\ref{hybrid_equation}) and build the new stacked model, see Fig. \ref{Fig:stacked_deep}. In training the new model, the parameters of the first two layers are not trained (are frozen). The advantage of this approach can be more noticed when analyzing large scale dataset.


Next we explore the use of competitive pooling layers in the context of neural-kernel architectures proposed in the previous subsection. In particular, here we discuss three types of neural-kernel blocks proposed in \cite{mehrkanoon2019deep} with average, maxout and convolutional pooling layers. The deep neural-kernel blocks are built using a stacking strategy which enables learning multi-scale representations of the data corresponding to different levels in network hierarchy.
The maxout neural-kernel architecture uses the maxout layer which receives as input multiple linear transformations of its previous layer. The output of the maxout flows through an explicit feature mapping constructed by random Fourier features.
The convolutional neural-kernel architecture uses a pointwise convolutional layer which receives the concatenated multiple linear transformations of the previous layer. Choosing less number of (1x1)-filters than the dimension of the input of a pointwise convolutional layer results in obtaining an output that has less dimension compared to the input dimension.  The average kernel block on the other hand is obtained by using an average layer which can be shown that it is a special case of the pointwise convolutional kernel block. 
\subsection{Deep maxout neural-kernel network}

Among important changes in classical artificial neural networks that the literature has witnessed is the use of piece-wise linear activation functions such as ReLU \cite{nair2010rectified,glorot2011deep}, maxout \cite{goodfellow2013maxout} and their variants. 

\begin{figure*}[htbp]
	\centering
	\includegraphics[width=4.5in, height=2.6in]{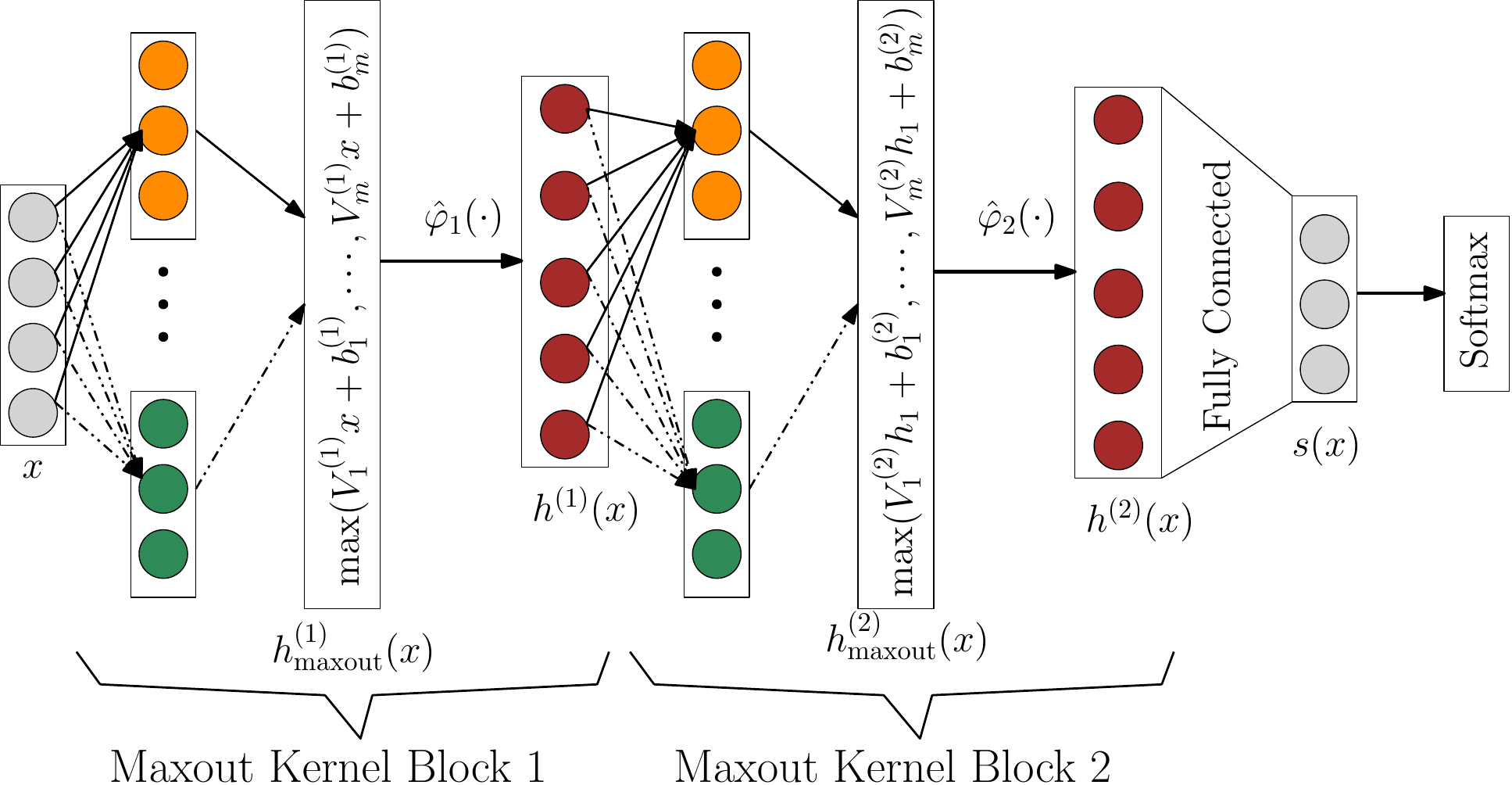}
	\caption{\label{Fig:maxout} Deep maxout neural-kernel networks. (Image adapted from \cite{mehrkanoon2019deep}).}
\end{figure*}

The piece-wise linear activation units partitions the input space into several and consequently simpler sub-problems can potentially be addressed by training several sub-networks jointly. 
In particular, the input space is partitioned into as many regions as the number of inputs when the the maxout activation units are applied \cite{goodfellow2013maxout,mehrkanoon2019deep}. As stated in \cite{goodfellow2013maxout}, the maxout activation function can approximate any convex function of the input by computing the maximum over a set of input linear functions. Here, we first begin by describing the maxout kernel block. The deep maxout neural-kernel networks is then built by stacking several maxout kernel blocks. The maxout neural-kernel block consists of constructing multiple linear transformations of the input and sending them through the maxout units. The explicit feature maps then are then applied on the output of the maxout units \cite{mehrkanoon2019deep}. Based on the definition of the maxout unit, it returns the largest value of its inputs. In other words, the activation function of each hidden unit is learned in the maxout layer. 
Assume that $h^{(\ell-1)}$ denotes the representation of the $(\ell-1)$-th layer, $V^{(\ell)}_{k} \in \mathbb{R}^{d_{\ell} \times d}$ and $b^{(\ell)}_{k} \in \mathbb{R}^{d_{\ell}}$ are the weight matrices and the bias vectors of the $(\ell)$-th layer for $k=1,\ldots,m$ respectively where $m$ is the number of linear transformations. Then the $(\ell)$-th maxout kernel block can be formulated as as follows \cite{mehrkanoon2019deep}:
\begin{align} \label{maxout-block}
h^{(\ell)} = \hat{\varphi}^{(\ell)}(h^{(\ell)}_{\text{maxout}}),  \;\;\; \textrm{where} \;\;\;\; h^{(\ell)}_{\text{maxout}} = \underset{k \in \{1,\ldots,m\}}{\max} \; V^{(\ell)}_k h^{(\ell-1)}  + b^{(\ell)}_k.
\end{align}
The original input $x$ corresponds to$h^{(0)}$ . A deep maxout neural-kernel network can then be formulated as follows \cite{mehrkanoon2019deep}:
\begin{equation} \label{maxout-maxout-neuarl-kernel_net}
\begin{cases}
h^{(1)} &= \hat{\varphi}^{(1)}(h^{(1)}_{\text{maxout}}), \;\;\; \textrm{where} \;\;\;\;
h^{(1)}_{\text{maxout}} = \underset{k \in \{1,\ldots,m\}}{\max} \;
V^{(1)}_k x  + b^{(1)}_k, \\  
h^{(2)} &=\hat{\varphi}^{(2)}(h^{(2)}_{\text{maxout}}), \;\;\; \textrm{where} \;\;\;\; h^{(2)}_{\text{maxout}} = \underset{k \in \{1,\ldots,m\}}{\max} \; V^{(2)}_k h^{(1)}  + b^{(2)}_k,\\
s(x) &= W h^{(2)}+ b.
\end{cases}
\end{equation}
Here $W \in \mathbb{R}^{Q \times d_2}$ and $b \in \mathbb{R}^{Q}$ denote the weight matrix and the bias vector of the fully connected layer before the softmax layer respectively. 
The parameters of the described model can be obtained by solving the following optimization problem \cite{mehrkanoon2019deep}:
\begin{equation} \label{cost_fun}
\begin{aligned}
\underset{\Theta}{\min} \; J(\Theta)= \gamma \Omega(\theta) + \frac{1}{n} \sum_{i=1}^{n} L(x_i,y_i),
\end{aligned}
\end{equation}
where $\Theta$, $\Omega(\Theta)$ and $L(\cdot,\cdot)$ are the trainable model parameters, regularization term and the cross-entropy loss. 
As stated previously, the emphasis given to the regularization term is controlled by the regularization parameter $\gamma$.
The architecture of the described maxout neural-kernel network is shown in Fig. \ref{Fig:maxout}.
The score variable $s_{\text{test}}$ of the test instance $x_{\text{test}}$, can be obtained by substituting $x_{\text{test}}$ in (\ref{maxout-maxout-neuarl-kernel_net}). Therefore, the class label for the test instance $x_{\text{test}}$ is computed as follows \cite{mehrkanoon2019deep}: 
\begin{equation}  \label{test_label_maxout}
\hat{y}_{\text{test}}= \underset{\ell=1,\ldots,Q}{\arg\!\max} (s_{\text{test}}).
\end{equation}
There are several possible strategies that one can choose to train the proposed deep architecture including block wise or end-to-end learning schemes. In the former, the parameters of the first block are learned, frozen and used when learning the parameters of the second block \cite{mehrkanoon2019deep}. In the later scheme, all the model parameters are jointly learned in an end-to-end fashion and this is the approach that we have adopted here.
Among existing regularization techniques one can for instance deploy dropout, batch normalization or applying various penalties on each block parameters. Here, an early stopping criterion which keeps track of the validation loss is applied. 

%

\subsection{Deep average neural-kernel network}

The average pooling layer can be used to fuse multiple representations by giving equal weight to each of the them. 
The authors in \cite{mehrkanoon2019deep} formulated the $(\ell)$-th average neural-kernel block as follows:
\begin{equation}
\begin{aligned} \label{avr-block}
h^{(\ell)}=& \hat{\varphi}^{(\ell)}(h^{(\ell)}_{\text{average}}), \\
\textrm{where} \; h^{(\ell)}_{\text{average}}=& \text{mean} \{V^{(\ell)}_1 h^{(\ell-1)}  + b^{(\ell)}_1, \ldots, V^{(\ell)}_m h^{(\ell-1)}  + b^{(\ell)}_m\}.
\end{aligned}
\end{equation}
Here, the input representation $h^{(\ell-1)}$, the weight matrices and the bias vectors $V^{(\ell)}_{k}$ and $b^{(\ell)}_{k}$ for $k=1,\ldots,m$ are defined as previously. In particular, the original input data $x$ is denoted by $h^{(0)}$.

\begin{figure*}[h]
	\centering
	\includegraphics[width=4.5in, height=2.6in]{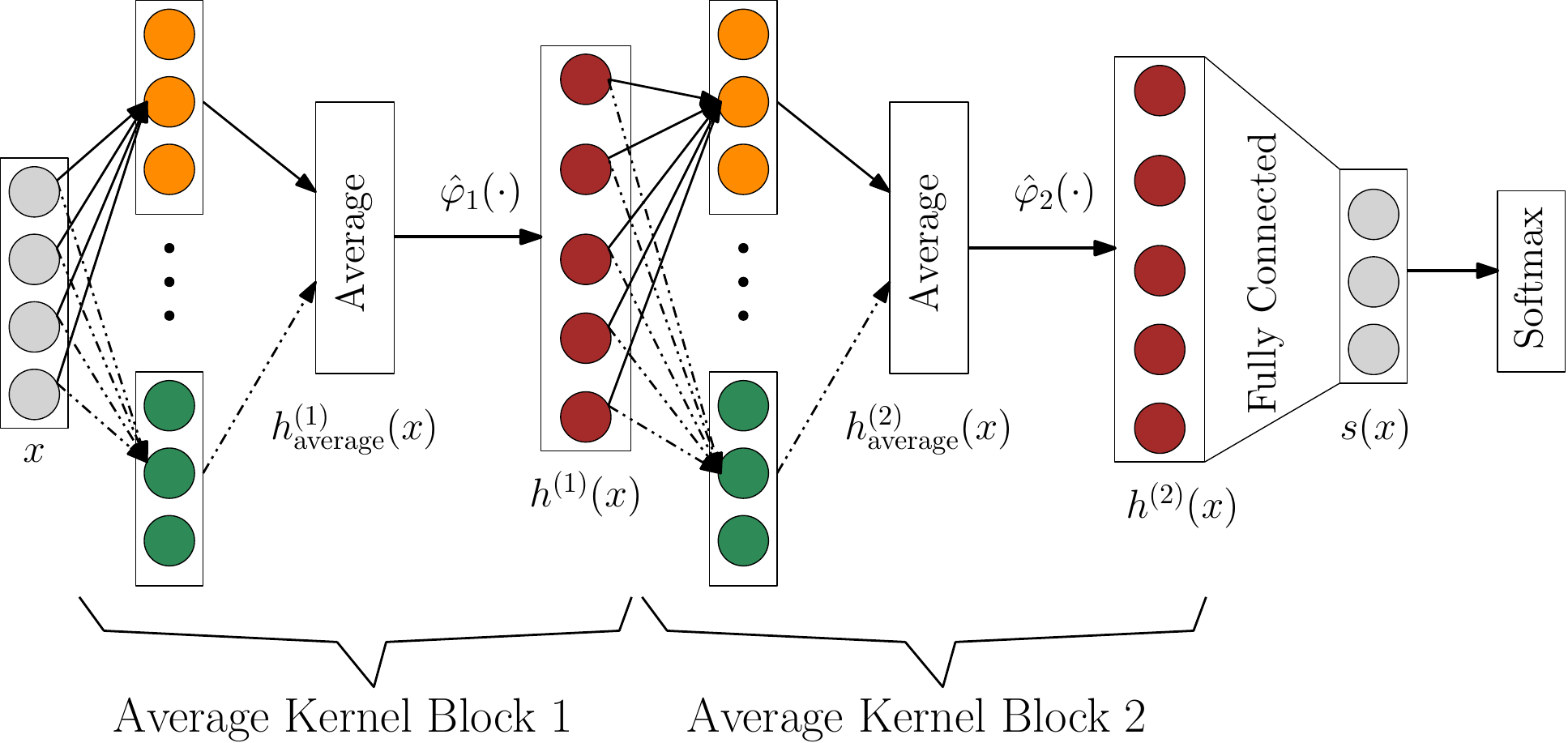}
	\caption{Deep average neural-kernel networks. Image adapted from \cite{mehrkanoon2019deep}).}\label{Fig:average}
\end{figure*}

One can stack several average neural-kernel blocks to form a deep average neural-kernel networks, see Fig. \ref{Fig:average}.
The extension of the (\ref{avr-block}) can then be made by formulating the deep average neural-kernel network as follows \cite{mehrkanoon2019deep}:
\begin{equation} \label{average-kernel-netwrok}
\begin{cases}
h^{(1)}_{\text{average}}=& \text{mean} \{V^{(1)}_1 x  + b^{(1)}_1, \ldots, V^{(1)}_m x  + b^{(1)}_m\},\\
h^{(1)}=& \hat{\varphi}^{(1)}(h^{(1)}_{\text{average}}), \\ 
h^{(2)}_{\text{average}}=& \text{mean} \{V^{(2)}_1 h^{(1)}  + b^{(2)}_1, \ldots, V^{(2)}_m h^{(1)}  + b^{(2)}_m\},\\
h^{(2)}=& \hat{\varphi}^{(2)}(h^{(2)}_{\text{average}}),\\ 
s(x) =& W h^{(2)}+ b.
\end{cases}
\end{equation}
The weight matrix and the bias vector of the fully connected layer before the softmax layer are also denoted by $W \in \mathbb{R}^{Q \times d_2}$ and $b \in \mathbb{R}^{Q}$ respectively. The hidden layers dimensions are defined as previously. 

\subsection{Deep convolutional neural-kernel network}

The convolutional neural-kernel network utilizes a pointwise convolutional pooling layer which aims at projecting multiple representations onto a new space \cite{mehrkanoon2019deep}. A pointwise convolution is a $(1\times 1)$-convolutional layer that applies the filters along the depth dimension to increase  or reduce the dimensionality of the given input (see \cite{lin2013network}). 
Following the lines of \cite{mehrkanoon2019deep}, in order to keep the number of hidden units among the maxout, average and convolutional neural-kernel networks comparable, here only one filter in the $(1\times 1)$-convolutional layer is used. However, it should be noted that in practice one could potentially deploy more number of filters. Assume that $m$ concatenated linear transformations of the input $h^{(\ell-1)}$ is denoted by $P^{(\ell)}=\bigg[V^{(\ell)}_1 h^{(\ell-1)}  + b^{(\ell)}_1, \ldots, V^{(\ell)}_m h^{(\ell-1)}  + b^{(\ell)}_m\bigg]_{d_{\ell} \times m}$. It is shown in \cite{mehrkanoon2019deep} that the result of applying the $(1 \times 1)$-convolution operations with one filter on $P^{(\ell)}$ is equivalent to calculating the following matrix multiplication:
\begin{equation} \label{cnn_formula}
P^{(\ell)} D^{(\ell)} 1_{m}.
\end{equation}
Here $D^{(\ell)}$ is a diagonal matrix whose diagonal elements are the filter parameter $d^{(\ell)}$ corresponding to the $\ell$-th convolutional layer.
$h^{(0)}$ is defined as previously and a $1_m$ denotes a vector of all ones of size $m$. 
\begin{figure*}[htbp]
	\centering
	\includegraphics[width=4.5in, height=2.6in]{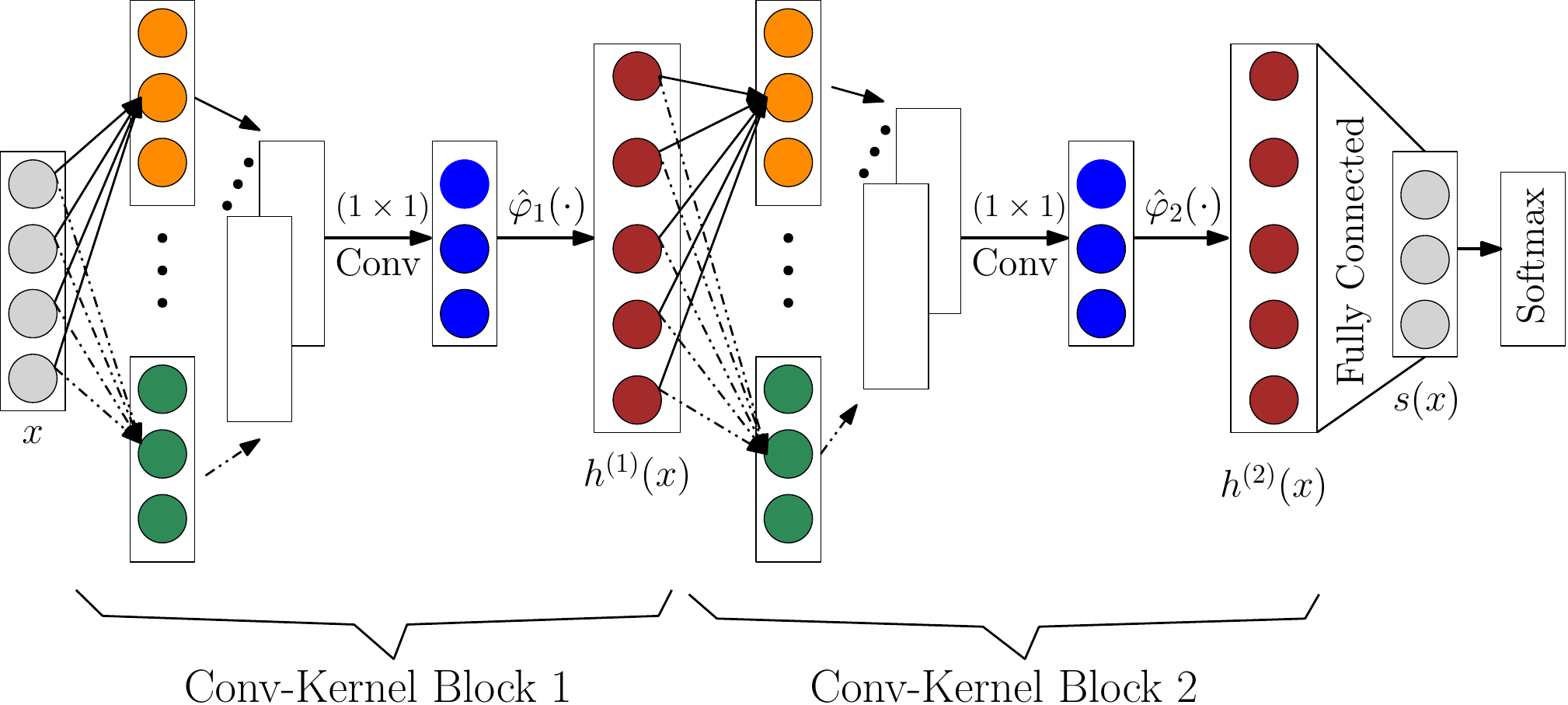}
	\caption{Deep convolutional neural-kernel networks. (Image adapted from \cite{mehrkanoon2019deep}).}\label{Fig:cnn}
\end{figure*}
Equation (\ref{cnn_formula}) reveals the difference between convolutional kernel blocks and the average kernel block layers. The convolutional kernel block can be reduced to the average kernel block, if we set the filter parameter $d^{(\ell)}$ to $\frac{1}{m}$. Therefore, average kernel block can be considered as a special case of the convolutional kernel block.
The $(\ell)$-th convolutional neural-kernel block can be formulated as follows \cite{mehrkanoon2019deep}:
\begin{equation} \label{avr-block2}
h^{(\ell)}=\hat{\varphi}^{(\ell)}(h^{(\ell)}_{\text{conv}}), \;\;\;\; \textrm{where} \;\;\;\; h^{(\ell)}_{\text{conv}} = P^{(\ell)} D^{(\ell)} 1_{m}.
\end{equation}
One can further Stack the convolutional kernel blocks as shown in Fig. \ref{Fig:cnn} to form a deep convolutional kernel network formulated as follows \cite{mehrkanoon2019deep}:
\begin{align} \label{cnn-kernel-netwrok}
\begin{cases}
h^{(1)}&= \hat{\varphi}^{(1)}(h^{(1)}_{\text{Conv}}), \;\;\;\; \textrm{where} \;\;\;\;
h^{(1)}_{\text{Conv}}= P^{(1)} D^{(\ell)} 1_{m},\\
h^{(2)}&= \hat{\varphi}^{(2)}(h^{(2)}_{\text{Conv}}), \;\;\;\; \textrm{where} \;\;\;\; 
h^{(2)}_{\text{Conv}} = P^{(2)} D^{(\ell)} 1_{m},\\
s(x) &= W h^{(2)}+ b.
\end{cases}
\end{align}
$W \in \mathbb{R}^{Q \times d_2}$ and $b \in \mathbb{R}^{Q}$ denote the  weight matrix and the bias vector of the fully connected layer before the softmax layer respectively. The hidden layers dimensions are defined as previously. 
\section{Experimental results and discussion}
We have conducted several experiments on real-life datasets taken from UCI machine learning repository \cite{UCI} and image benchmark datasets. The descriptions of the used UCI datasets are summarized in Table \ref{dataset}. 
We follow the same setup as in \cite{mehrkanoon2019deep} and therefore the given UCI dataset is randomly partitioned to $80\%$ training and $20\%$ test sets respectively. 
The hyper-parameters of the discussed deep neural-kernel networks are the dimension of the explicit feature maps and the variance of the normal distribution \cite{mehrkanoon2019deep}. Following the lines of \cite{mehrkanoon2019deep}, the dimensions of the intermediate layers are sought in the range of $[50, 600]$ using the random search scheme.  In all the experiments the regularization parameter is set to $\gamma=0.0001$. The employed architectures composed of two neural-kernel blocks with four linear transformations within each block (see Fig. \ref{Fig:maxout}).

\begin{figure*}[htbp]
\centering
\includegraphics[width=4in, height=2in]{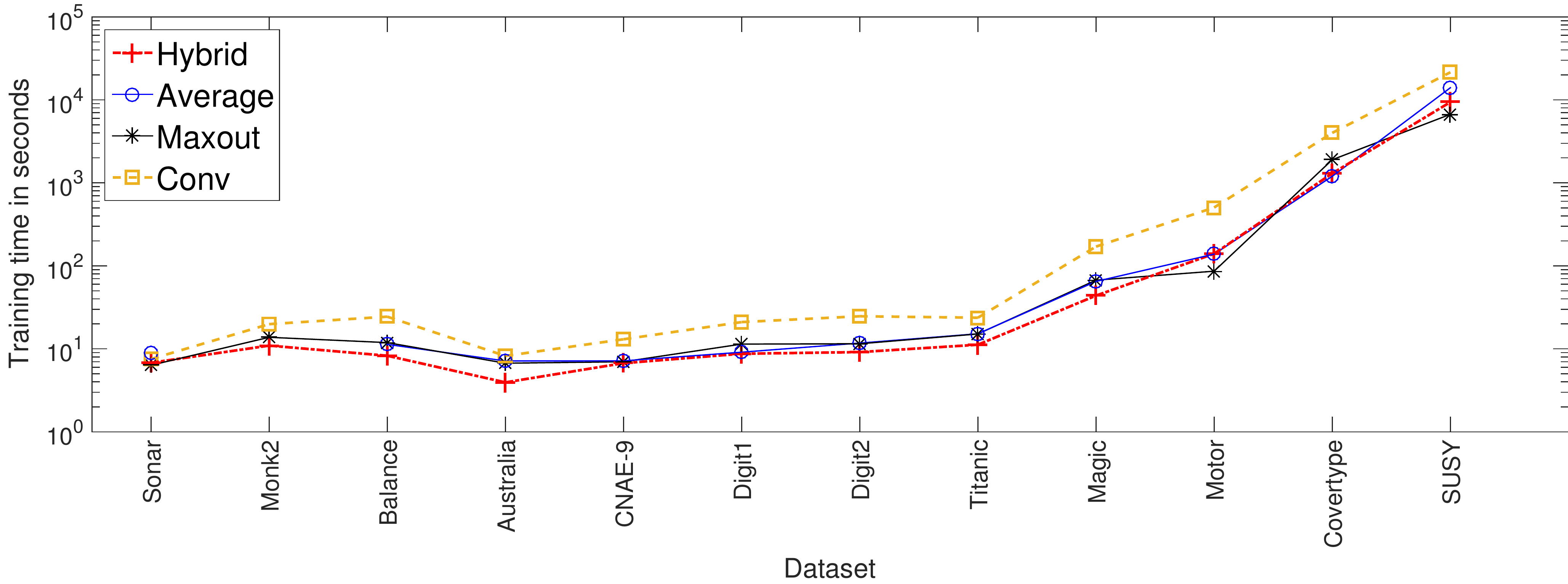} \\	
\caption{The training computation time of deep hybrid, maxout, average and convolutional neural-kernel networks.}\label{training_time}
\end{figure*}

\begin{figure*}[htbp]
\centering
\includegraphics[width=4in, height=2in]{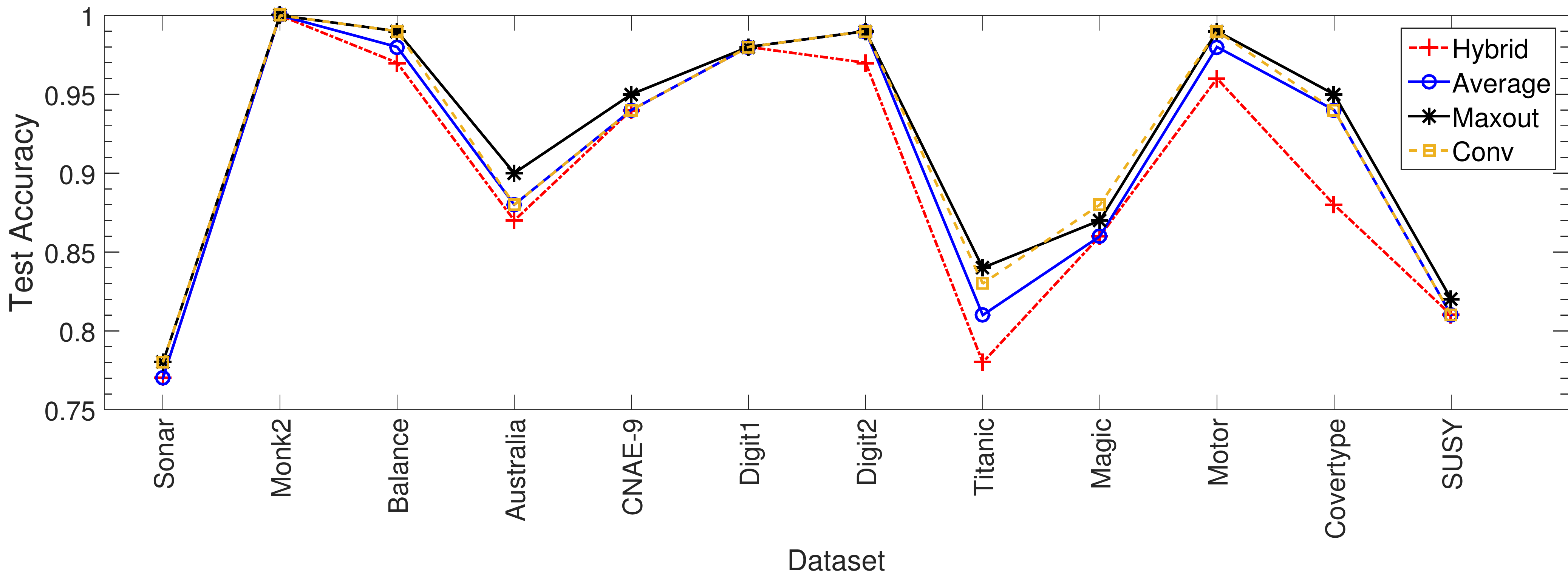} \\	
\caption{The test accuracy of deep hybrid, maxout, average and convolutional neural-kernel networks.}\label{test_acc_comp}
\end{figure*}

Table \ref{result} reports the obtained results of the introduced models in section \ref{deep_nk}.
It can observed that deep maxout neural-kernel network architecture outperforms average neural-kernel model and is comparable to the convolutional neural-kernel model. In particular one can notice a significant test accuracy improvement for the Titanic, Covertype and Motor datasets. 
The training computation time and test accuracy of the discussed deep hybrid, average, maxout and convolutional neural-kernel networks are shown in \ref{training_time} and \ref{test_acc_comp} respectively.  Fig. \ref{monk2_val_acc_loss} and Fig. \ref{motor_val_acc_loss} illustrate the accuracy, training and validation loss of the deep neural-kernel models discussed in section \ref{deep_nk} for the Covertype and Motor datasets respectively.


Here, an early stopping criterion is used to avoid overfitting. In particular, the validation loss has been monitored and after twenty consecutive epochs of no improvement the training process is stopped. It should be mentioned that the changing patterns of the validation loss is not the same for all the models and therefore as can be seen from Fig. \ref{monk2_val_acc_loss} the training process of some models stop earlier than the others.
In general we observed that the training/validation losses of the introduced deep maxout, average and convolutional neural-kernel networks are lower than those of deep hybrid model. Therefore, the deep maxout, average and convolutional neural-kernel models are expected to show a better generalization performance on the test in case there is no distributional mismatch between test and training datasets.

Fig. \ref{motor_weights_1} shows the learned weights of the first block of the deep maxout neural-kernel networks applied on the Motor dataset. The the learned weights of the second block for the same dataset are also depicted in Fig. \ref{motor_weights_2}. It can be noted that the magnitude and structure of the weight matrices corresponding to the first and second block differs.
In particular, the weights of the first block seem to have more sparse preserved structure where only in few positions one can observe the picks in the magnitude. This is due to the fact that still the nonlinear transformation is not applied. One can notice changes in the structure of the weights corresponding to the second block compared to the first block. This can be explained by the imposed nonlinearity.
Inspecting the the magnitude can also potentially reveal the amount of emphasis the network is giving to each transformation. Fig. \ref{Motor} illustrates the The t-SNE visualization \cite{maaten2008visualizing} of the hidden layer projections and the score variables corresponding to the employed deep maxout neural-kernel architecture for the Motor dataset. 
Thanks to this visualization technique, one can observe the changes in the data representations as the data flows through the stacked hierarchical layers. Ideally, one may expect that the learnt representations in the deeper layer form a more separable clusters corresponding to the existing classes \cite{mehrkanoon2018deep}.

\normalsize{
	\begin{table}[htbp]%
		\centering
		\caption{ \label{dataset} Dataset statistics}
		\begin {tabular}{lrrrr}
		
		Dataset  & \# Instances  & \# Attributes  & \# Classes  \\  \hline 
		
		Sonar & 208 & 60 & 2  \\
		
		Monk2    &  432 & 6 &  2 \\
				
		Balance  &  625 &  4&  3 \\
		
		Australian & 690 & 14 &  2 \\
				
		CNAE-9   & 1080 & 856 & 9  \\

		Digit-MultiF1  & 2000 & 240 & 10 \\

		Digit-MultiF2  & 2000 & 216 & 10 \\

		Titanic  & 2201 & 3  &  2 \\
		
		Magic   & 19,020  & 10 &  2 \\

		Motor  & 58,509 & 49 &  11\\

		Covertype  & 581,012 & 54 &  3\\

		SUSY  & 5,000,000 & 18 & 2 \\ \hline
		
	\end{tabular}
\end{table}
}

\begin{table*}[htbp]   
\centering \caption{ \label{result} The average test accuracy of deep hybrid, maxout, average and convolutional neural-kernel architectures on several benchmark datasets.}
\renewcommand*{\arraystretch}{1.4}
\setlength{\tabcolsep}{5pt}
\begin{tabular}{lllllllllll}
	
	
	&   &  \multicolumn{3}{c}{\textbf{Deep neural-kernel networks \cite{mehrkanoon2019deep}}}  \\ \cline{3-5}

	Dataset & \textbf{deep hybrid \cite{mehrkanoon2018deep}}  &   \textbf{Average} & \textbf{Maxout} &  \textbf{CNN}  \\  [0.5ex] \hline  

	Sonar &  $0.77\pm 0.04$   & $0.77\pm 0.02$ & $\underline{0.78 \pm 0.01}$ & $\underline{0.78\pm 0.01}$\\
	
	Monk2    &   $\underline{1.00\pm 0.00}$   & $\underline{1.00\pm 0.00}$ & $\underline{1.00 \pm 0.00}$ & $\underline{1.00\pm 0.00}$\\

	Balance  &   $0.97\pm 0.02$  &  $0.98\pm 0.01$ & $\underline{0.99 \pm 0.01}$ & $\underline{0.99\pm 0.01}$\\
	
	Australian &   $0.87\pm0.01$  & $0.88\pm 0.01$ & $\underline{0.90 \pm 0.01}$  &  $0.88\pm 0.02$ \\

	CNAE-9   &  $0.94\pm 0.02$  &   $0.94\pm 0.01$ & $\underline{0.95 \pm 0.01}$ & $0.94\pm 0.01$\\

	Digit-MultiF1  &  $\underline{0.98\pm 0.01}$  & $\underline{0.98\pm 0.01}$ & $\underline{0.98 \pm 0.02}$ &  $\underline{0.98\pm 0.01}$\\

	Digit-MultiF2  &  $0.97\pm 0.02$   &   $\underline{0.99\pm 0.01}$ & $\underline{0.99 \pm 0.02}$ & $\underline{0.99\pm 0.01}$\\

	Titanic  &   $0.78\pm 0.02$ &  $0.81\pm 0.01$ & $\underline{0.84 \pm 0.01}$ &  $0.83\pm 0.01$\\

	Magic   &   $0.86\pm 0.01$  &  $0.86\pm 0.01$ & $0.87 \pm 0.02$ &  $\underline{0.88\pm 0.01}$\\

	Motor  &  $0.96\pm 0.01$  &   $0.98\pm 0.01$ & $\underline{0.99 \pm 0.01}$ & $\underline{0.99\pm 0.01}$\\

	Covertype  &   $0.88\pm 0.02$  & $0.94\pm 0.01$ & $\underline{0.95 \pm 0.01}$ & $0.94\pm 0.01$\\

	SUSY & $0.81\pm 0.01$ &  $0.81\pm 0.02$ & $\underline{0.82 \pm 0.01}$ &  $0.81\pm 0.01$ 
	
	\\ \hline 
\end{tabular}
\end{table*}

\normalsize

%
%
%
%
%
%
%
%
%
%
%
%
%
%

\begin{figure*}[htbp]
\begin{center}
	
	\renewcommand*{\arraystretch}{0.5}
	\setlength{\tabcolsep}{2pt}
	
	\begin{tabular}{cc}
		

		\subfigure[]
		{
			\includegraphics[width=2in, height=1.8in]{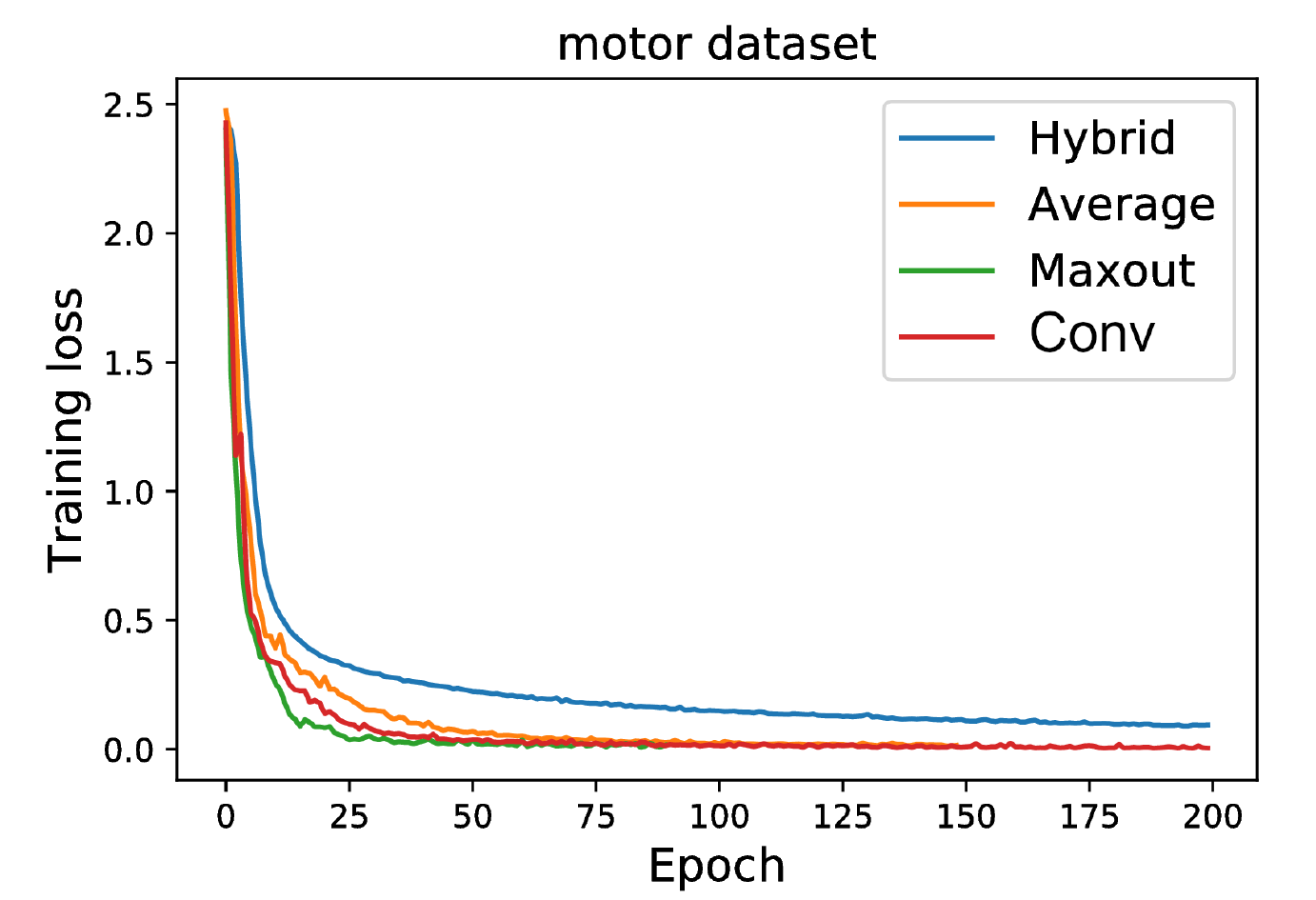}
		}
		
		&
		
		\subfigure[]
		{
			\includegraphics[width=2in, height=1.8in]{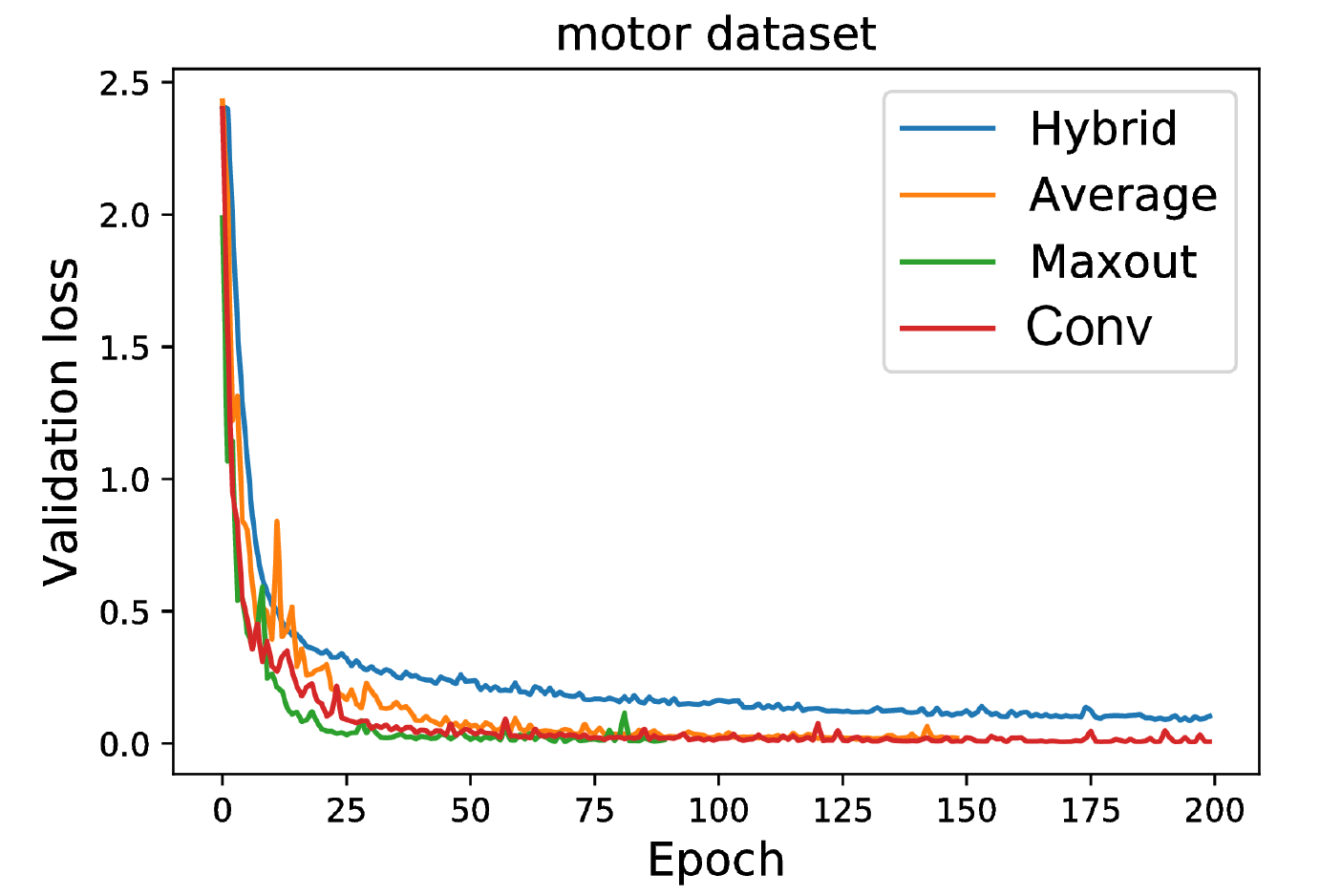}
		}
		
		\\
		
		\subfigure[]
		{
			\includegraphics[width=2in, height=1.8in]{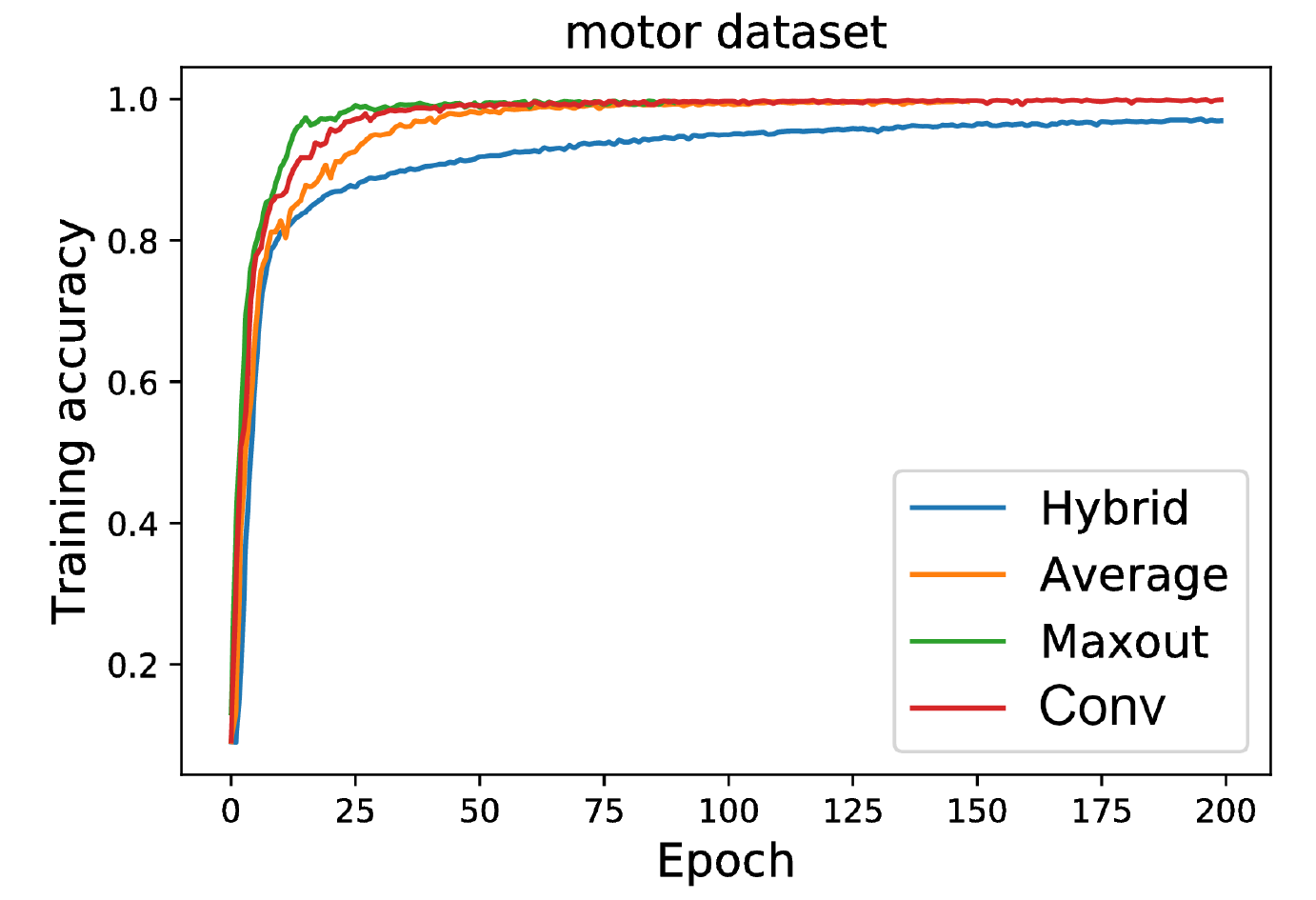}
		}

		&
		
		\subfigure[]
		{
			\includegraphics[width=2in, height=1.8in]{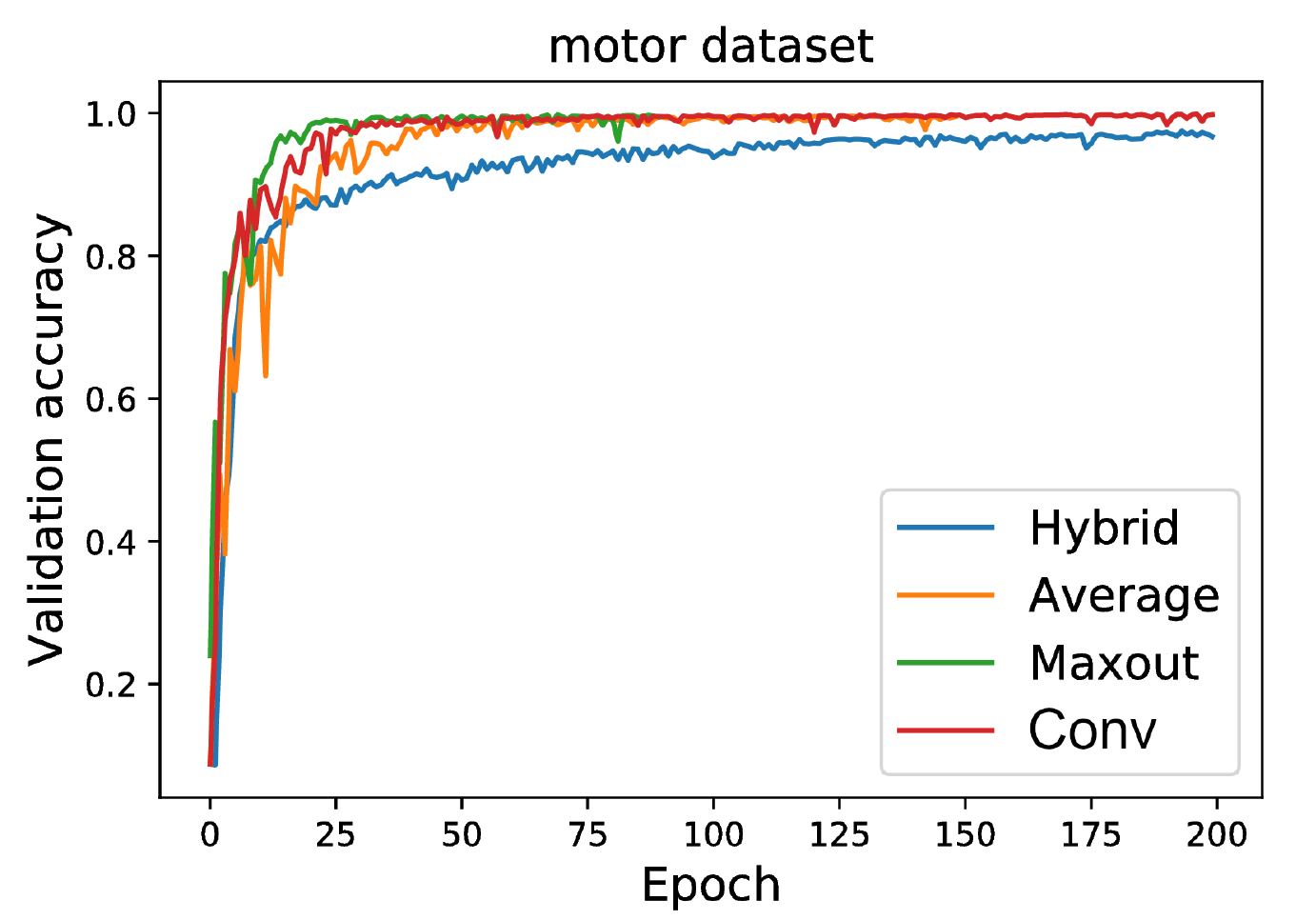}
		}
		
	\end{tabular}
\end{center}
\vspace{-0.1in} \caption{ \label{motor_val_acc_loss} The Motor dataset. (a,b,c,d) The training/validation loss and the accuracy of the deep neural-kernel networks discussed in section \ref{deep_nk}.}
\end{figure*}


\begin{figure*}[ht]
\begin{center}
	
	\renewcommand*{\arraystretch}{-10}
	\setlength{\tabcolsep}{-15pt}
	
	\begin{tabular}{cc}

		
		\subfigure[]
		{
			\includegraphics[width=2.5in, height=1.9in]{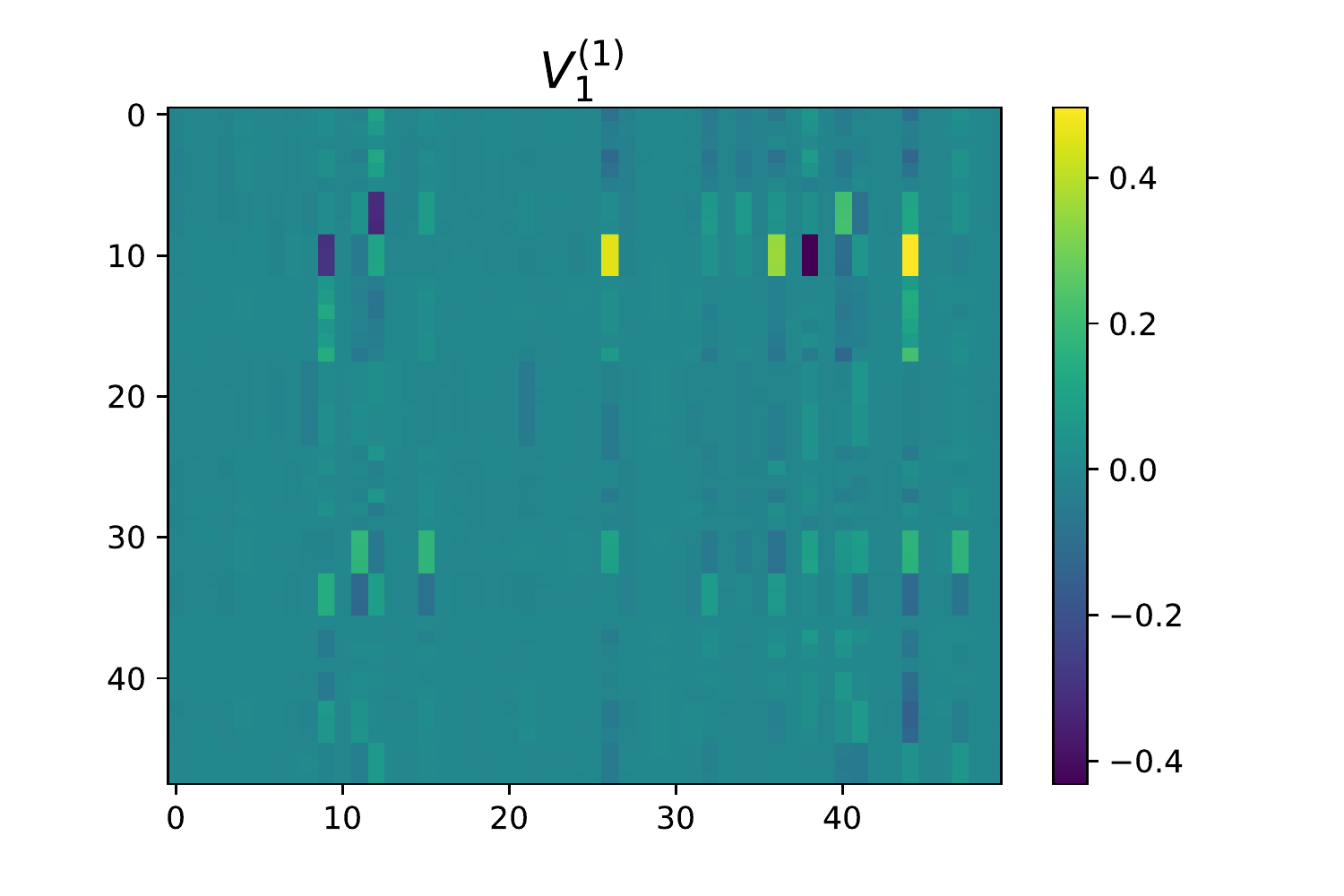}
		}
		
		&
		
		\subfigure[]
		{
			\includegraphics[width=2.5in, height=1.9in]{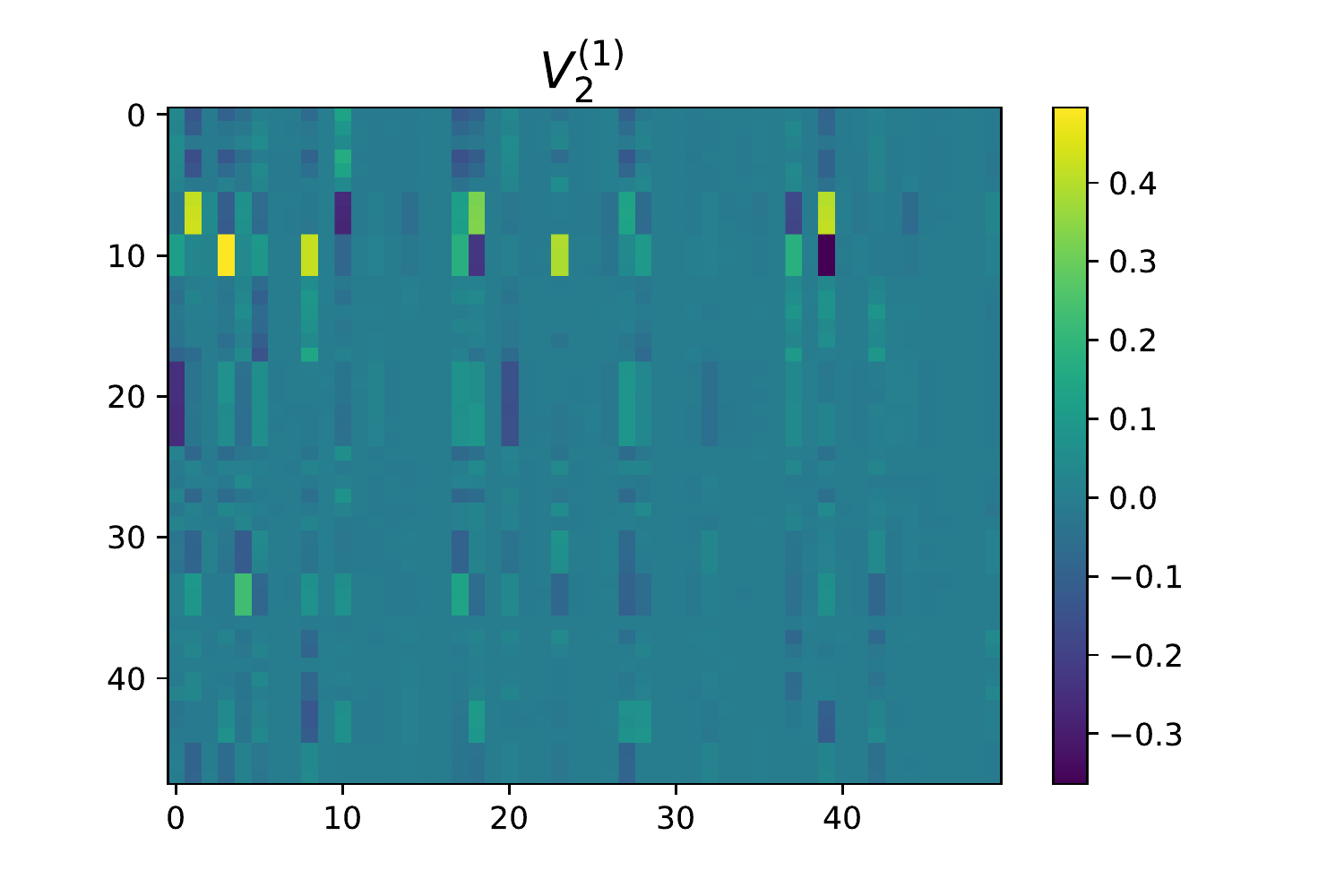}
		}

		\\
		
		\subfigure[]
		{
			\includegraphics[width=2.5in, height=1.9in]{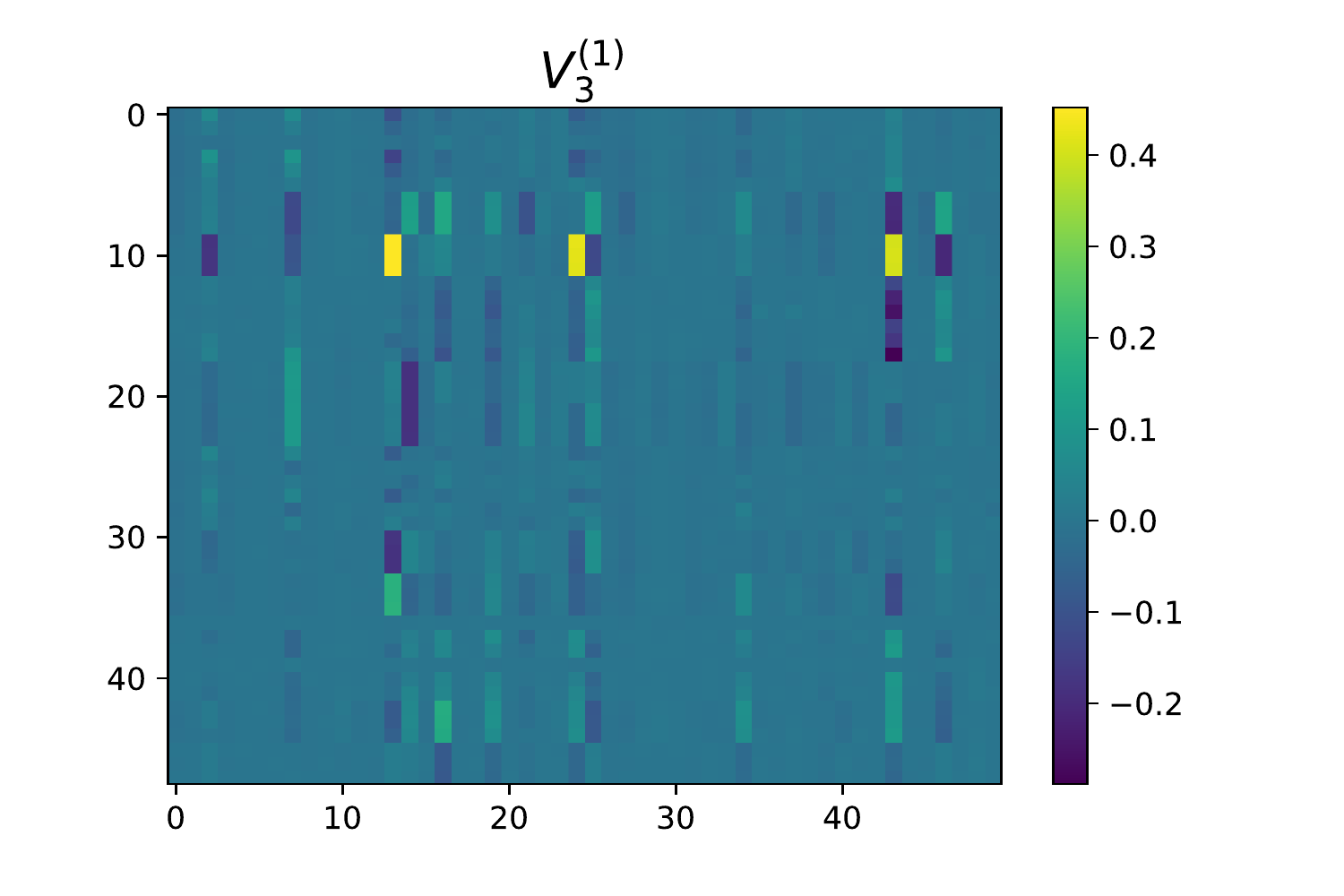}
		}
		
		&
		
		\subfigure[]
		{
			\includegraphics[width=2.5in, height=1.9in]{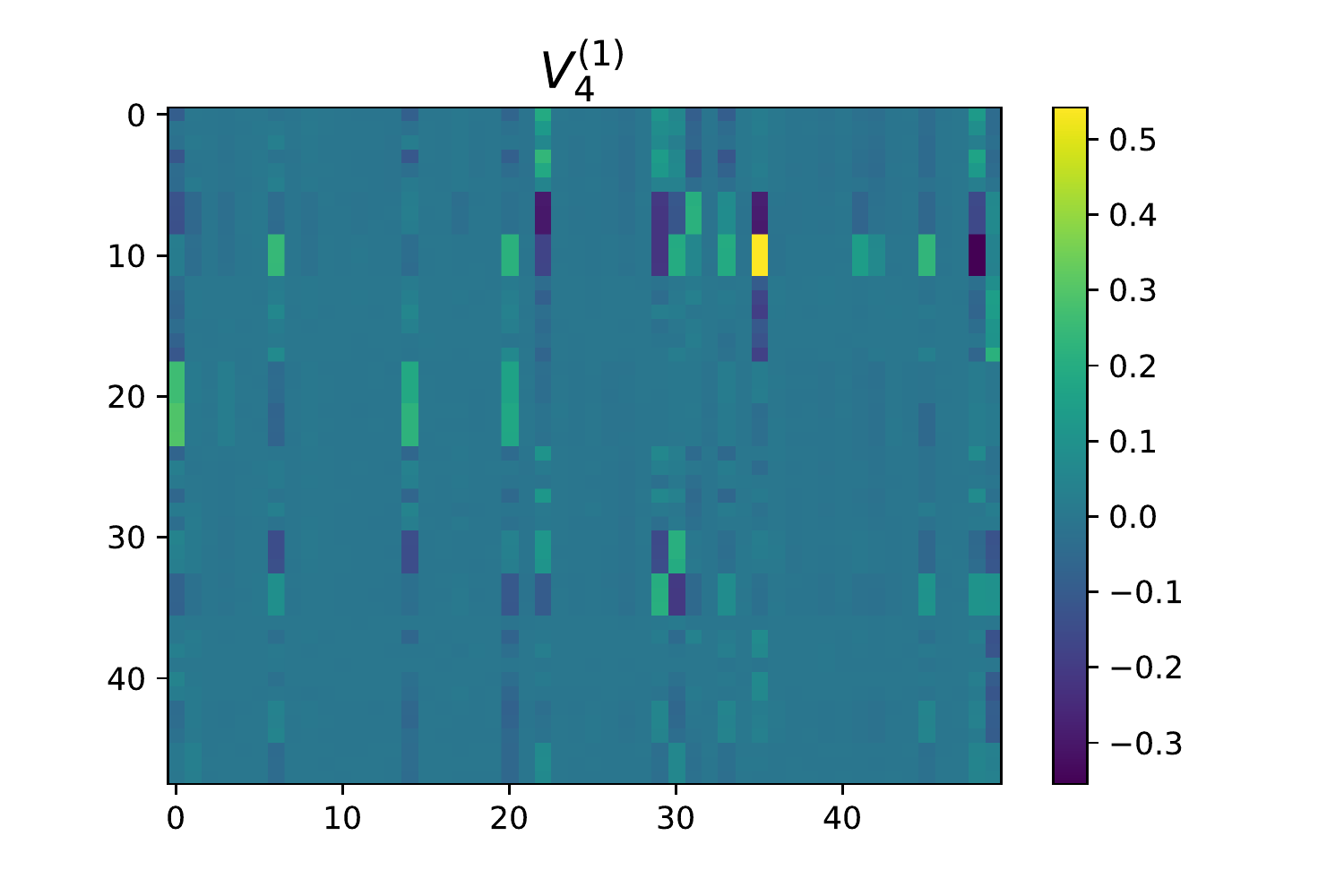}
		}

	\end{tabular}
\end{center}
\caption{ \label{motor_weights_1} The Motor dataset. (a,b,c,d) The learned weights of the first block of the maxout neural-kernel networks. (See Fig. \ref{Fig:maxout})}
\end{figure*}

\begin{figure*}[htbp]
\begin{center}

	\renewcommand*{\arraystretch}{-10}
	\setlength{\tabcolsep}{-15pt}
	
	\begin{tabular}{cc}

		\\  

		\subfigure[]
		{
			\includegraphics[width=2.5in, height=1.9in]{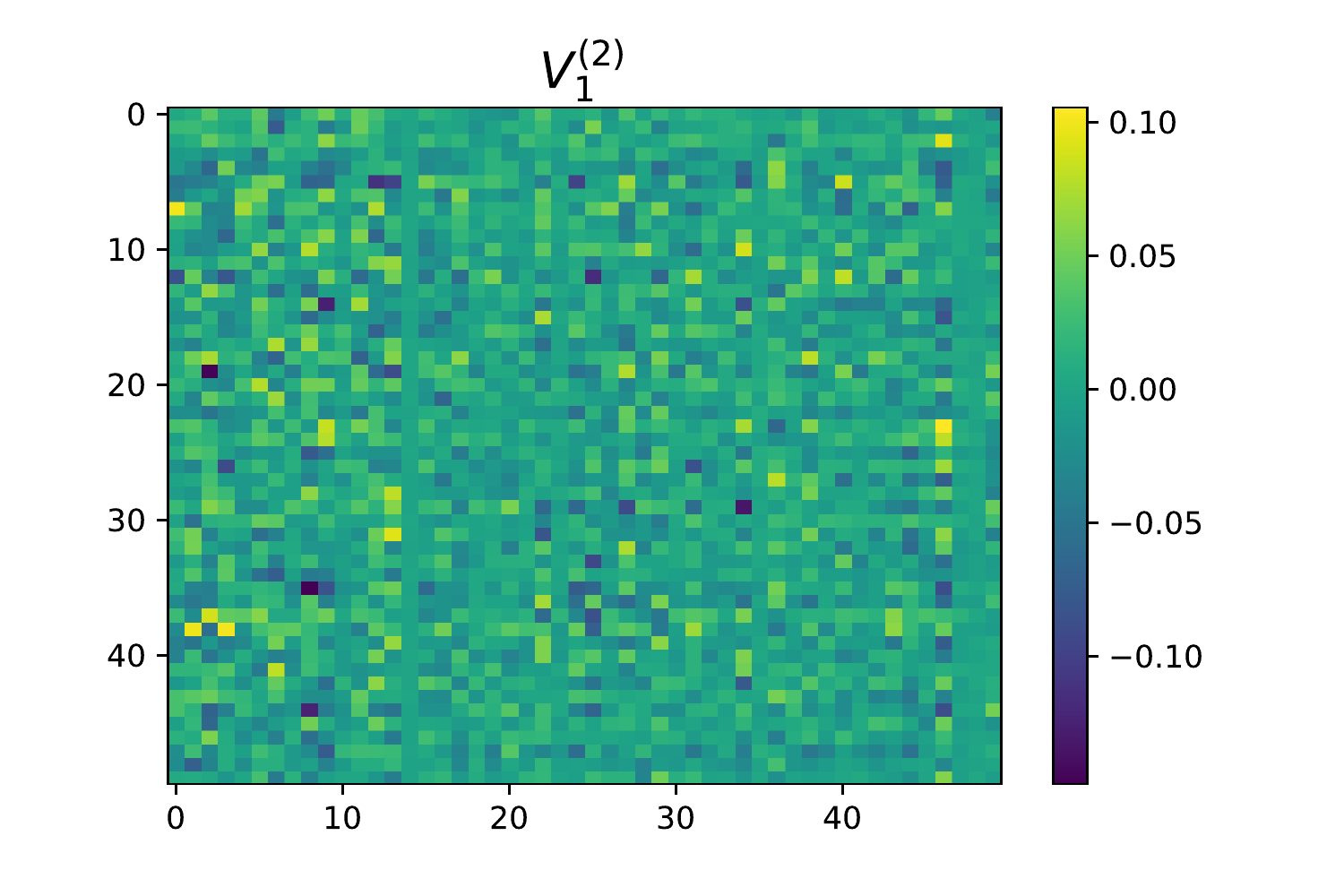}
		}
		
		&
		
		\subfigure[]
		{
			\includegraphics[width=2.5in, height=1.9in]{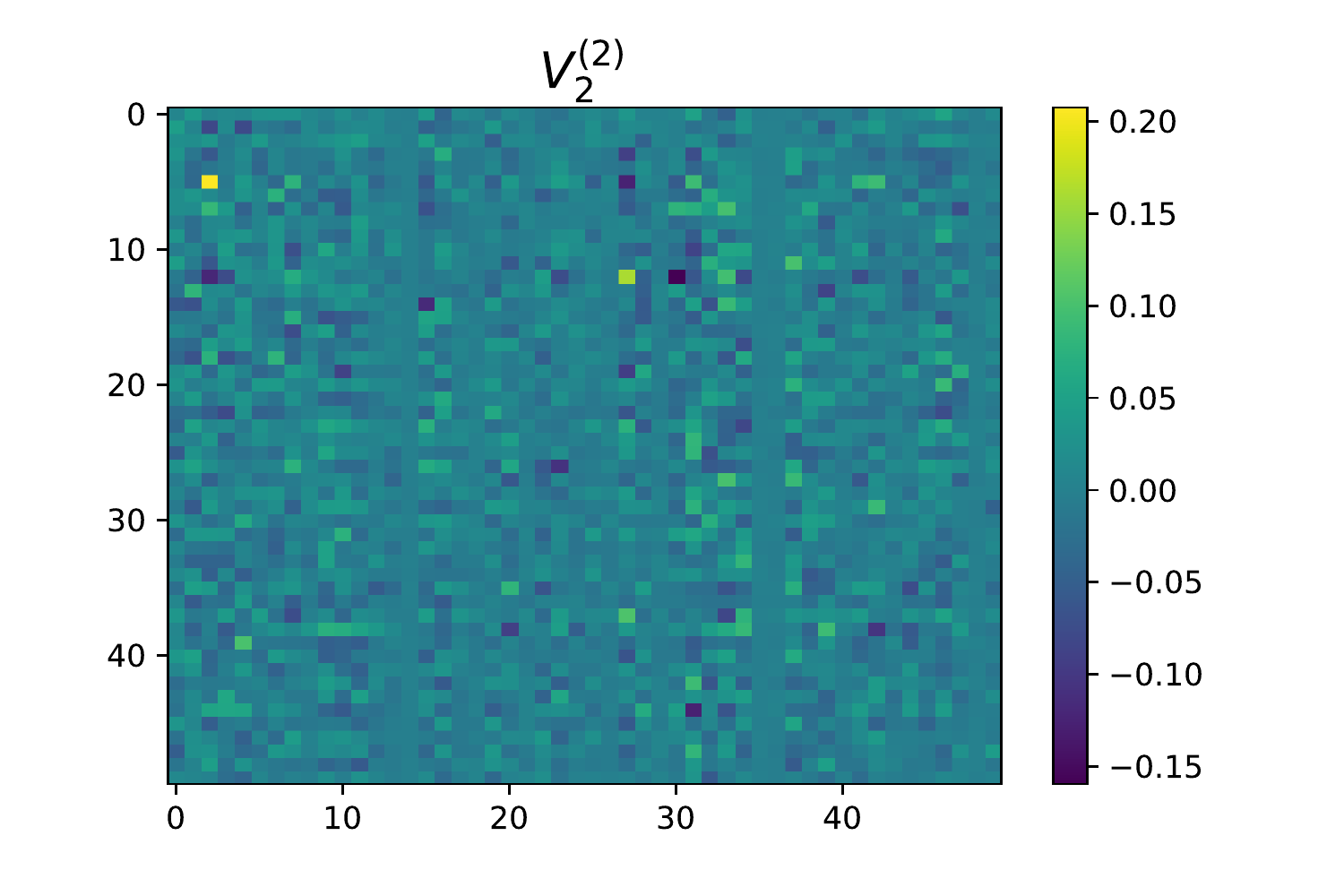}
		}

		\\
		
		\subfigure[]
		{
			\includegraphics[width=2.5in, height=1.9in]{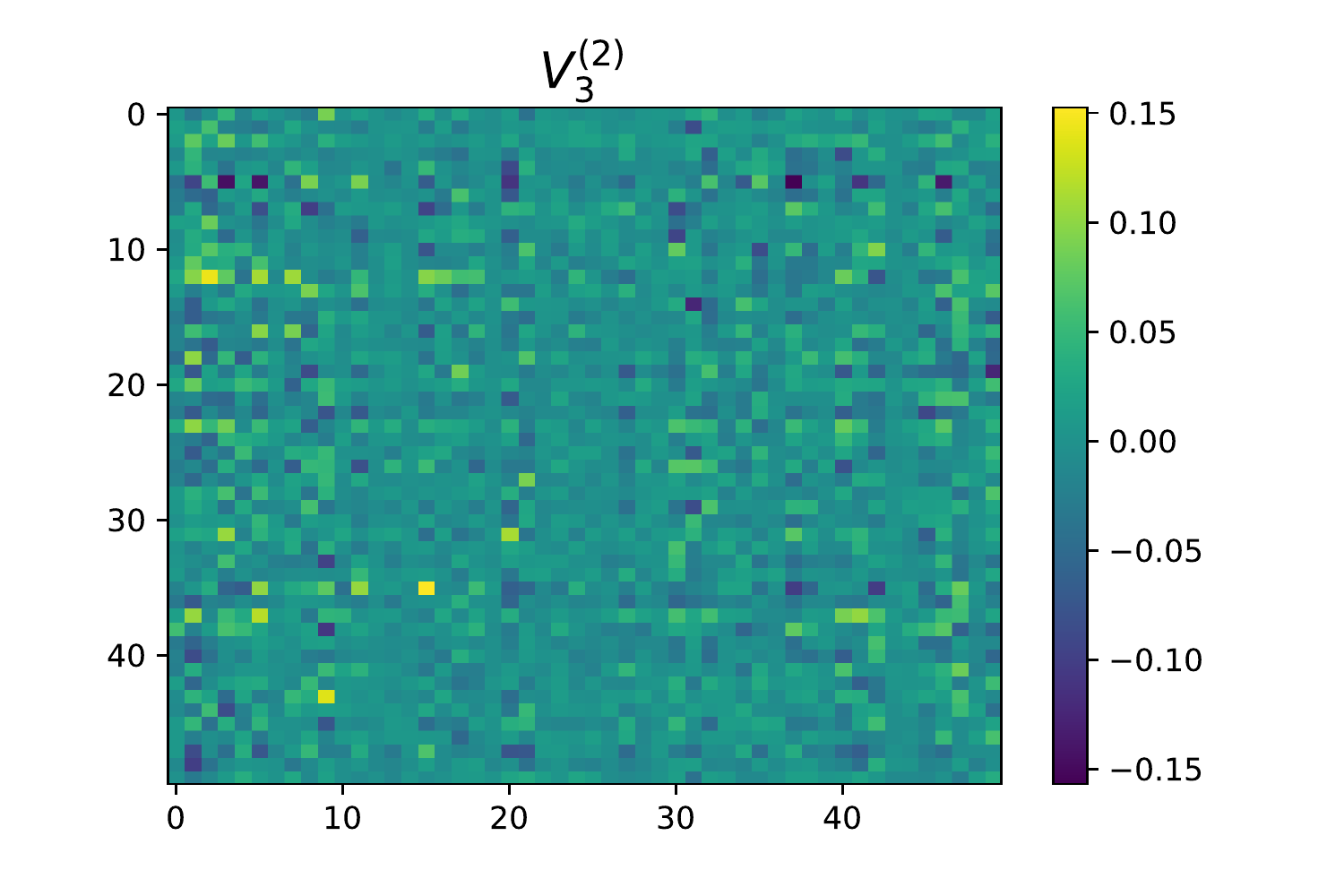}
		}
		
		&
		
		\subfigure[]
		{
			\includegraphics[width=2.5in, height=1.9in]{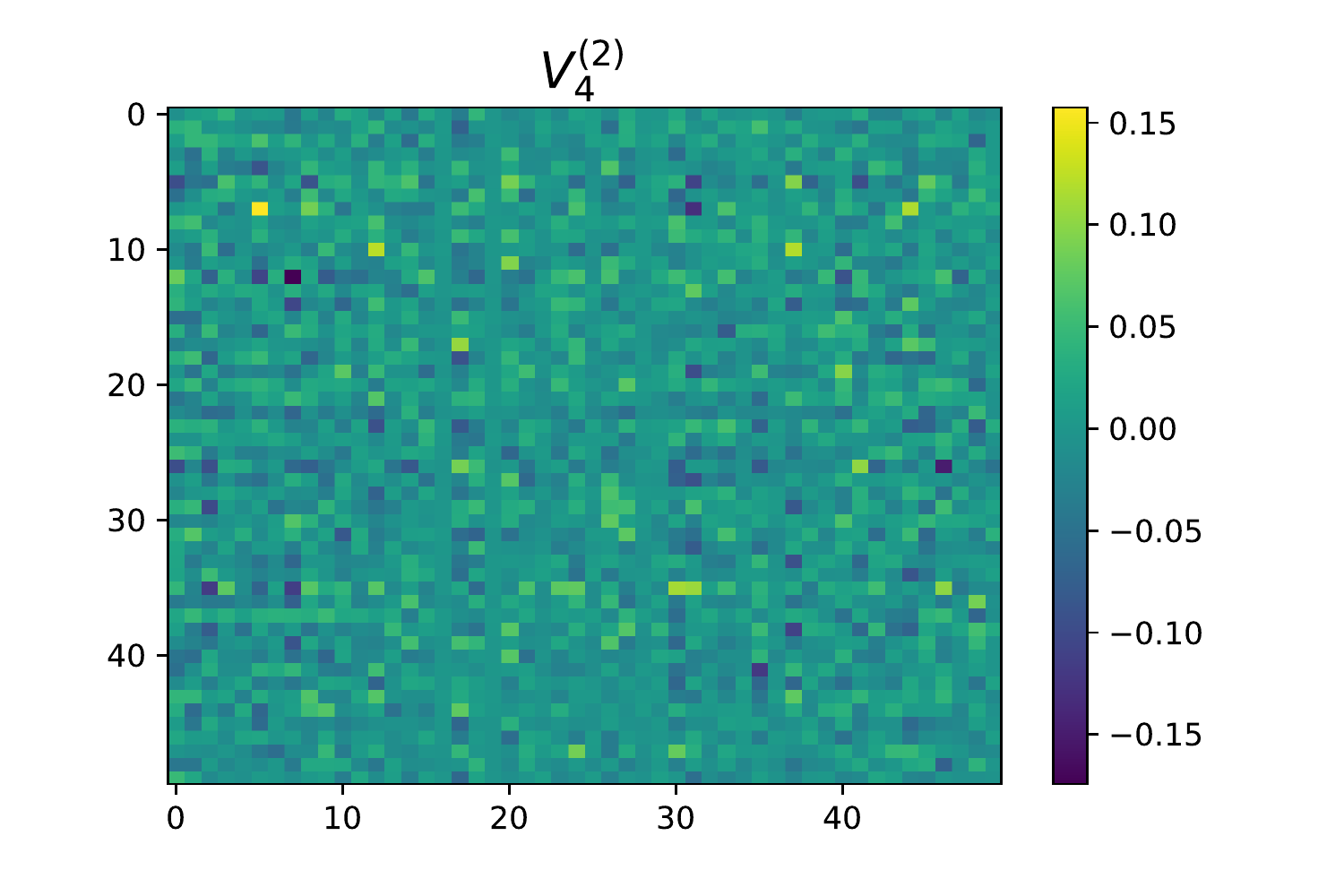}
		}

	\end{tabular}
\end{center}
\caption{ \label{motor_weights_2} The Motor dataset. (a,b,c,d) The learned weights of the second block of the maxout neural-kernel networks. (See Fig. \ref{Fig:maxout})}
\end{figure*}

\begin{figure*}[htbp]
\centering

\renewcommand*{\arraystretch}{-20}
\setlength{\tabcolsep}{2pt}

\begin{tabular}{cccc}


	\subfigure[]
	{
		\includegraphics[width=1.5in, height=1.2in]{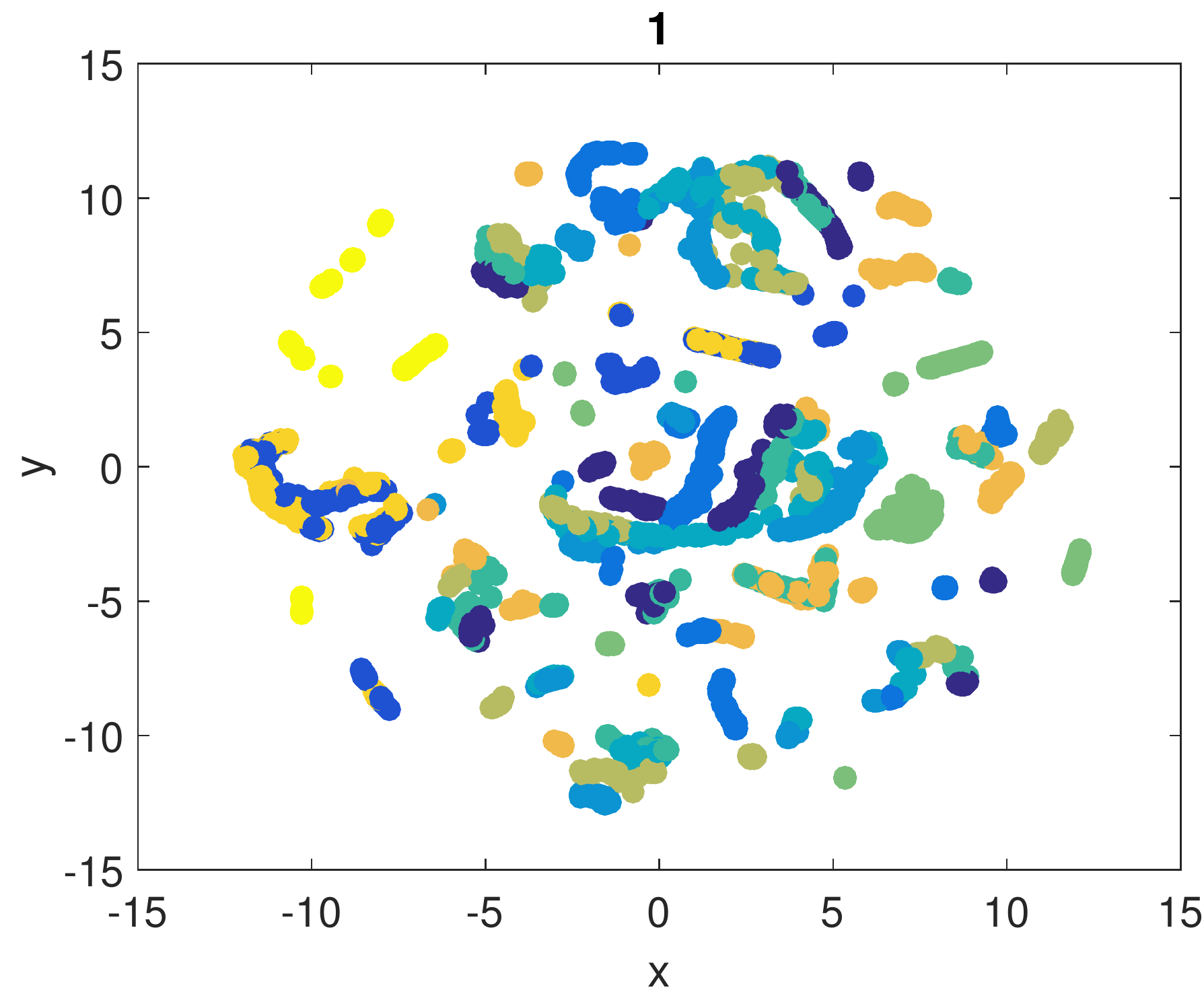}
		
	}
	
	&
	
	\subfigure[]
	{
		\includegraphics[width=1.5in, height=1.2in]{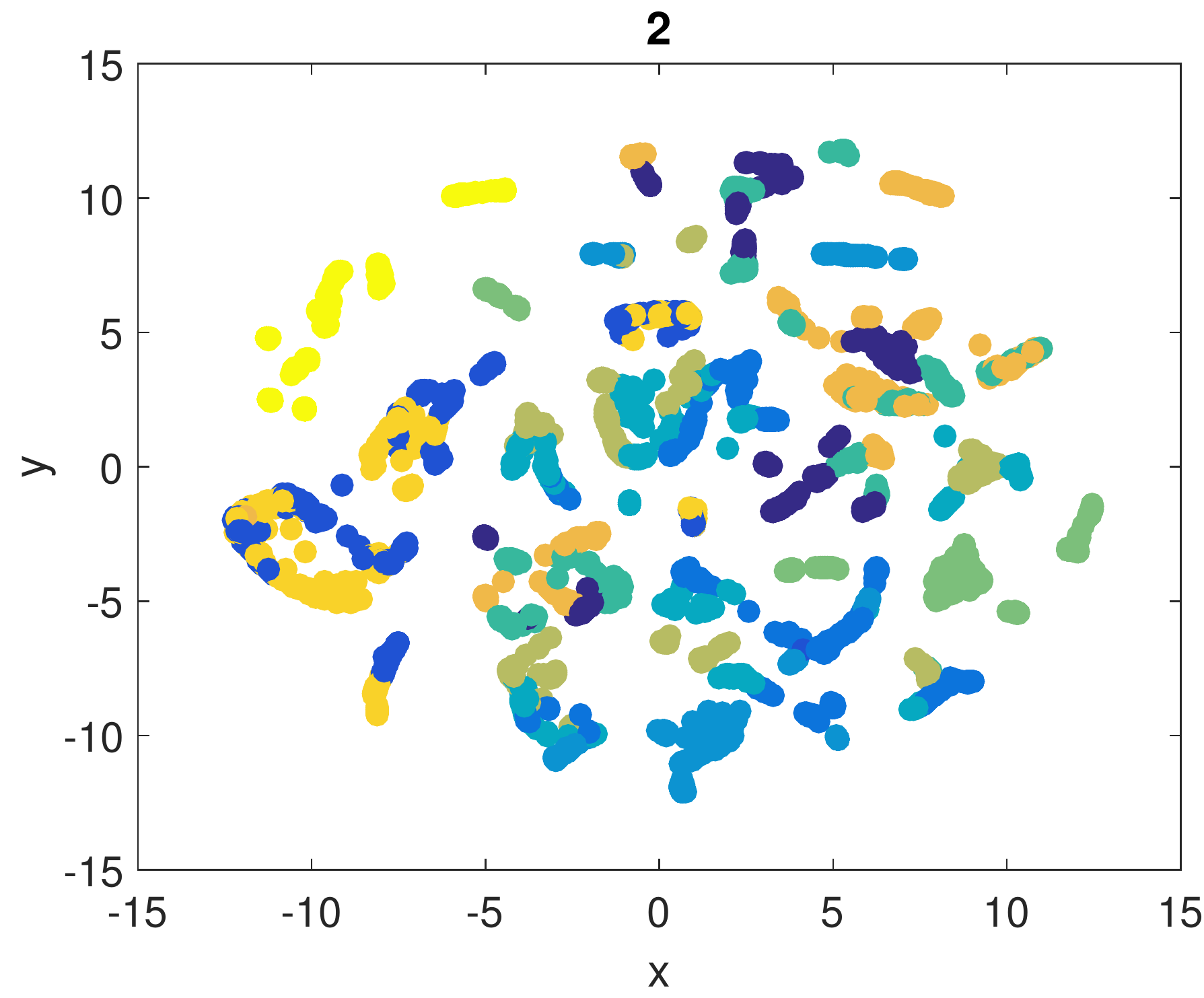}
	}
	
	&
	
	\subfigure[]
	{
		\includegraphics[width=1.5in, height=1.2in]{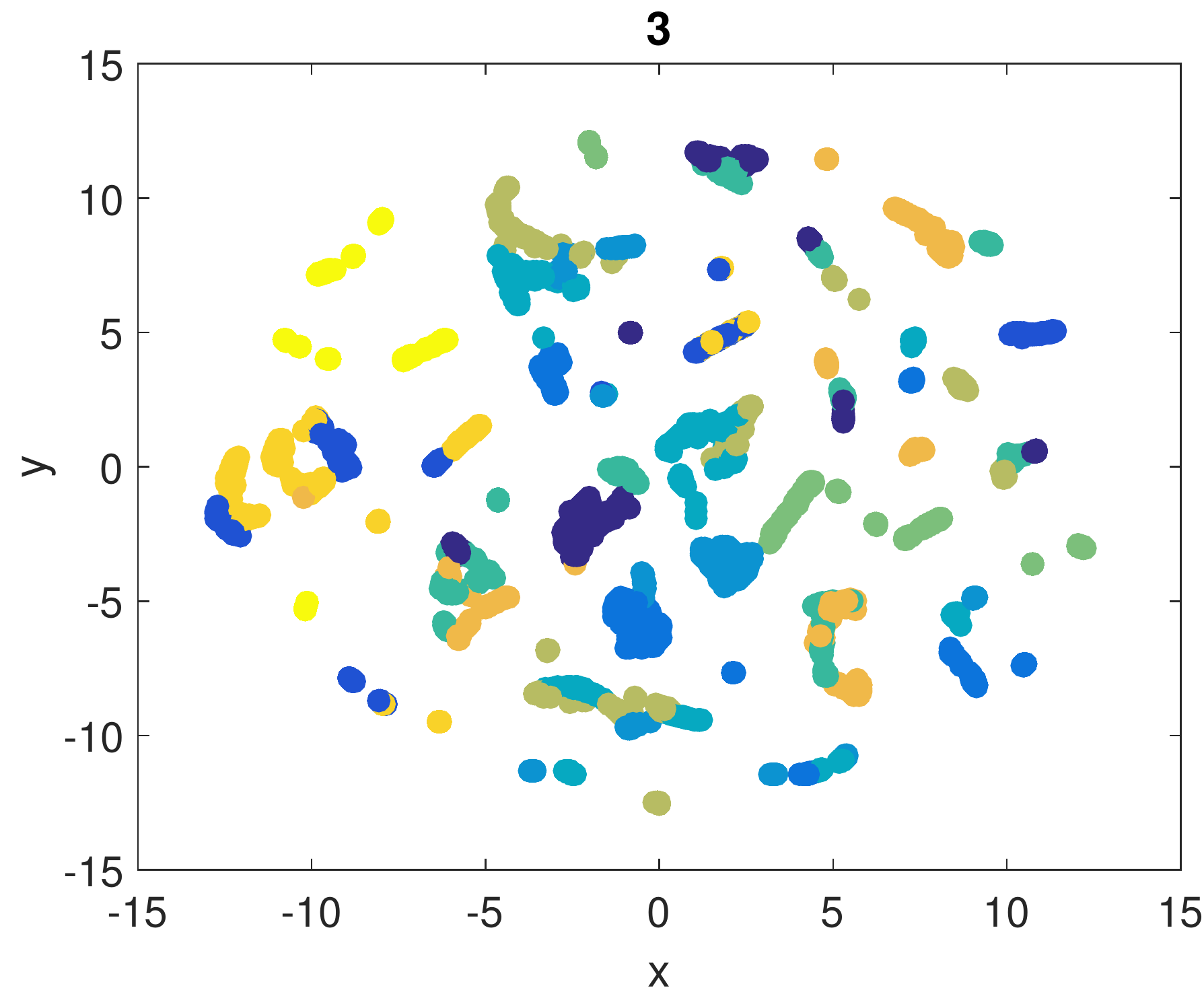}
	}

	\\
	
	\subfigure[]
	{
		\includegraphics[width=1.5in, height=1.2in]{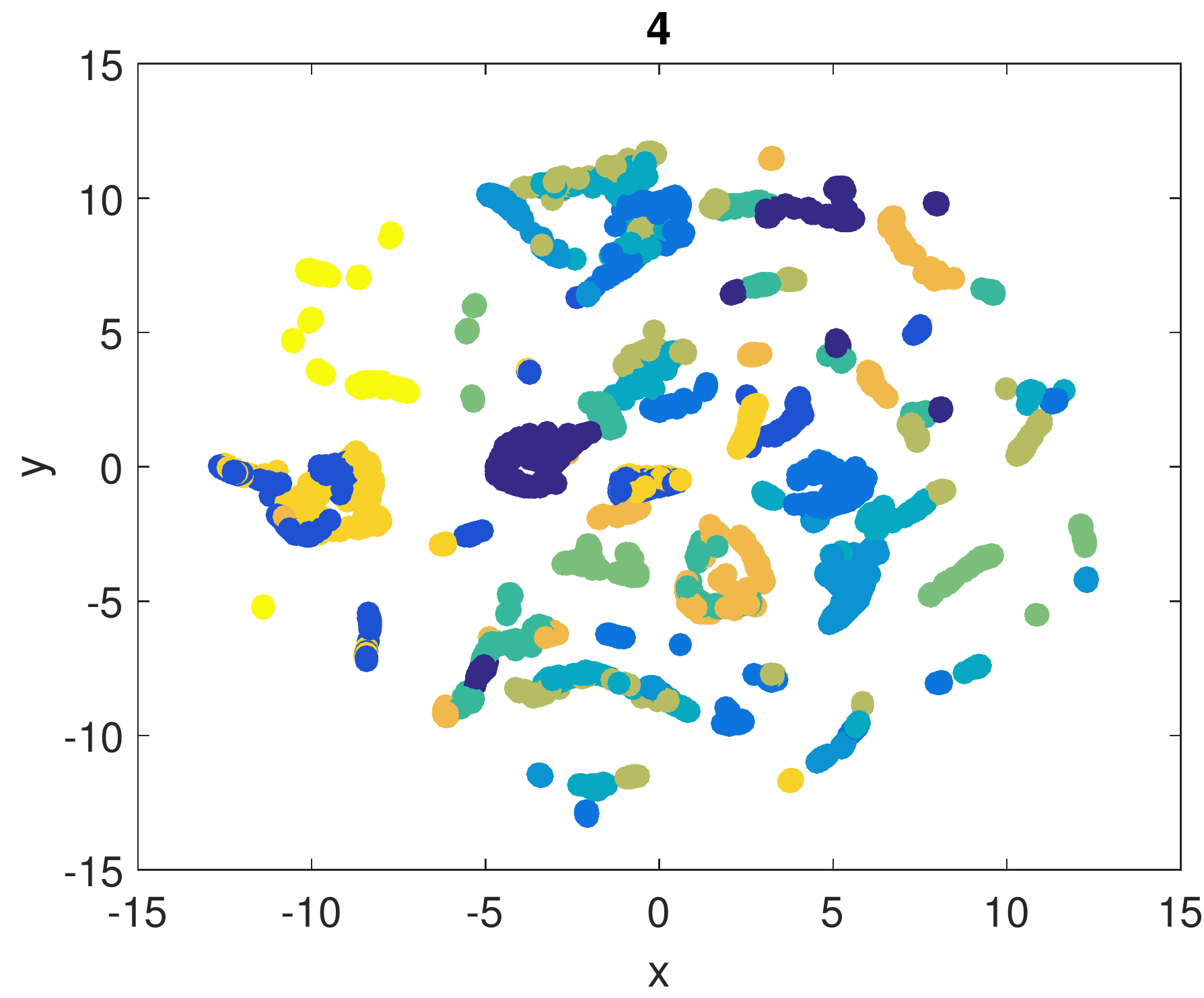}
	}
	
	&
	
	\subfigure[]
	{
		\includegraphics[width=1.5in, height=1.2in]{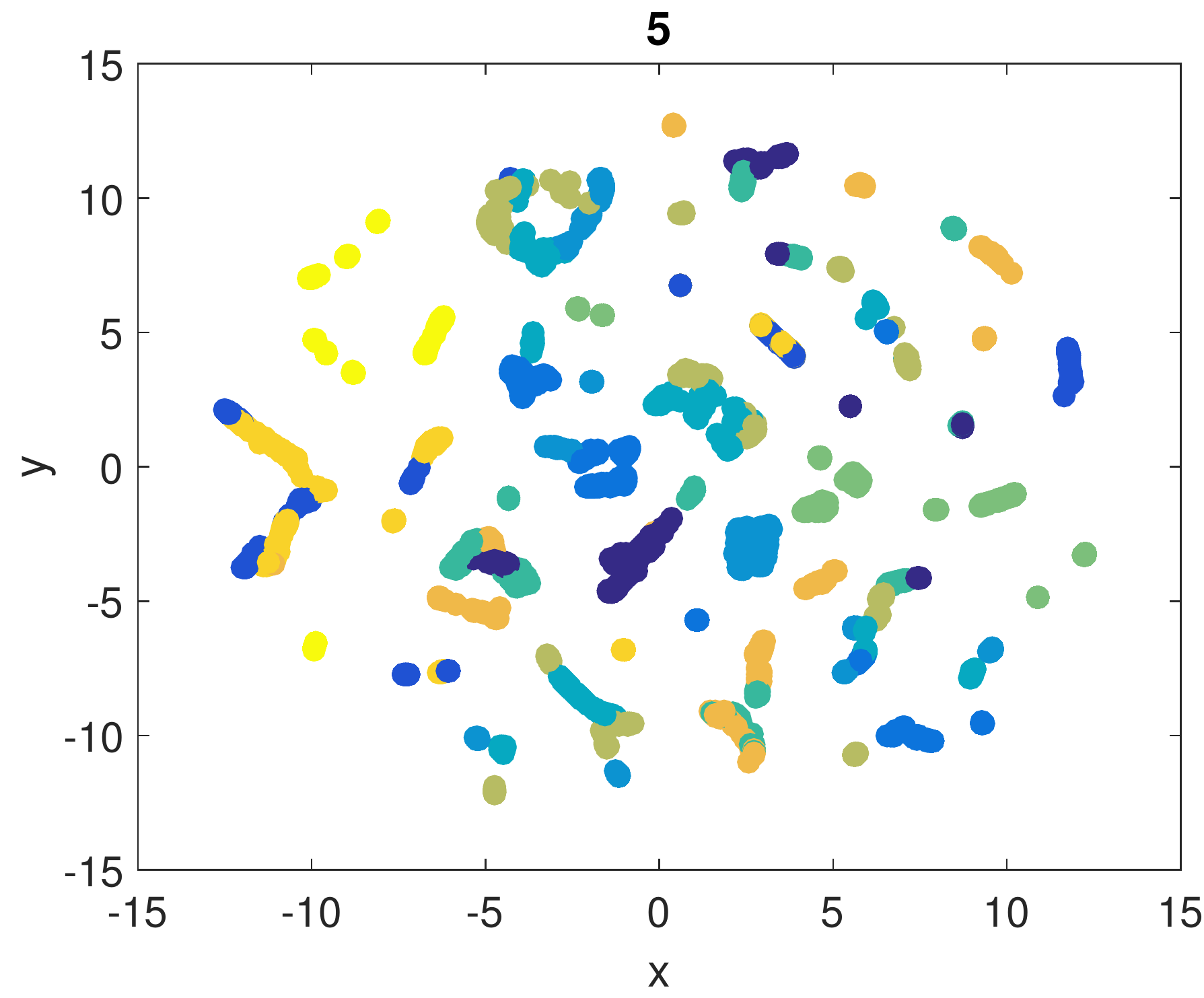}
	}

	&
	
	\subfigure[]
	{
		\includegraphics[width=1.5in, height=1.2in]{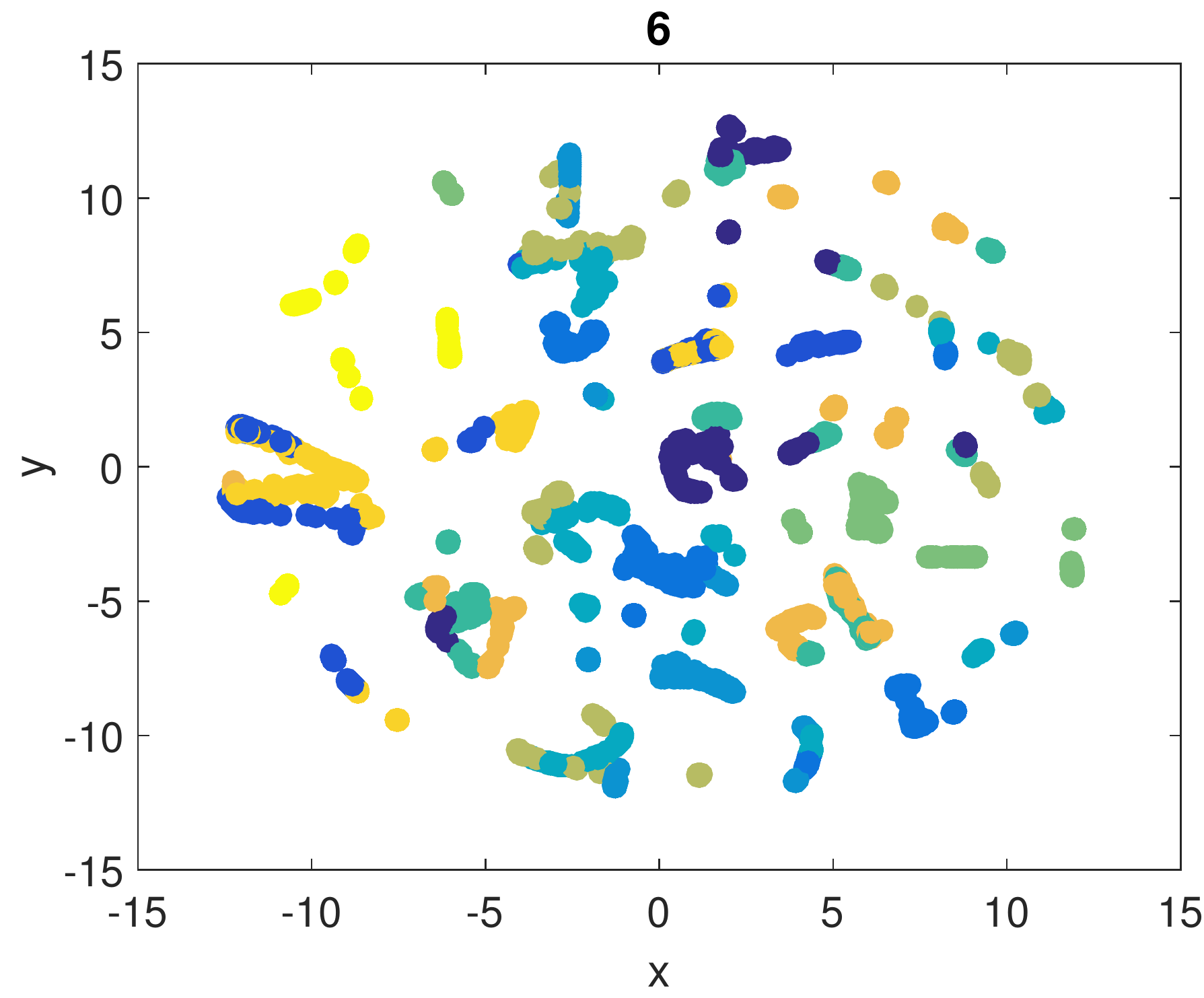}
	}

	\\
	
	\subfigure[]
	{
		\includegraphics[width=1.5in, height=1.2in]{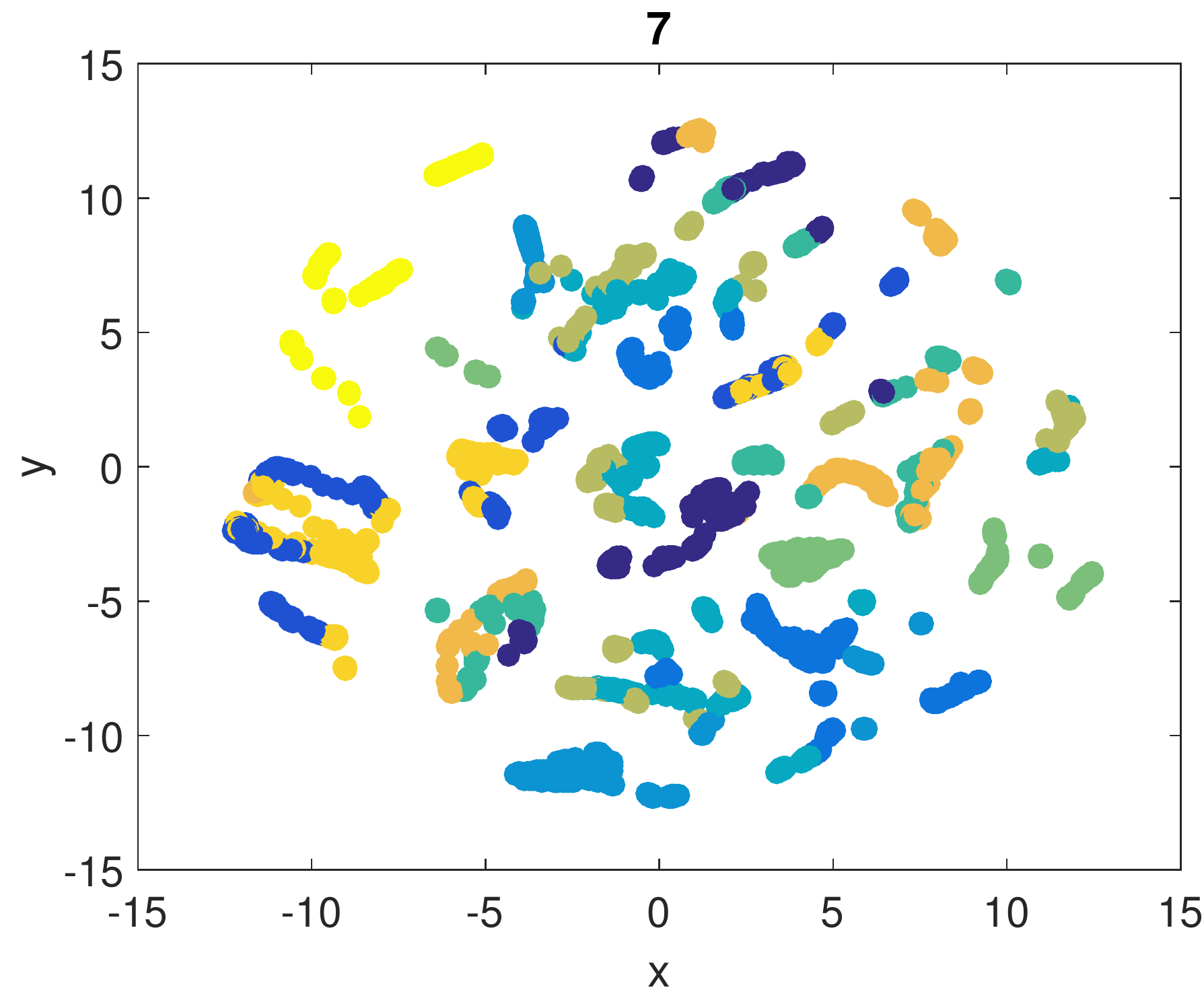}
	}
	
	&
	
	\subfigure[]
	{
		\includegraphics[width=1.5in, height=1.2in]{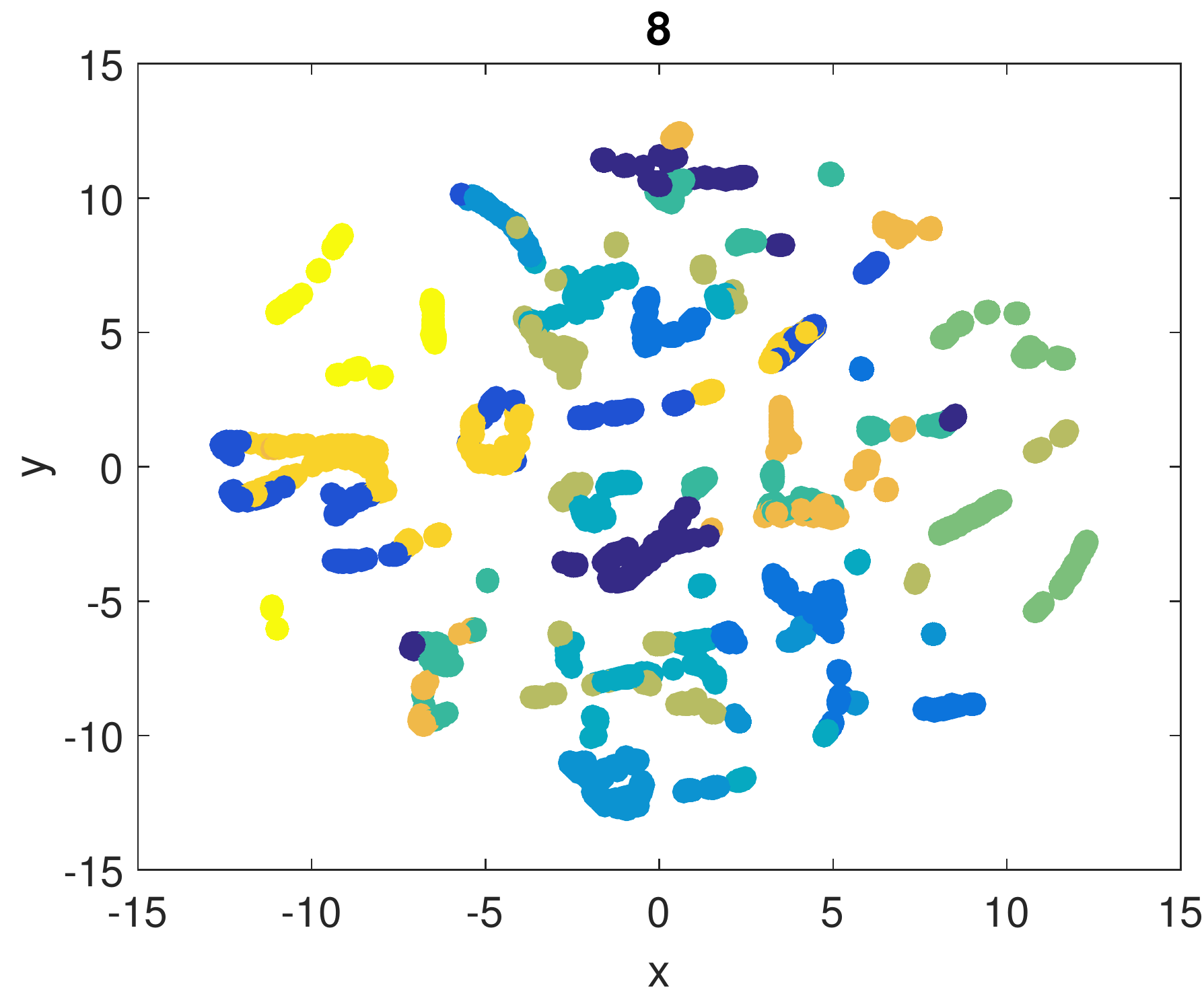}
	}

	&
	
	\subfigure[]
	{
		\includegraphics[width=1.5in, height=1.2in]{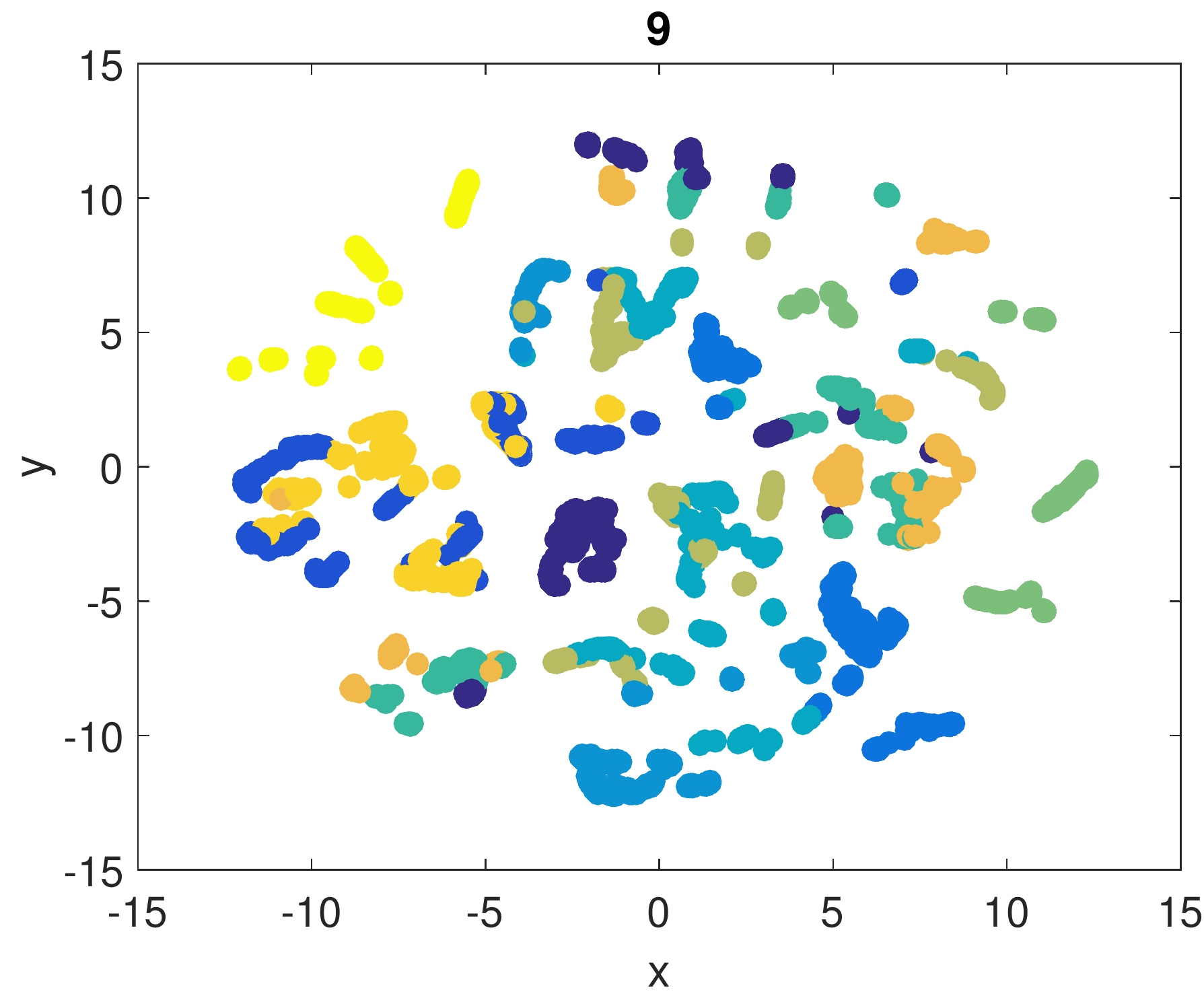}
	}

	\\
	
	\subfigure[]
	{
		\includegraphics[width=1.5in, height=1.2in]{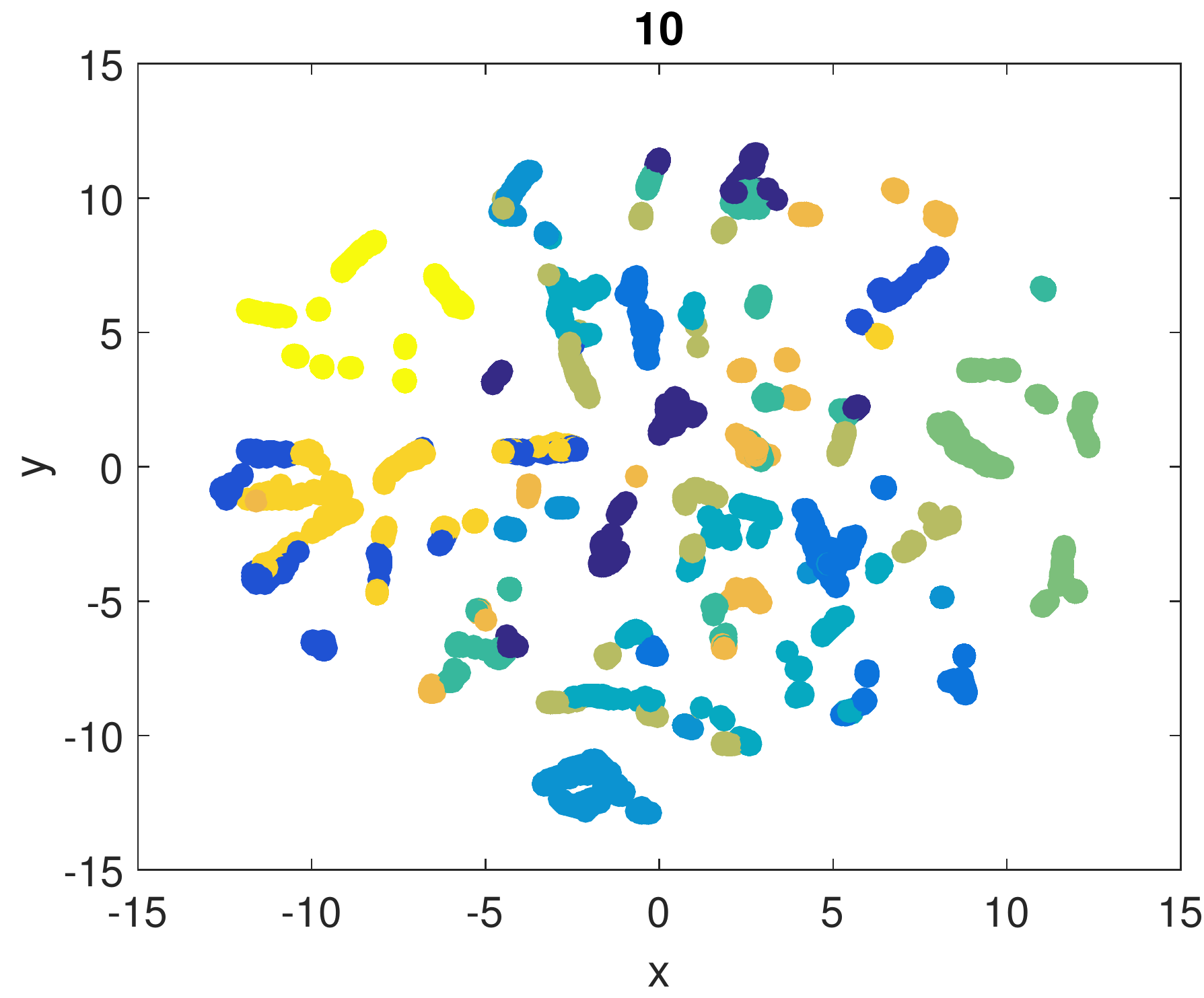}
	}

	&
	
	\subfigure[]
	{
		\includegraphics[width=1.5in, height=1.2in]{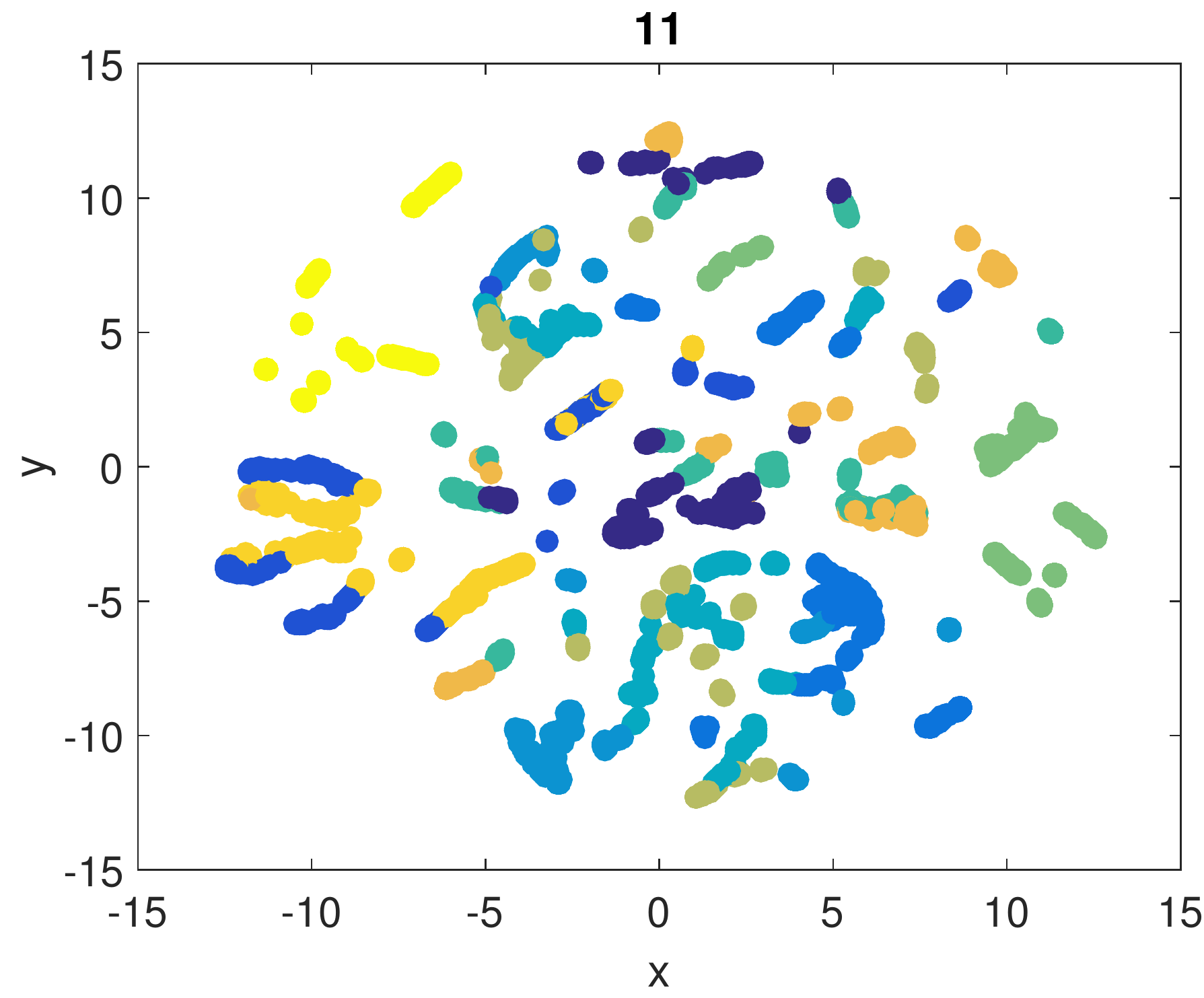}
	}
	
	&
	
	\subfigure[]
	{
		\includegraphics[width=1.5in, height=1.2in]{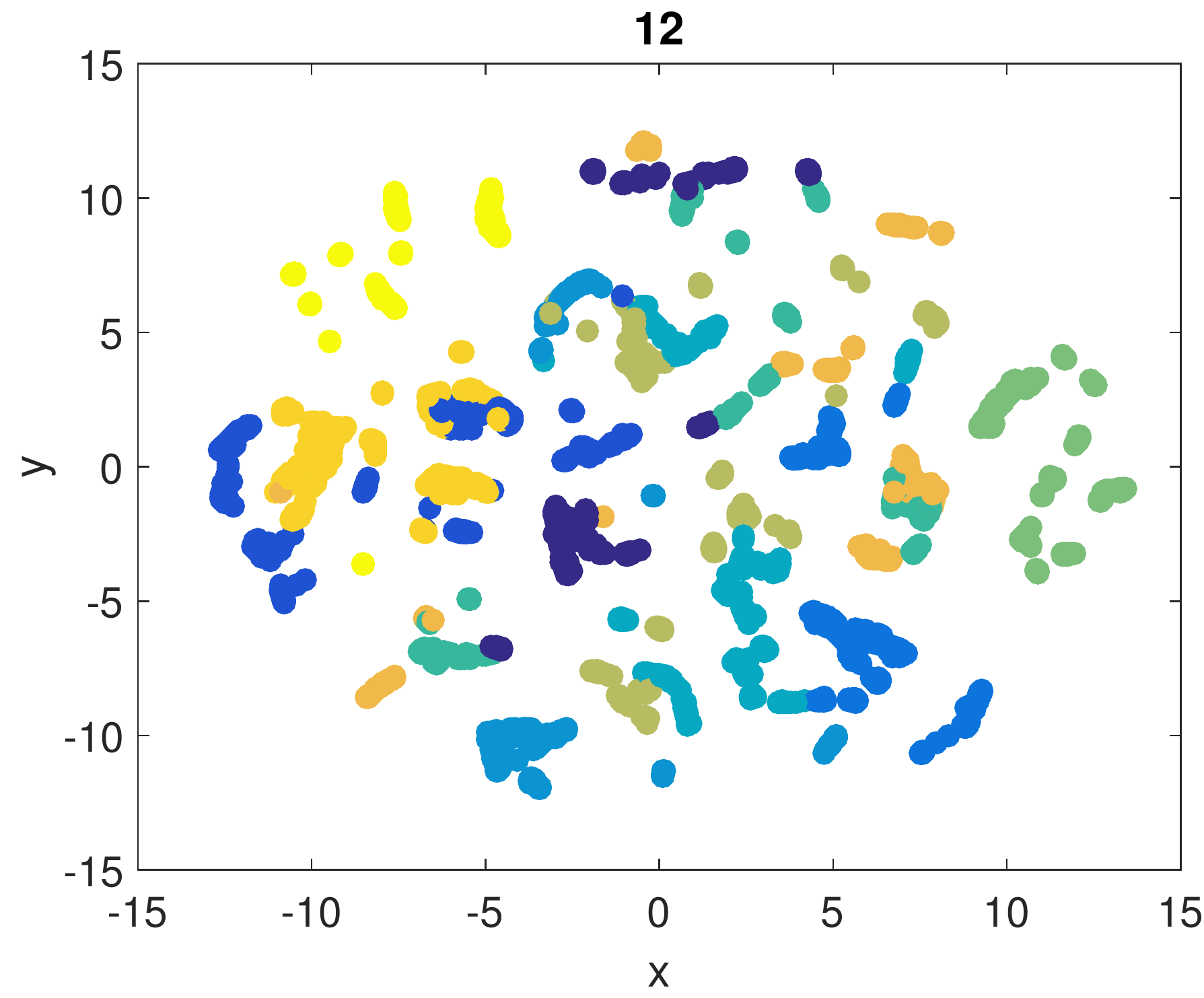}
	}

	\\
	
	\subfigure[]
	{
		\includegraphics[width=1.5in, height=1.2in]{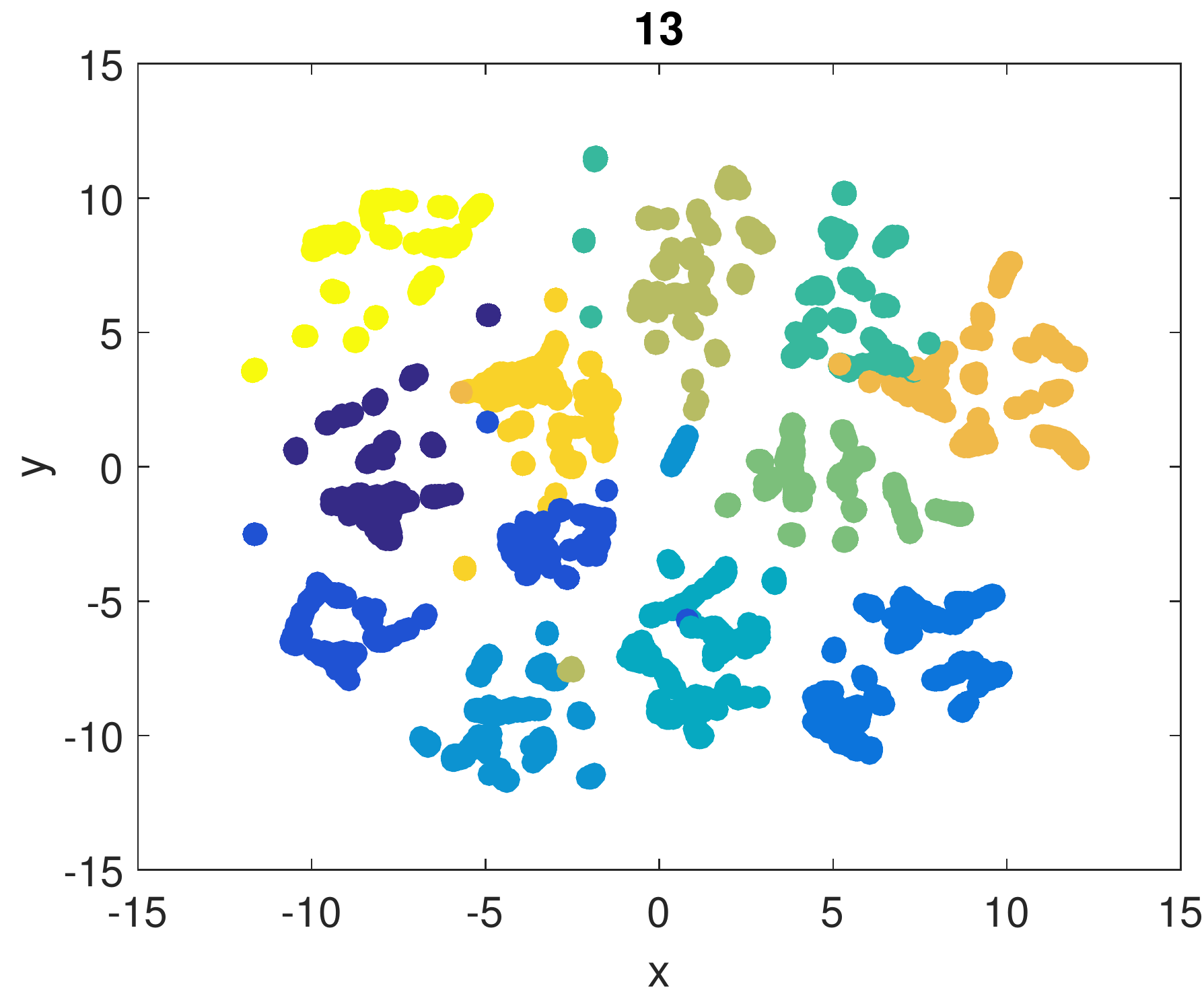}
		
	}

	&
	
	\subfigure[]
	{
		\includegraphics[width=1.5in, height=1.2in]{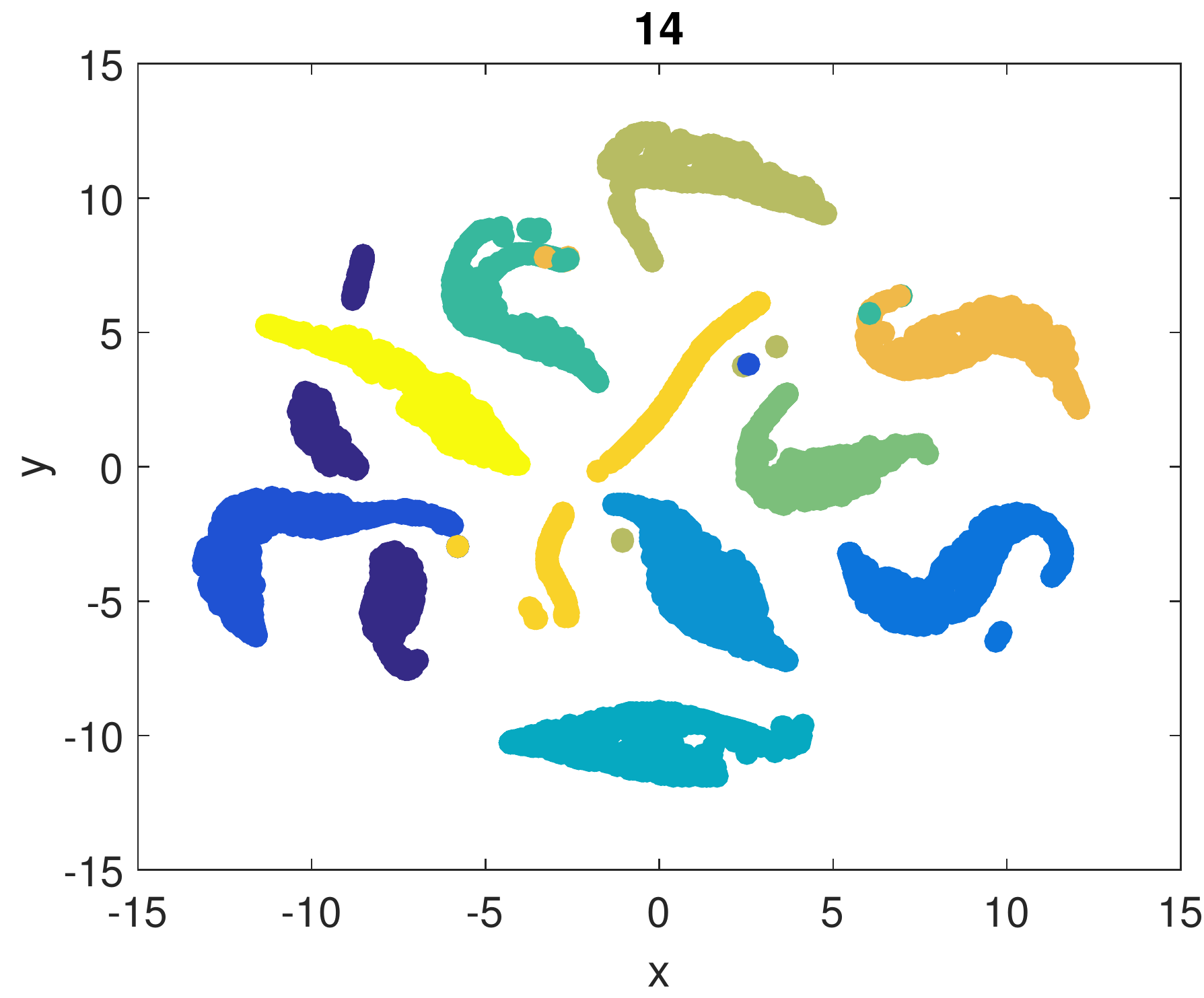}
	}
	
\end{tabular}
\caption{ \label{Motor} The Motor dataset. The t-SNE projections of: (a) the test data. (b,c,d,e) the linear transformations of the first block. (f) the maxout layer of the first block. (g) the explicit feature mapping of the first block. (h,i,j,k) the linear transformations of the second block. (l) the maxout layer of the second block. (m) the explicit feature mapping of the second block. (n) the score variables $S(x)$.}
\end{figure*}

The performance of the discussed deep maxout and convolutional neural-kernel networks are also evaluated on four real-life benchmark datasets CIFAR-10 \cite{krizhevsky2009learning}, CIFAR-100 \cite{krizhevsky2009learning}, SVHN \cite{netzer2011reading} and MNIST \cite{bengio2013representation}. The architectures used for these datasets follow a combination of convolutional neural networks (CNN) for feature learning and the proposed models for the classification task. The employed CNN networks consists of three convolutional layers and then downsampling the feature maps by a factor of two using spatial pooling layer. The results of the CNN model is fed as input to the deep neural-kernel networks. The parameters of the two models are jointly learned in an end-to-end fashion. In this experiment the neural-kernel networks is composed of  one block with three linear transformations. The number of filters of CNN is gradually increased i.e. 32, 64 and 128. 
The data statistics of the used images can be found in Table \ref{dataset_2}.  Next the pre-processing steps applied to each of the image dataset will be explained. The MNIST images are scaled to [0,1] before they are fed into the networks. The global contrast normalization as well as ZCA whitening \cite{goodfellow2013maxout} are applied for preprocessing the CIFAR-10 images. The weight decay and the size of the local receptive field are tuned using the validation set. The implemented settings for CIFAR-100 follow those applied on CIFAR-10 dataset. The task of SVHN dataset is to classify the digit in the centre of each image. Following the lines of Goodfellow et al. \cite{goodfellow2013maxout} here a local contrast normalization is applied in order to preprocess the data.

\normalsize{
\begin{table}[htbp]%
	\centering
	\caption{ \label{dataset_2} Dataset statistics for Images}
	\renewcommand*{\arraystretch}{1.4}
	\setlength{\tabcolsep}{5pt}
	\begin {tabular}{lrrrr}
	
	Dataset  & \# Image-size  & \# Training & \# Test  & \# Classes   \\   \hline  
	
	MNIST &  $28 \times 28$ & 60,000 & 10,000 & 10 \\
	
	CIFAR-10   &   $32 \times 32$ & 50,000 & 10,000 & 10 \\

	CIFAR-100   &   $32 \times 32$ & 50,000 & 10,000 & 100 \\
	
	SVHN  &   $32 \times 32$ & 73,257 & 26,032 & 10 \\ \hline 
	
	
\end{tabular}
\end{table}
}

\begin{figure}[htbp]
\centering
\includegraphics[width=3.5in, height=2.5in]{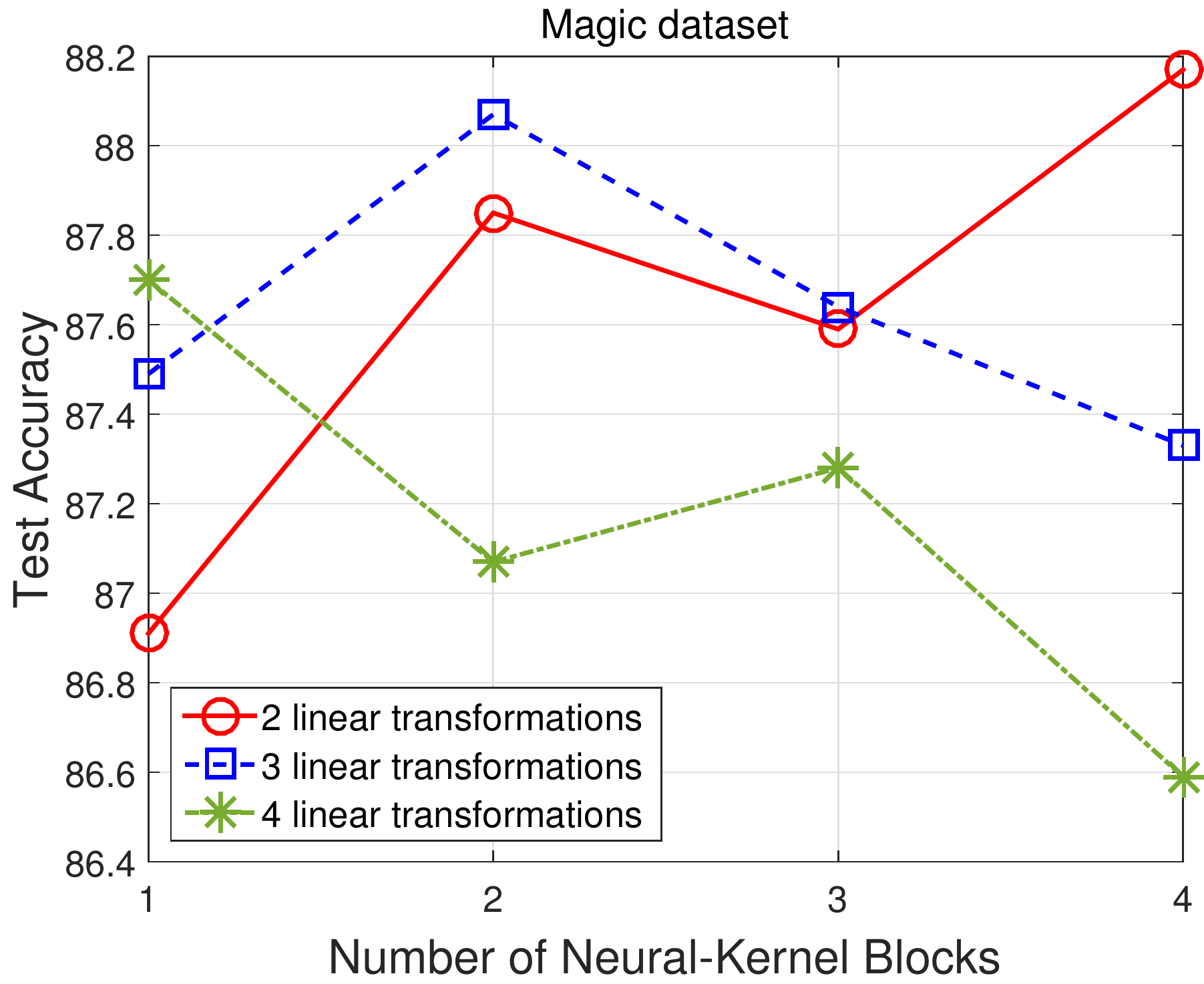} \\	
\caption{The test accuracy of the deep maxout neural-kernel networks corresponding to different pairs of the number of blocks and linear transformations for the Magic datasets.}\label{test_acc_blocks}
\end{figure}

\begin{figure}[t]
\begin{center}

\renewcommand*{\arraystretch}{1.0}
\setlength{\tabcolsep}{2pt}

\begin{tabular}{ccc}

	
	\subfigure[]
	{
		\includegraphics[width=1.5in, height=1.5in]{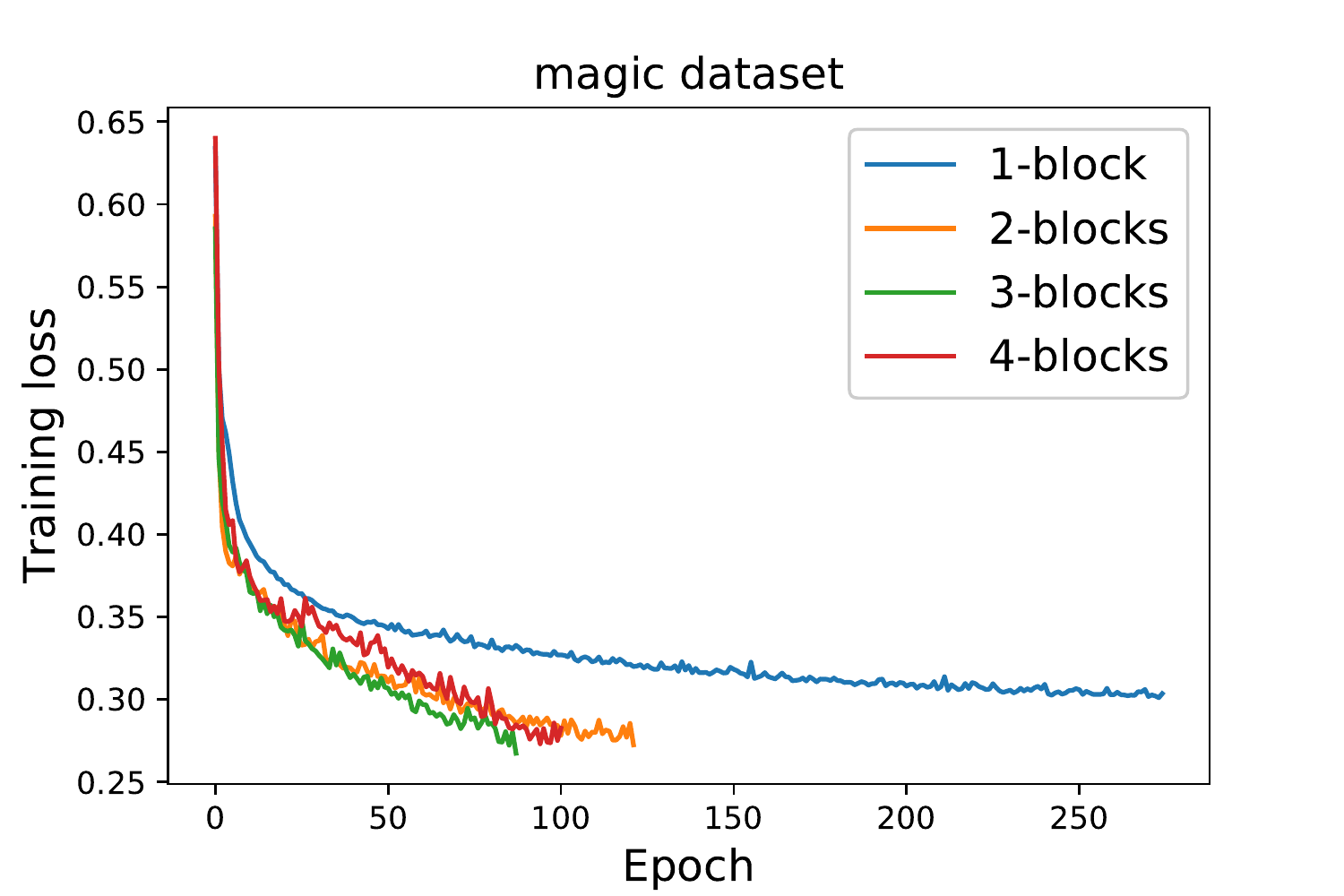}
	}
	
	&
	
	{
		\includegraphics[width=1.5in, height=1.5in]{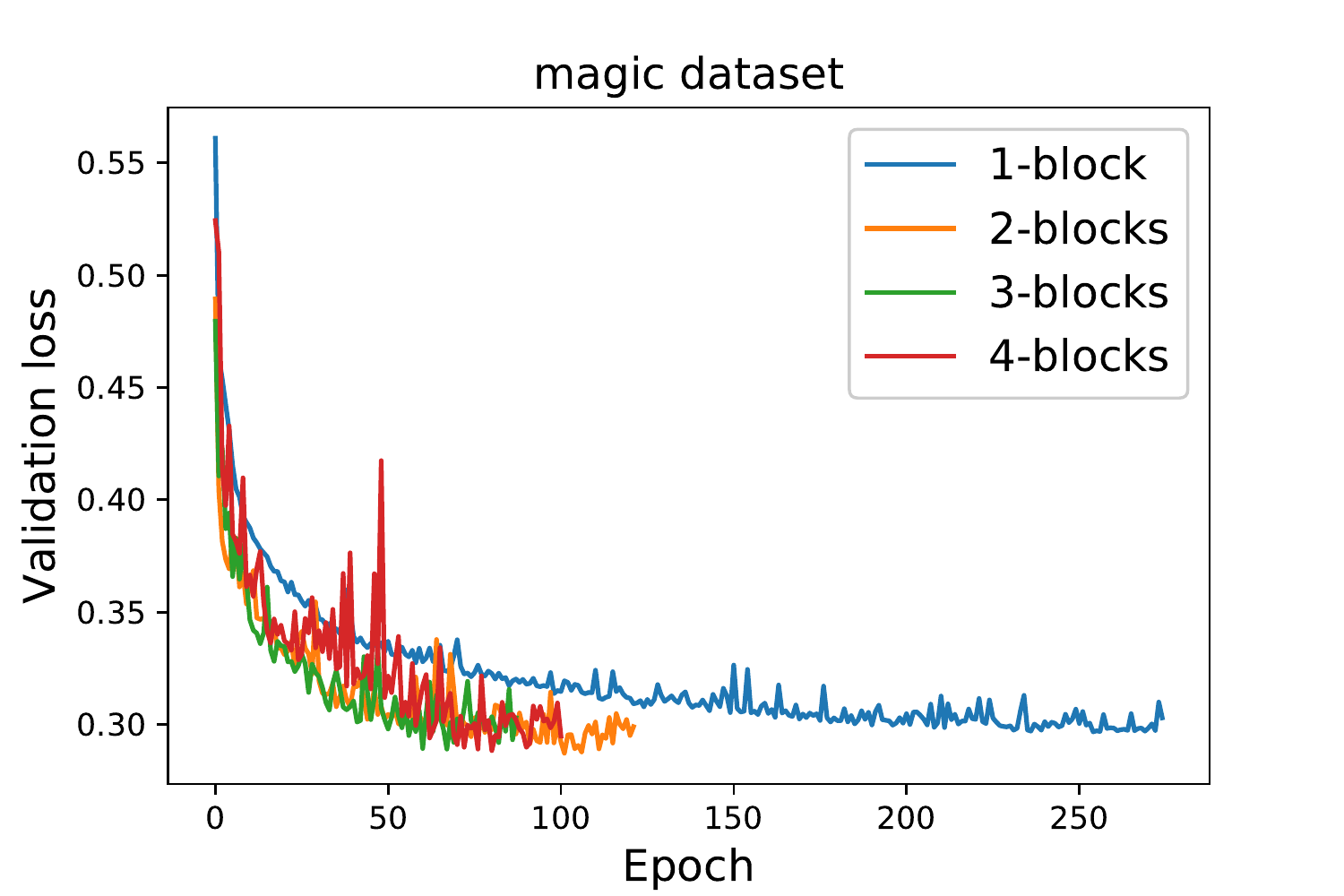}
	}

	\\
	
	\subfigure[]
	{
		\includegraphics[width=1.5in, height=1.5in]{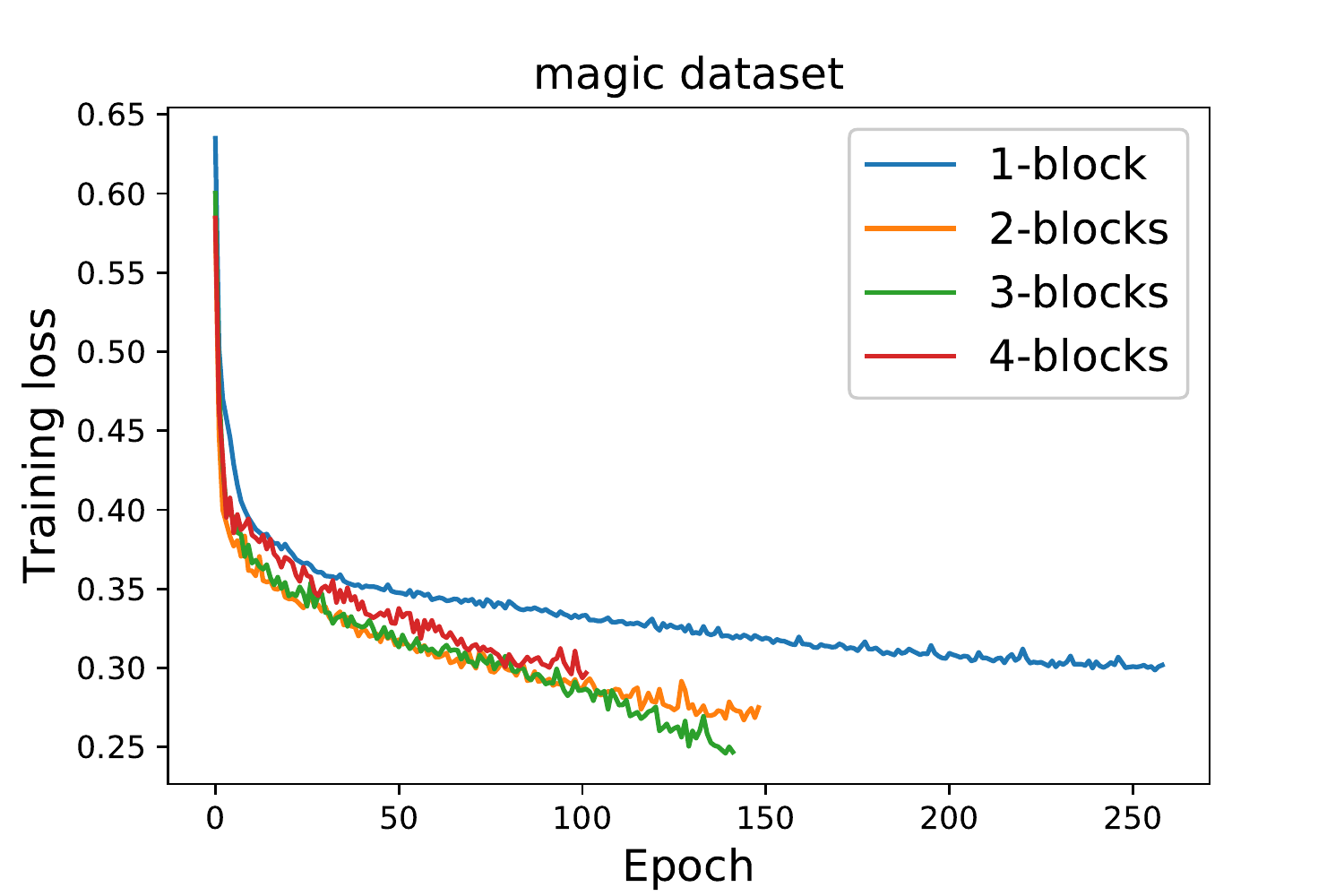}
	}
	
	&

	\subfigure[]
	{
		\includegraphics[width=1.5in, height=1.5in]{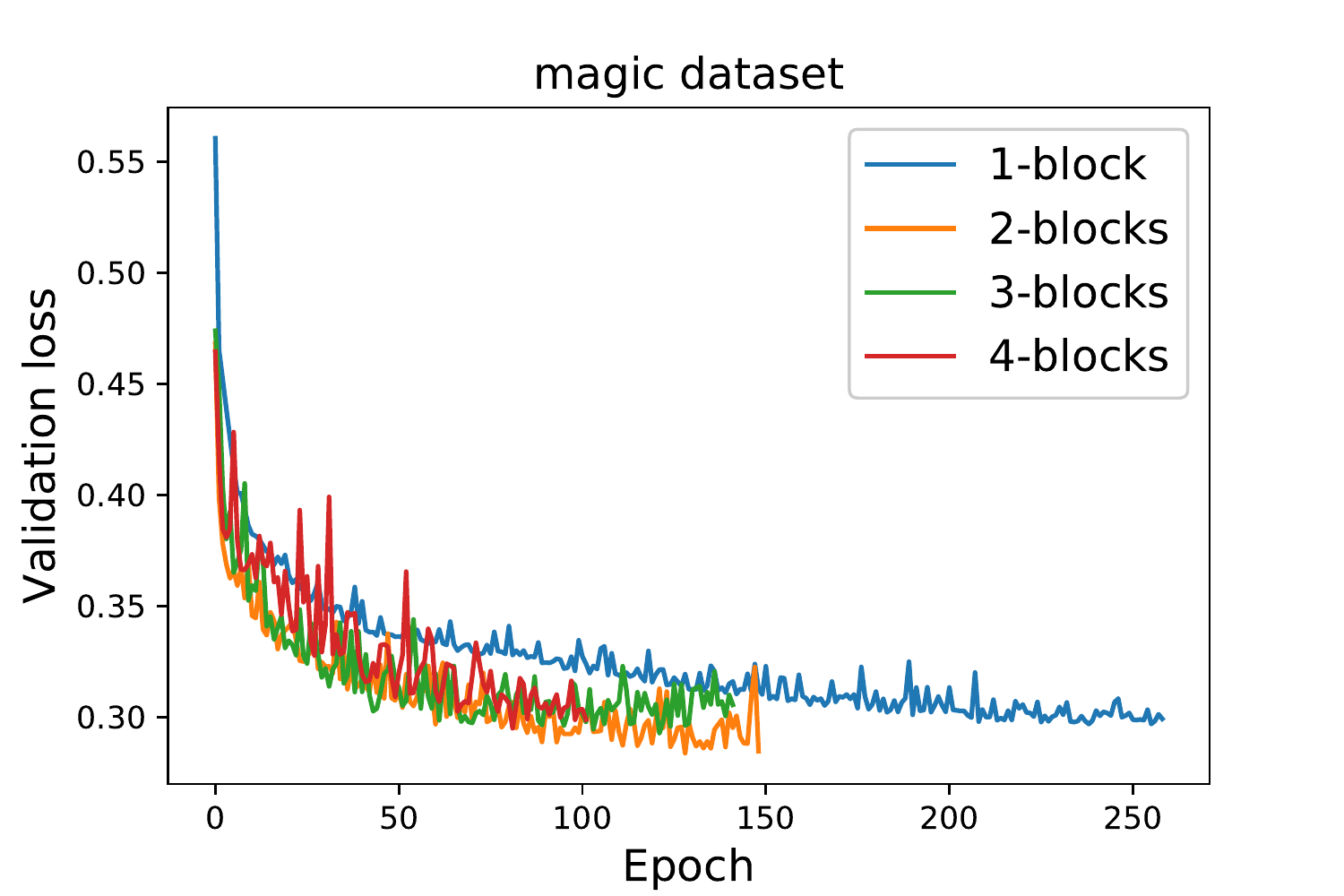}
	}

	\\
	
	\subfigure[]
	{
		\includegraphics[width=1.5in, height=1.5in]{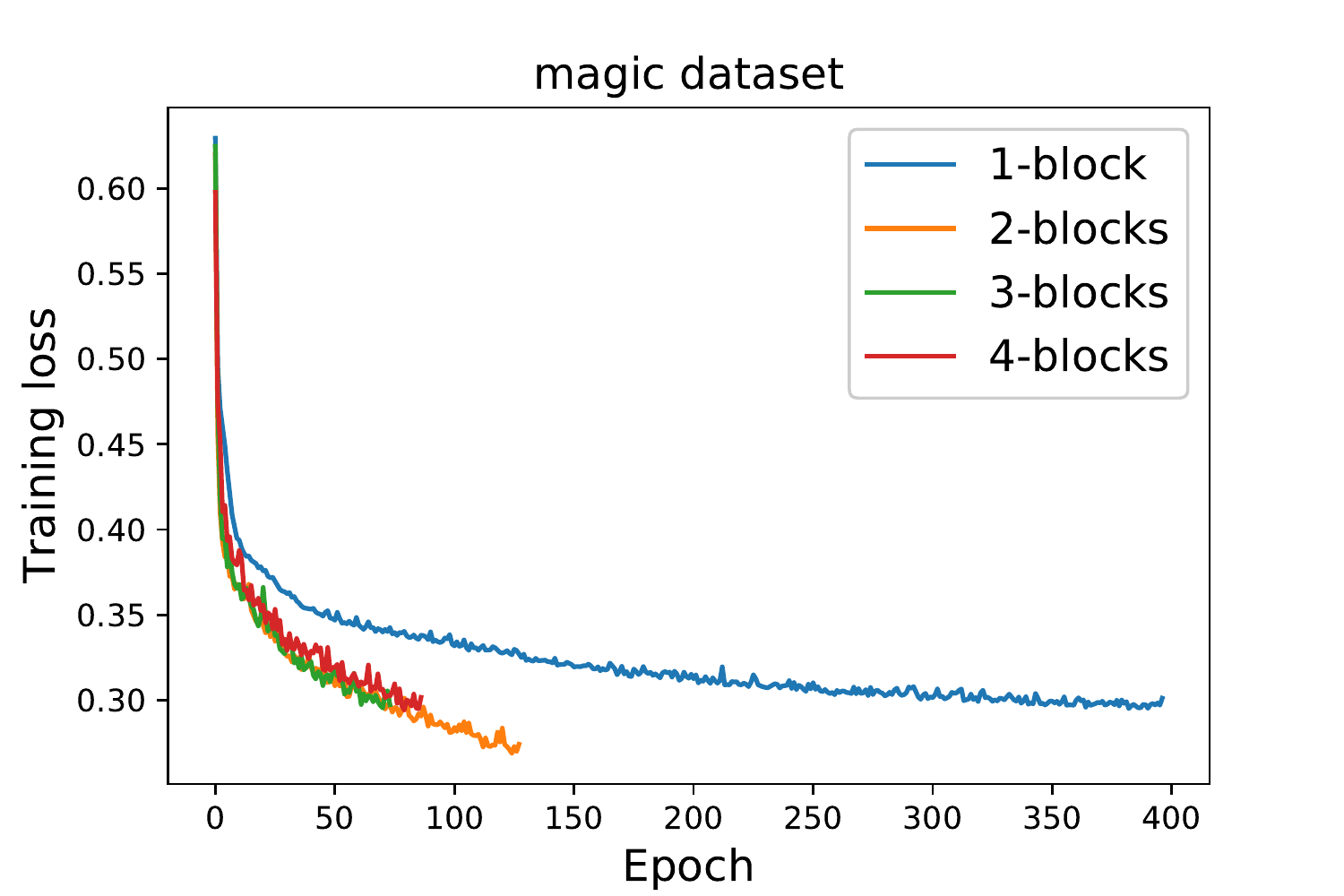}
	}

	&

	\subfigure[]
	{
		\includegraphics[width=1.5in, height=1.5in]{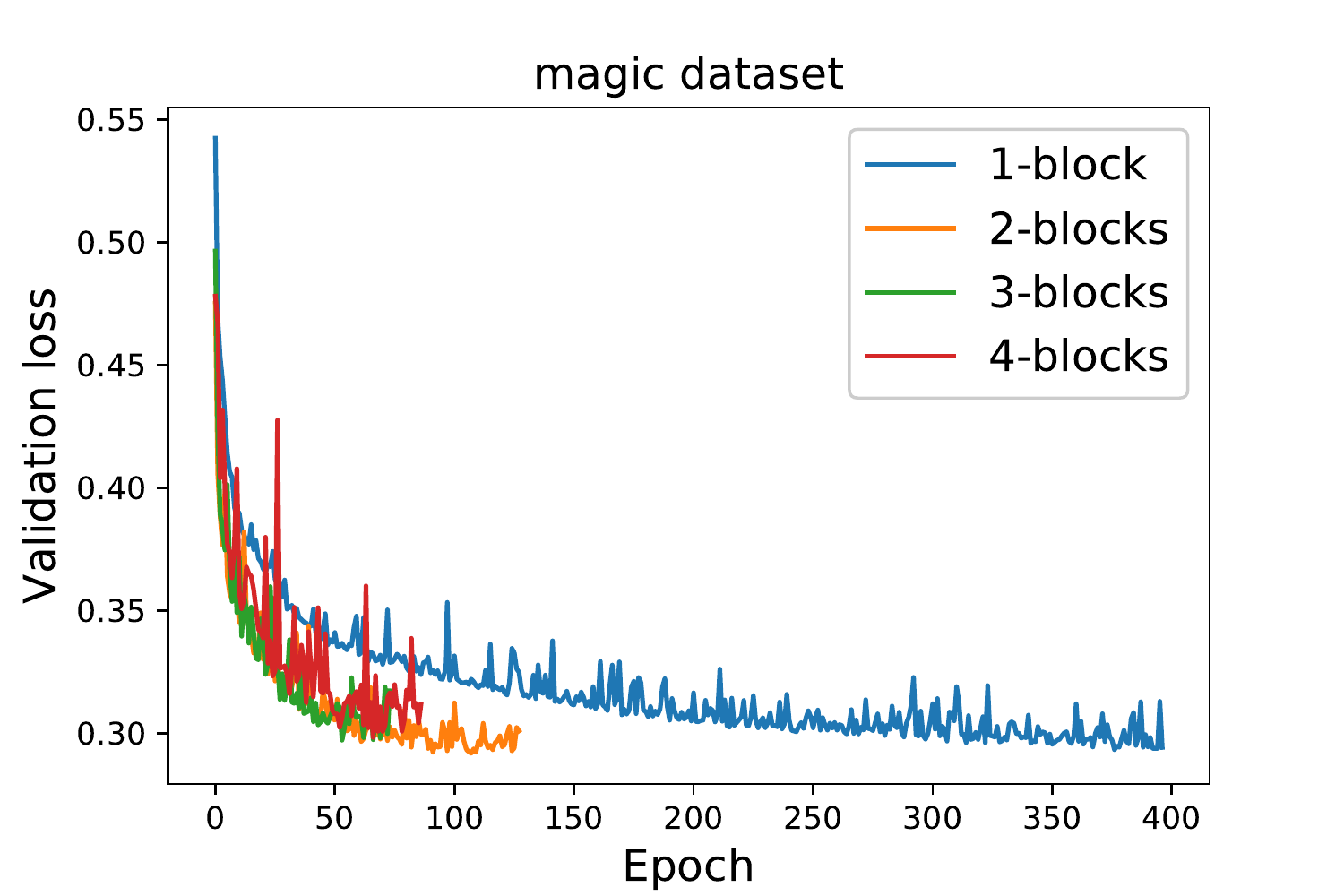}
	}

\end{tabular}
\end{center}
\vspace{-0.1in} \caption{ \label{magic_train_val_loss} Magic dataset: the training/validation loss of the deep maxout neural-kernel networks when different number of blocks and linear transformations are used. (a,b) two linear transformations. (c,d) three linear transformations. (e,f) four linear transformations.}
\end{figure}

Table \ref{img_results} summarizes a comparison between the introduced models and those in \cite{zeiler2013stochastic,tang2013deep,goodfellow2013maxout, zeiler2013stochastic, tang2013deep}. 
It can be observed from Table \ref{img_results} that the performance of the deep neural-kernel models is either comparable or better than those of SVM and convolutional neural networks. It should be noted that in our experiments we have not fully explored other regularization techniques such as dropout, hidden unit sparsity, weight constraints or an exhaustive model selection search. Next we analyse the influence of the number of neural-kernel blocks and linear transformations on the performance of the deep maxout neural-kernel networks. In particular, we vary the number of blocks in the range of 1 to 4 and the number of linear transformation in the range of 2 to 4.  
Fig. \ref{test_acc_blocks} and Fig. \ref{magic_train_val_loss} show the effect of the number of blocks and linear transformations on the accuracy, training and validation loss for the Magic dataset. In general, we can observe that the larger the number of blocks, the higher the complexity of the model is and the faster the training loss converges. However, it should be noted that increase in the complexity of the model results in increasing the risk of overfitting. Therefore, in practice, one can seek to find the optimal combinations of these two parameters using for instance a validation dataset.

\begin{figure}
	\begin{center}
		
		\renewcommand*{\arraystretch}{0.5}
		\setlength{\tabcolsep}{2pt}
		
		\begin{tabular}{cccc}

			
			\subfigure[]
			{
				\includegraphics[width=1.5in, height=1.5in]{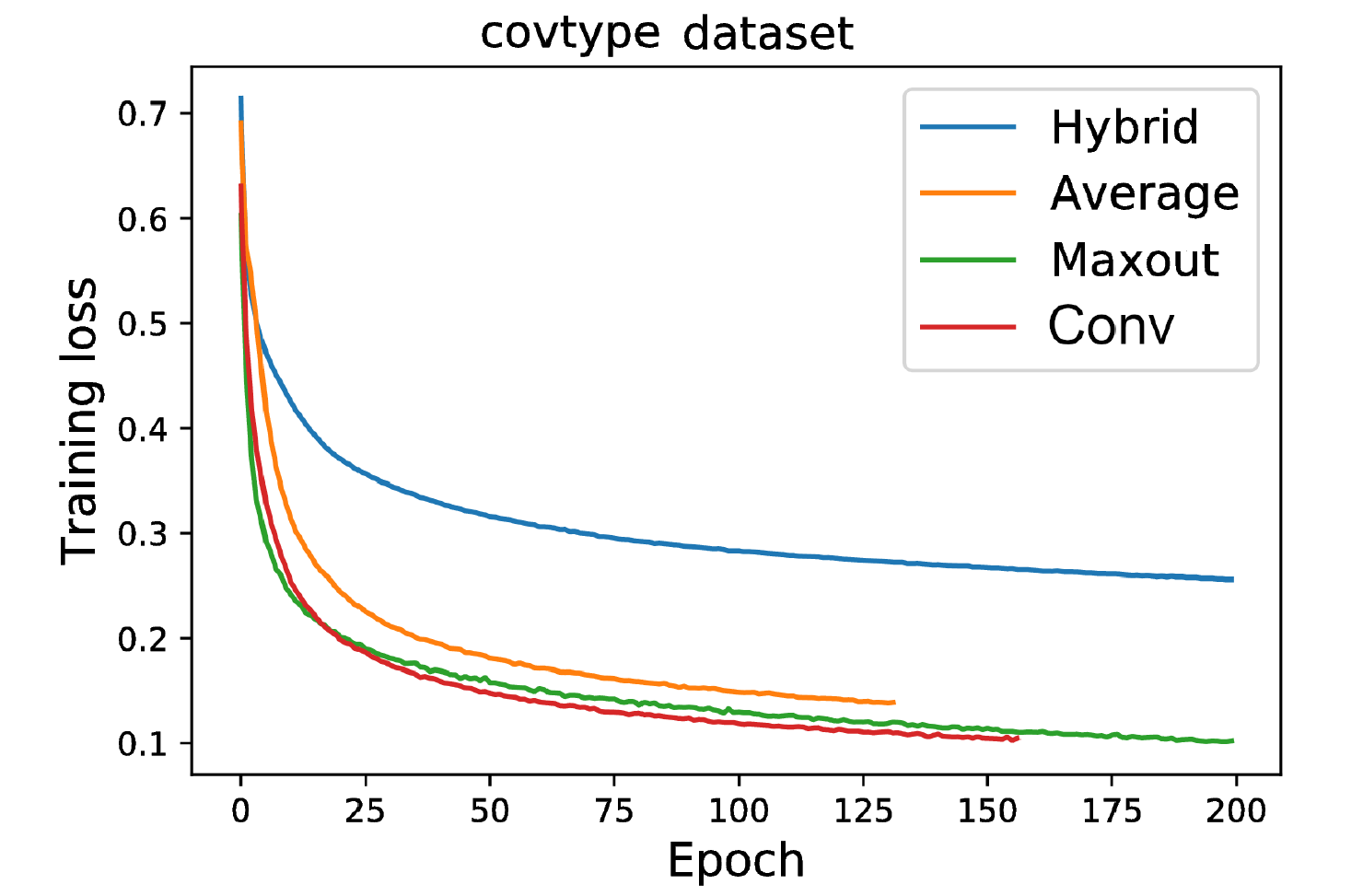}
			}
			
			&
			
			\subfigure[]
			{
				\includegraphics[width=1.5in, height=1.5in]{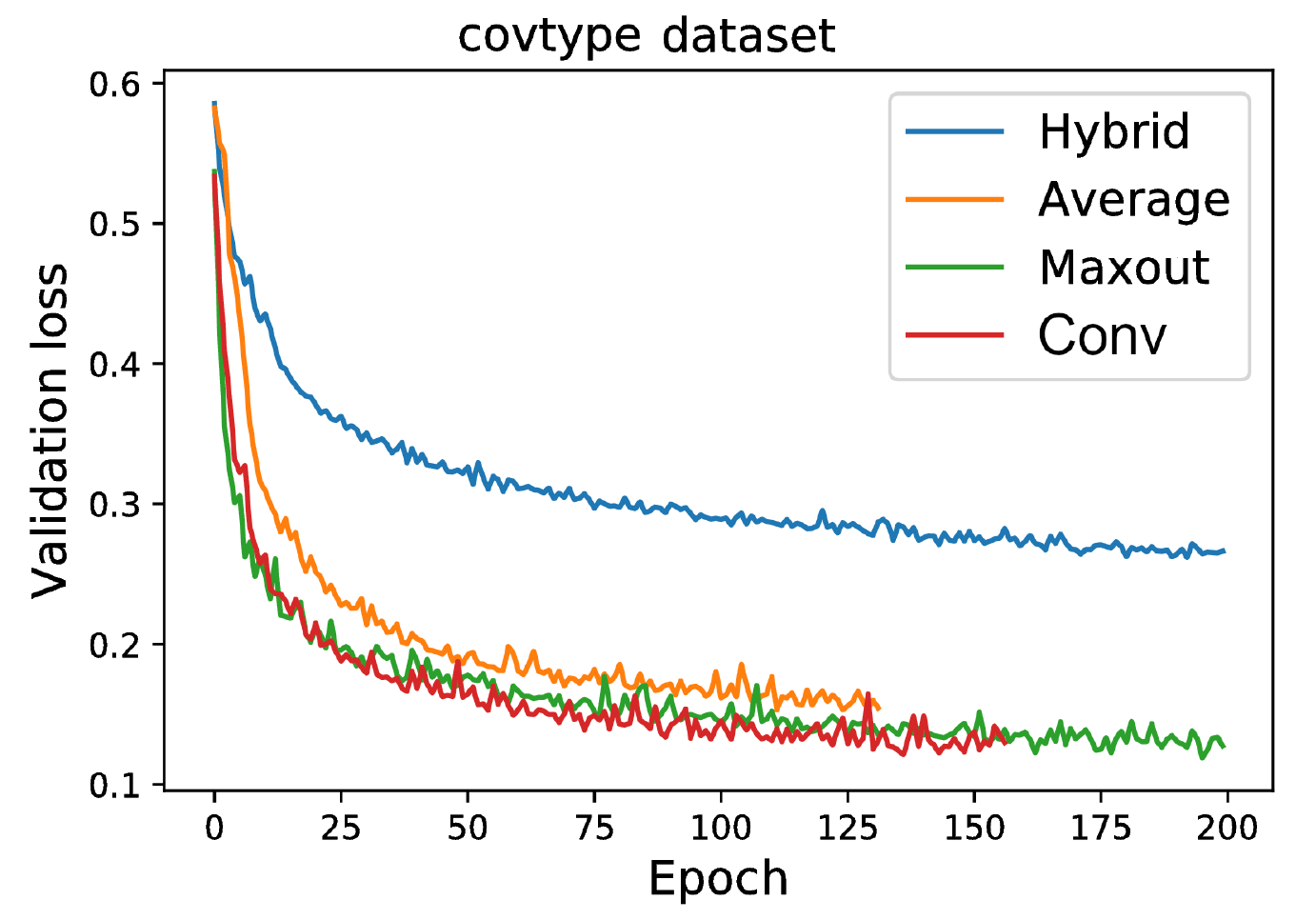}
			}
			
			\\
			
			\subfigure[]
			{
				\includegraphics[width=1.5in, height=1.5in]{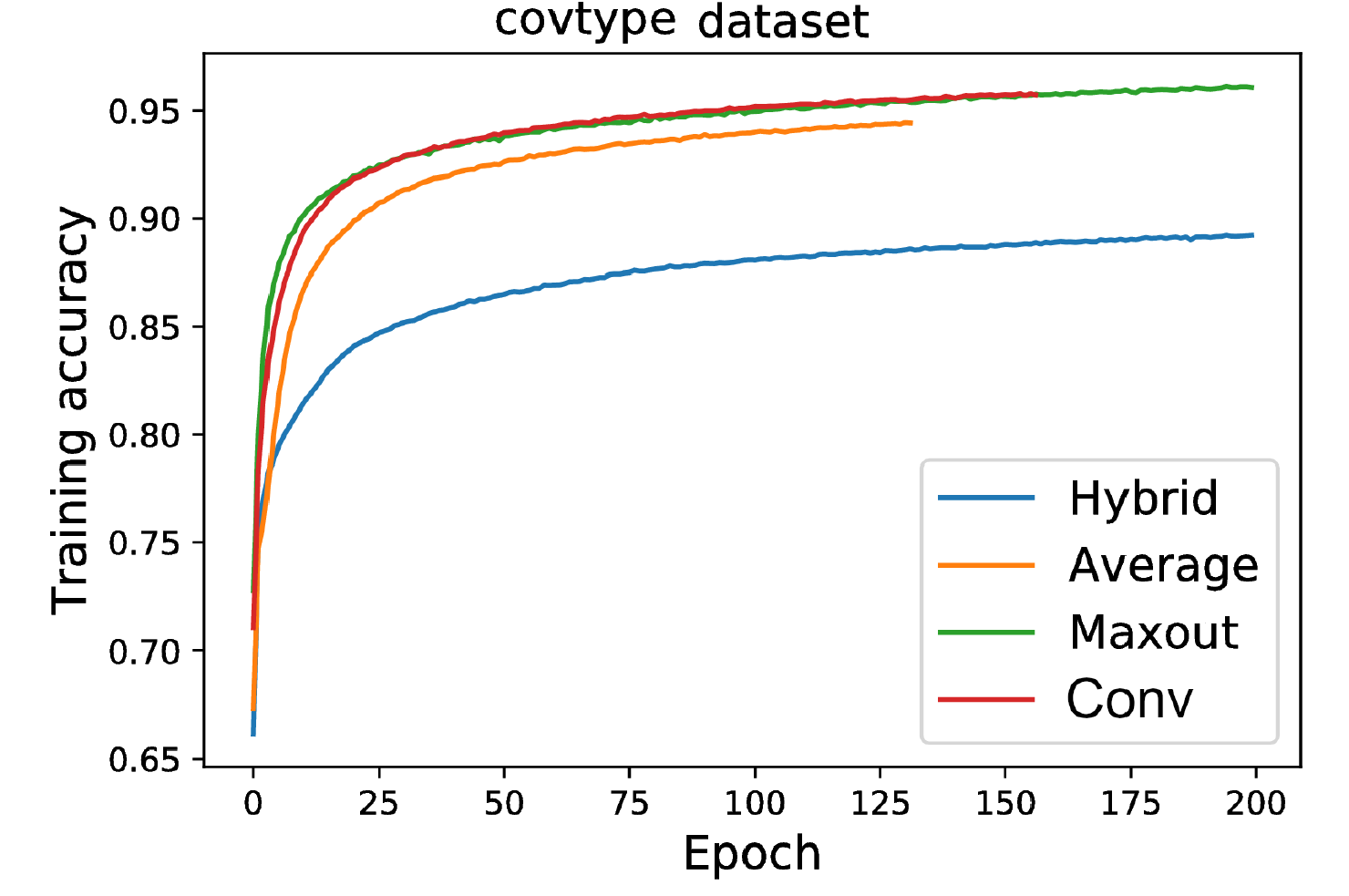}
			}

			& 
			
			\subfigure[]
			{
				\includegraphics[width=1.5in, height=1.5in]{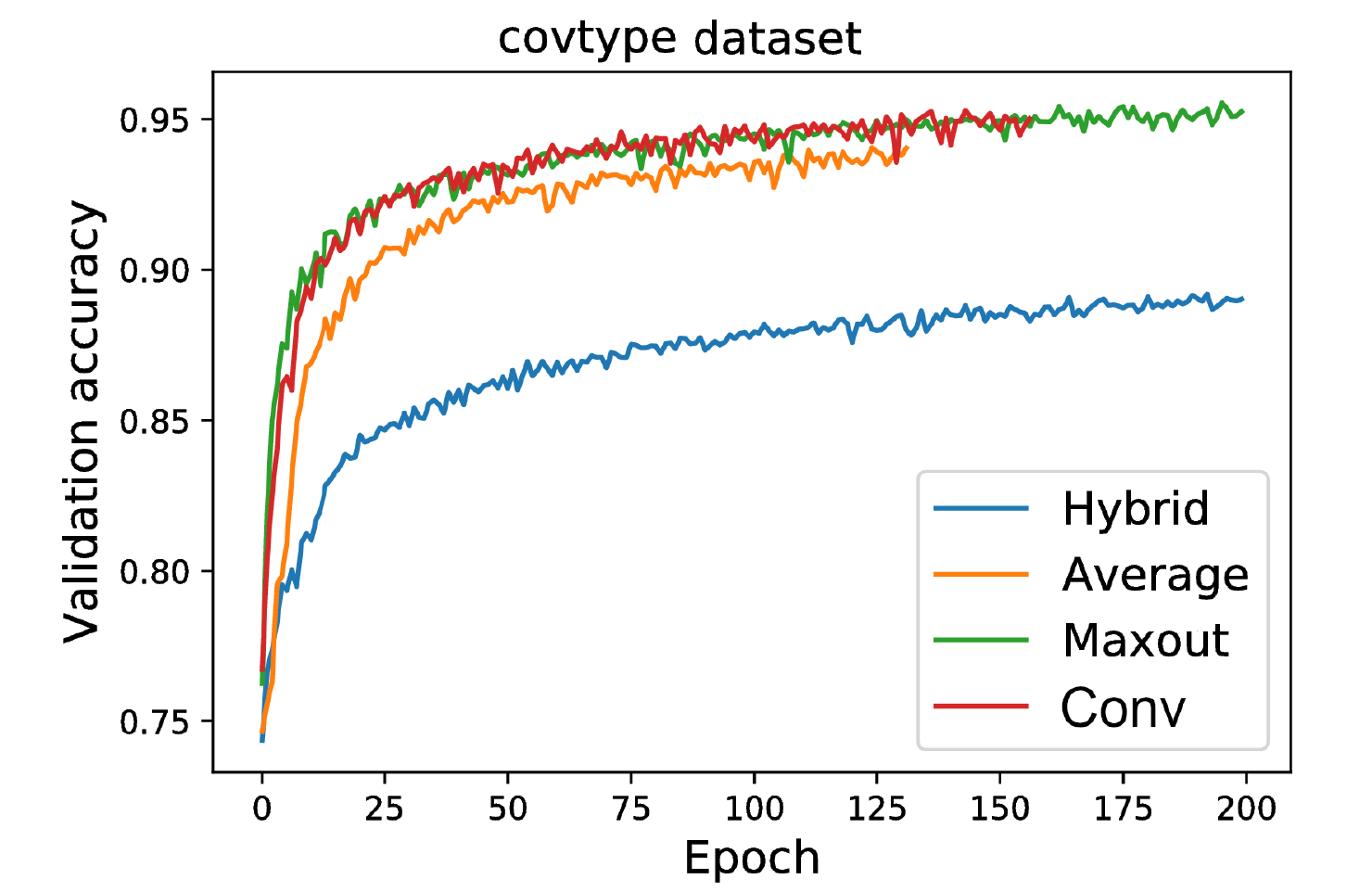}
			}

		\end{tabular}
	\end{center}
	\vspace{-0.1in} \caption{ \label{monk2_val_acc_loss} The Covertype dataset. (a,b,c,d) The training/validation loss and accuracy of the deep neural-kernel networks discussed in section \ref{deep_nk}.}
\end{figure}

\begin{table*}[htbp]   
\centering \caption{ \label{img_results} Test set error rates of various networks for MNIST, CIFAR-10, CIFAR-100 and SVHN datasets}
\renewcommand*{\arraystretch}{1.4}
\setlength{\tabcolsep}{5pt}

\begin{tabular}{lllllllllll}


& \multicolumn{4}{c}{\textbf{Test error (\%) }}  \\ \cline{2-5} 

\textbf{Method} &  MNIST & CIFAR-10 & CIFAR-100 & SVHN (\%)  \\  [0.5ex]  \hline   

CNN + Maxout Kernel Blocks &  $0.48 \%$ & $11.52 \%$   & $38.77 \%$  & $2.41 \%$   \\
CNN + Conv Kernel Blocks & $0.51 \%$ &  $11.61 \%$   &   $39.21 \%$    & $2.56 \%$  \\
Conv. maxout + Dropout \cite{goodfellow2013maxout} &  $0.45\%$ &  $11.68 \%$   &   $38.57 \%$   &  $2.47 \%$  \\
Stochastic Pooling \cite{zeiler2013stochastic} &  $0.47\%$  &    $15.13 \%$  &  $42.51 \%$  & $2.80 \%$   \\ 
2-Layer CNN + 2-Layer NN \cite{zeiler2013stochastic} &  $0.53\%$ & N.A &  N.A & N.A & \\
DLSVM \cite{tang2013deep} &  $0.87 \%$ &   $11.90 \%$  &  N.A  &  N.A  & 

\\ \hline \\ 
\end{tabular}
\end{table*}

\newpage
\section{Conclusions}
In this chapter we have discussed neural-kernel machines a framework that bridges the artificial neural network and kernel based models.
In particular, deep hybrid, average, maxout and convolutional neural-kernel architectures have been explained in details.
The exiting connections between artificial neural networks and kernel machines with explicit and implicit feature mapping are highlighted.
In this context, different ways of combining multi-scale representations of the data are explored and in particular we showed that the introduced average neural-kernel model is a special case of the neural-kernel model with pointwise convolutional pooling layer. Finally, the validity and applicability of the proposed models on several real-life benchmark datasets have been examined.



\bibliography{main}

\end{document}